\documentclass[10pt, twocolumn, journal]{IEEEtran}

\usepackage{cite}
\usepackage{graphicx}
\usepackage{epsf}
\usepackage{subcaption}
\usepackage[labelformat=simple]{subcaption}

\usepackage{amsmath}
\usepackage{comment}
\usepackage{amssymb}
\usepackage{array}
\usepackage{setspace}
\usepackage{amsthm,amssymb}
\usepackage[ruled,lined]{algorithm2e}
\usepackage{multirow}
\usepackage{tabularx}
\usepackage{xfrac}
\usepackage{caption}
\usepackage{breqn}
\usepackage{bbm}
\usepackage{bm}
\usepackage{enumerate}
\usepackage{algorithmic}
\usepackage{dsfont}
\usepackage{float}
\usepackage{diagbox}
\usepackage{mathtools}
\usepackage{adjustbox}
\usepackage{comment}
\usepackage{color}

\newcommand{\tabincell}[2]{\begin{tabular}{@{}#1@{}}#2\end{tabular}}

\graphicspath{{Fig/},{fig/}}

\newcommand{\fig}{Fig.\xspace}

\begin{document}

\title{BTS: Bifold Teacher-Student in Semi-Supervised Learning for Indoor Two-Room Presence Detection Under Time-Varying CSI}

\author{
Li-Hsiang Shen,~\IEEEmembership{Member,~IEEE},
An-Hung Hsiao, 
Kai-Jui Chen,
Tsung-Ting Tsai,
Kai-Ten Feng,~\IEEEmembership{Senior Member,~IEEE}
}


\maketitle

\begin{abstract}
In recent years, indoor human presence detection based on supervised learning (SL) and channel state information (CSI) has attracted much attention. However, existing studies that rely on spatial information of CSI are susceptible to environmental changes which degrade prediction accuracy. Moreover, SL-based methods require time-consuming data labeling for retraining models. Therefore, it is imperative to design a continuously monitored model using a semi-supervised learning (SSL) based scheme. In this paper, we conceive a bifold teacher-student (BTS) learning approach for indoor human presence detection in an adjoining two-room scenario. The proposed SSL-based primal-dual teacher-student network intelligently learns spatial and temporal features from labeled and unlabeled CSI datasets. Additionally, the enhanced penalized loss function leverages entropy and distance measures to distinguish drifted data, i.e., features of new datasets affected by time-varying effects and altered from the original distribution. Experimental results demonstrate that the proposed BTS system accomplishes an averaged accuracy of around 98\% after retraining the model with unlabeled data. BTS can sustain an accuracy of 93\% under the changed layout and environments. Furthermore, BTS outperforms existing SSL-based models in terms of the highest detection accuracy of around 98\% while achieving the asymptotic performance of SL-based methods.
\end{abstract}

\begin{IEEEkeywords}
	Teacher-student learning, semi-supervised learning, time-varying environment, human presence detection, channel state information (CSI).
\end{IEEEkeywords}

%

{\let\thefootnote\relax\footnotetext
{Li-Hsiang Shen is with the Department of Communication Engineering, National Central University, Taoyuan 320317, Taiwan
(email: shen@ncu.edu.tw).}
}

{\let\thefootnote\relax\footnotetext
{An-Hung Hsiao, Kai-Jui Chen, Tsung-Ting Tsai, and Kai-Ten Feng are with the Department of Electronics and Electrical Engineering, National Yang Ming Chiao Tung University, Hsinchu 300093, Taiwan. (email: e.c@nycu.edu.tw, cary.ee09@nycu.edu.tw, yoyo6339948.c@nycu.edu.tw, and ktfeng@nycu.edu.tw) (Corresponding author: Kai-Ten Feng)}
}

\section{Introduction}\label{Introduction}
Wireless networking devices have gained popularity, especially Wi-Fi, offering convenience in daily lives. However, many Internet-of-Thing (IoT) devices lack security and safety features, emphasizing the need for wireless devices with intrusion protection as one of the applied scenarios of human presence. Additionally, in-home wireless sensing devices can assist other indoor applications, such as elderly healthcare, room vacancy, and remote monitoring \cite{add5}. In practical applications, device-based wireless gadgets such as smart phones \cite{Smartphone1, Smartphone2} and smart watches \cite{Smartwatch1, Smartwatch2} are commonly used for home security and safety sensing. Nevertheless, these devices may not always be carried around in indoor scenarios, rendering them less effective. On the contrary, device-free solutions, such as cameras \cite{IPcamera1, IPcamera2, IPcamera3}, are also utilized for home security and safety sensing applications. However, privacy concerns arise when these methods rely on image analysis for indoor detection. Alternatively, some applications focus on the utilization of radar-based sensing owing to its asymptotic resolution to imaging \cite{add1}. However, both images and radar sensors may lead to a difficulty of detecting blind spots due to potential obstacles in the indoor environments. Therefore, there has been a surge of interest among researchers in exploring the utilization of commercial Wi-Fi devices to detect people indoors, as radio frequency (RF) wireless signals have unique characteristics that can penetrate walls \cite{add2}. With the aid of wireless signals, channel state information (CSI) has become increasingly popular due to its versatile and informative features in both spatial, temporal, and spectral domains \cite{CSI1, CSI2, CSI3, acm}. In \cite{add6}, conventional optimization-based signal preprocessing of WiFi CSI is performed for sensing applications. The work of \cite{add3} has discussed the subtle changes in CSI observed under stationary human presence through a wall, compared to the more pronounced changes caused by a moving person. Moreover, with the advancements of deep learning, researchers have attempted to design deep learning-based Wi-Fi detection systems by using Wi-Fi CSI signals \cite{Deep_Learning1, Deep_Learning2, Deep_Learning3, Deep_Learning4, Deep_Learning5, Deep_Learning6, Deep_Learning7, Deep_Learning8, Deep_Learning9, add7}. The authors in \cite{add4} have also investigated the future advances, challenges, and opportunities for WiFi-based human sensing using deep learning.

Various approaches have been proposed for indoor human detection based on CSI. The fine-grained fingerprinting method in \cite{CSI1} compares real-time streaming CSI data with an offline radio map in a database. Learning-based methods, such as support vector machines (SVM) classification \cite{CSI2} and autoencoder (AE) approaches \cite{Deep_Learning1, new_attention}, have also been employed to improve detection performance  based on CSI data. Alternatively, some approaches focus on capturing the temporal correlation of sequential CSI data by applying recurrent neural networks (RNNs) \cite{tk_rnn, new_chu} such as long short-term memory (LSTM) \cite{Deep_Learning3, tk_lstm} and GRU \cite{Deep_Learning4, Deep_Learning5}. Other related architectures combine AE, CNN, and LSTM for device-free human activity recognition, such as autoencoder long-term recurrent convolutional network (AE-LRCN) in DeepSence \cite{Deep_Learning6}. More advanced architecture of WiSMLF adopts module fusion and scoring on attention is proposed in \cite{add7}. Additionally, some methods attempt to overcome the lack of CSI data using generative adversarial networks (GANs) during training \cite{Deep_Learning7}. However, CSI is susceptible to time-varying effects induced by environmental changes \cite{time_varying}. When severe time-varying effects occur, the CSI pattern can be completely different from the original one, which requires model retraining in order to adapt to the new data distribution. Unfortunately, most of the aforementioned approaches for indoor human detection are supervised learning (SL) methods, which require time-consuming and laborious labeling of new data when encountering time-varying effects.

To address the above issue, semi-supervised learning (SSL) \cite{tk_ssl} is beneficial to adopt a portion of the labeled CSI dataset to train a new and unlabeled data. This scheme can eliminate the need to manually label new data, which can be further reused for subsequent retraining, improving efficiency and reducing costs. Several existing SSL approaches have been proposed for human activity recognition based on CSI data. MCBAR \cite{MCBAR} and CsiGAN \cite{CsiGAN} utilize GAN to distinguish between real and fake data, whereas DADA-AD \cite{DADA} leverages transfer learning and domain adaptation. SemiC-HAR \cite{SemiC_HAR} employs a temporal classifier to label the unlabeled data before training an encoder on both labeled and unlabeled datasets. However, these methods often require ground truth for some unlabeled data. Most of existing works do not consider significant differences in the distributions of labeled and unlabeled datasets. Additionally, they only conduct experiments in a single room, simplifying the training difficulty. Note that only CsiGAN has demonstrated multi-room scenarios without limiting the labeled and unlabeled data to similar distributions. However, CsiGAN is with high computational complexity, which requires a long training time due to the adoption of CycleGAN \cite{CycleGAN} in their algorithm.

\begin{table}
\begin{center}
\footnotesize
\setstretch{1.1}
\caption {Acronyms}
    \begin{tabular}{|l|l|}
    \hline
        Acronym & Definition \\        
        \hline\hline
		AE	&	autoencoder	\\ \hline
AP	&	access point	\\ \hline
BTS	&	bifold teacher-student	\\ \hline
CD	&	confidence distribution 	\\ \hline
CNN	&	convolutional neural network	\\ \hline
CSI	&	channel state information	\\ \hline
CTQ	&	cross-teacher quadratic	\\ \hline
DNN	&	deep neural network	\\ \hline
GAN	&	generative adversarial network	\\ \hline
IoT	&	Internet-of-Thing	\\ \hline
MIMO	&	multiple-input multiple-output	\\ \hline
MLP	&	multilayer perceptron	\\ \hline
MPL	&	meta pseudo label	\\ \hline
OFDM	&	orthogonal frequency-division multiplexing	\\ \hline
ReLU	&	rectified linear unit	\\ \hline
RF	&	radio frequency	\\ \hline
RNN	&	recurrent neural network	\\ \hline
RX	&	receiver	\\ \hline
SL	&	supervised learning	\\ \hline
SSL	&	semi-supervised learning	\\ \hline
SVM	&	support vector machine	\\ \hline
TCE	&	transformative cross entropy	\\ \hline
TS	&	teacher-student	\\ \hline
TX	&	transmitter	\\ \hline
UICE	&	unlabeled indication cross entropy	\\ \hline

		\end{tabular} \label{abbr}
\end{center}
\end{table}

Therefore, it becomes compellingly imperative to address the challenges of time-varying effects and achieve high-precision for human presence detection in adjoining room scenarios. Inspired by the concept of knowledge discillation \cite{kn}, we propose a novel bifold teacher-student (BTS) network. Our approach is motivated by the meta pseudo labels (MPL) \cite{MPL} and communicative teacher-student model \cite{tk_ts}, which frames the teacher-student training process as a bi-level optimization problem. To further improve performance and address a more challenging time-varying scenario, we extend MPL to a two teacher-student network design that incorporates both temporal and spatial perspectives using transformer encoder \cite{Attention} and ResNet \cite{ResNet} architectures. Our previous work \cite{SAS_PD} has demonstrated the feasibility of using a transformer encoder based teacher-student network for this problem, but this paper aims to further enhance accuracy through leveraging the response of data in different cases. Our proposed system is conducted under a more complicated problem with adjacent room scenario and with fewer number of equipped antennas. A table for acronyms is established in Table \ref{abbr}. The main contributions of this paper are summarized as follows.
\begin{itemize}
	\item We have conceived a semi-supervised CSI-based bifold teacher-student presence detection system in an adjoining two-room scenario to resolve the problems of indoor human presence detection, especially the laborious labeling and drifted datasets issue due to time-varying channel. We consider four cases including empty room, human presence in either one of two rooms, and people presence in both rooms.
	
	\item BTS overcomes the time-varying problem by leveraging the characteristics of fluctuating channels in different human presence cases. We measure a time-invariant indicator for each case by computing subcarrier-wise entropy within a time period, which generates confidence distribution of cases viewed as the prior knowledge to CSI data.

	\item A series of loss functions are designed to enhance performance and resolve various problems, including (1) self-supervised learning between two teacher-student networks by minimizing their distance in latent space, (2) data drift evaluation by constraining the dataset in hypersphere, and (3) regulating inconsistent distribution of drifted data by adding confidence distribution.

	\item We have collected and observed different responses of CSI in the adjoining room scenario in different fluctuation levels of time-varying problem. Experiments are conducted to demonstrate that the proposed BTS system can achieve the highest accuracy compared to different semi-supervised based methods. Moreover, the accuracy of BTS can asymptotically approach the performance of supervised learning and even surpasses them in some scenarios.

\end{itemize}
\section{System Architecture and Preliminary Observations} \label{SYS_PRE}

\subsection{System Architecture}
\label{SYS_ARCH}

\begin{figure}[t]
\centering
\includegraphics[width=3.3in]{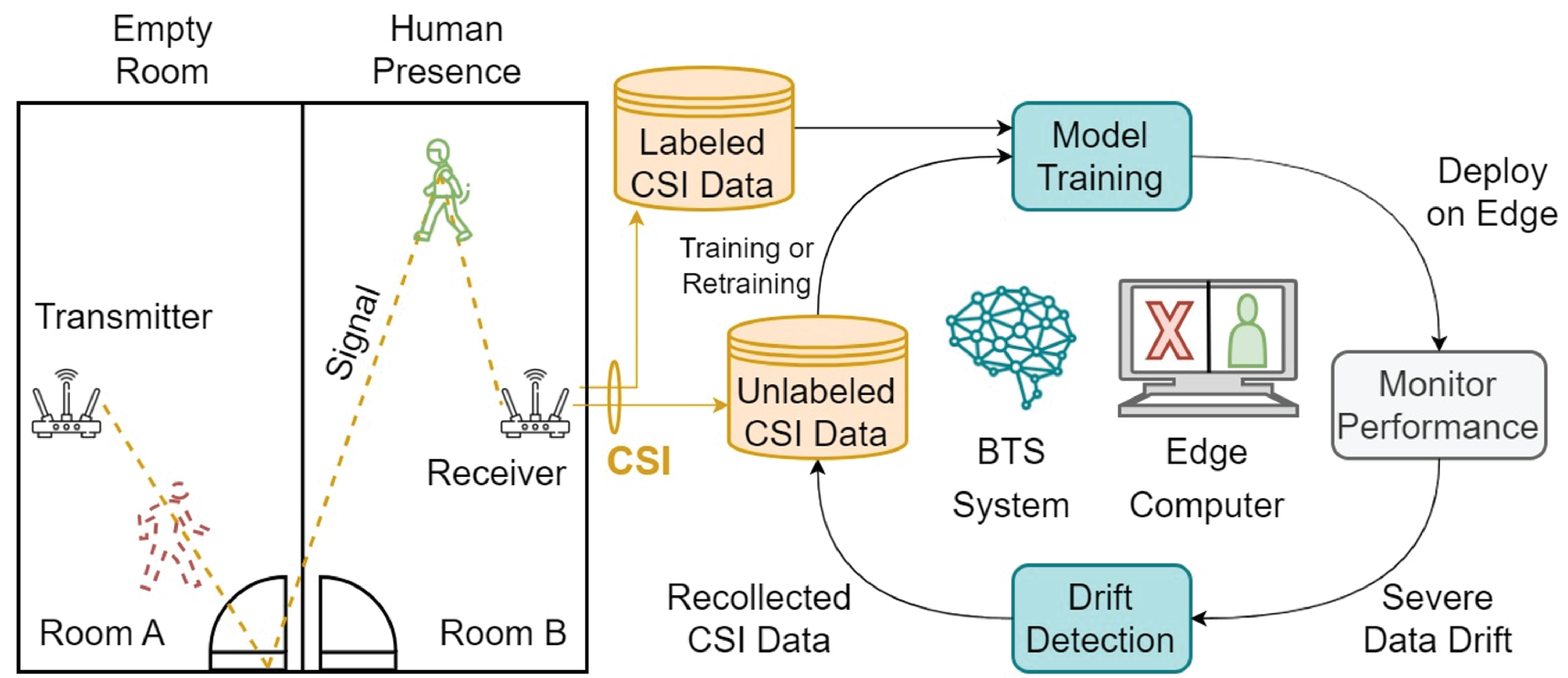}
\caption{\small System architecture for human presence detection in adjacent rooms.}
\label{fig:systemArchitecture}
\end{figure}

We consider two Wi-Fi access points (APs) in each room, as illustrated in \fig \ref{fig:systemArchitecture}. A single pair of APs, serving as a transmitter (TX) and a receiver (RX), is considered in an adjoining room, with or without human presence. For example, the TX in room A periodically sends an RF signal (dotted line) to the RX in room B, where human presence only takes place in room B. The RX estimates the CSI and delivers it to the database in the edge computer for further data training. Both labeled and unlabeled data will be fed into the computing edge for model training of BTS system. After deploying the well-trained model on the edge, human existence can be predicted in the two-room scenario. However, CSI is not always stable due to external factors. Our proposed system can detect the severe \textit{data drift} in CSI and retrain the deep neural networks (DNN) model from the obtained drifted and recollected data. Hence, the unlabeled database will be updated every time the drift occurs, whilst the labeled database will remain unchanged. We consider $C=4$ cases of presence detection\textsuperscript{\ref{note1}}\footnotetext[1]{Localization can be another potential application for detecting an empty room and human presence. However, it requires a more complex system and deep learning design. Furthermore, data collection and labeling for a large number of reference points for localization are potential challenges. \label{note1}}, including \textbf{Case 1} ($c=1$): Empty in both rooms, \textbf{Case 2} ($c=2$): A person in room A and empty in room B, \textbf{Case 3} ($c=3$): Empty in room A and a person in room B, as well as \textbf{Case 4} ($c=4$): people in both rooms. Note that we consider the most challenging task that at most one people being present in each room, which leads to subtle changes in CSI compared to lots of people in rooms.

\subsection{CSI Modeling} 
\label{CSI}

\begin{figure}[t]
     \centering
     \begin{subfigure}[b]{0.4\textwidth}
         \centering
         \includegraphics[width=1\linewidth]{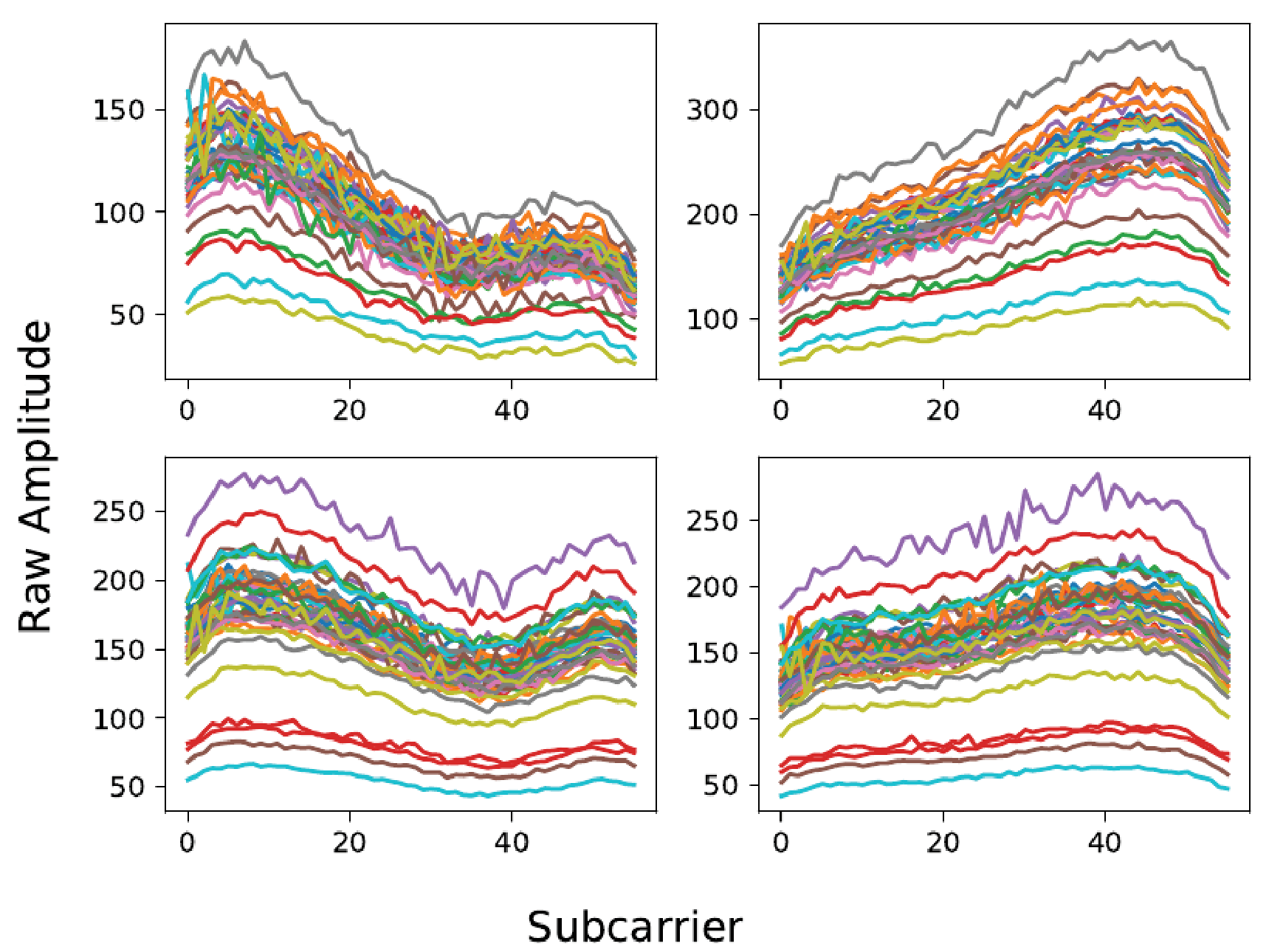}
         \caption{}
         \label{fig:raw_CSI}
     \end{subfigure}
     \begin{subfigure}[b]{0.4\textwidth}
         \centering
         \includegraphics[width=1\linewidth]{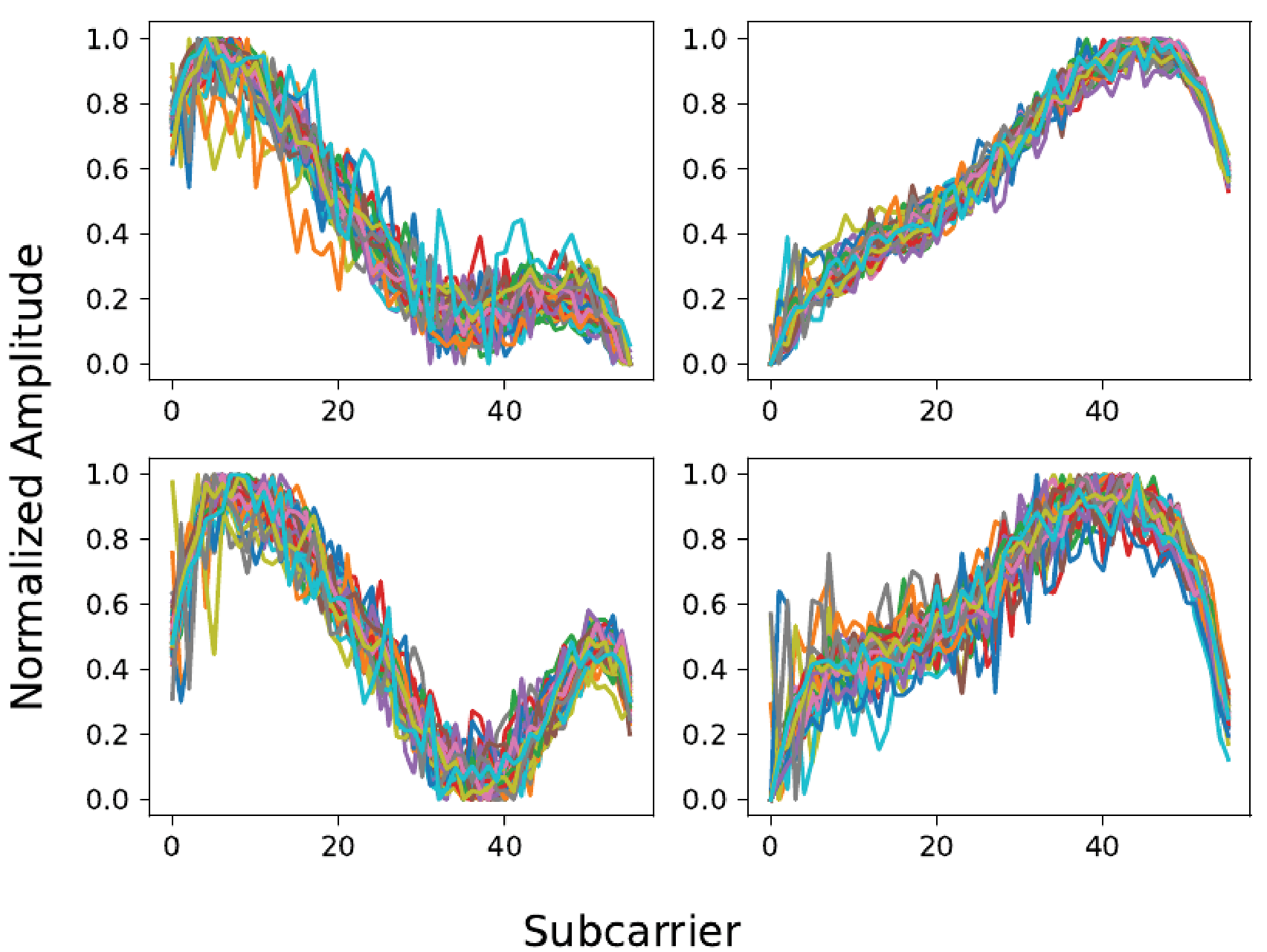}
         \caption{}
         \label{fig:normalized_CSI}
     \end{subfigure}
        \caption{\small Amplitude response of CSI (1) before and (2) after normalization. Note that we consider two TX and two RX antennas generating $K=4$ transmission pairs, labeled as $k\in\{1,2,3,4\}$ for respective subplots at top-left, top-right, bottom-left and bottom-right. The x-axis indicates the index of $S=52$ subcarrier, whilst y-axis is value of CSI amplitude.}
        \label{fig:raw_norm_CSI}
\end{figure}

CSI is considered essential information for estimating the wireless channel, as it represents the combination of fading and propagation effects. This information can be acquired based on orthogonal frequency-division multiplexing (OFDM) and multiple-input multiple-output (MIMO) compatible with existing up-to-date IEEE 802.11 protocols. In the OFDM system, all subcarriers are orthogonal to each other within a channel, which diversifies channel utilization. On the other hand, the MIMO technique provides channel diversity to enrich spatial information by propagating the RF signal from TX to RX, supporting multiple antenna pairs for CSI. We consider $S$ subcarriers, $K$ antenna pairs and $T$ packets. Employing these two techniques, we can establish a channel model by the combined MIMO-OFDM system in an indoor environment, which can be formulated as
\begin{align}
    y_{t,s,k}=h_{t,s,k}x_{t,s,k}+w_{t,s,k},
\end{align}
where $y_{t,s,k}$ and $x_{t,s,k}$ represent the received and transmitted signals with the $s^{th}$ subcarrier of the $k^{th}$ antenna pair at the $t^{th}$ packet, respectively. Notation $h_{t,s,k}$ is the channel response, whilst $w_{t,s,k}$ is additive white Gaussian noise (AWGN). To estimate the channel model, the CSI of the $s^{th}$ subcarrier and the $k^{th}$ antenna pair at the $t^{th}$ packet can be acquired as \cite{new_cronos}
\begin{align}\label{eq:h_t,s,k}
	\hat{h}_{t,s,k} = \frac{y_{t,s,k}}{x_{t,s,k}} = \left| \hat{h}_{t,s,k} \right| e^{j\sin \left( \angle \hat{h}_{t,s,k} \right)},
\end{align}
where $|\hat{h}_{t,s,k}|$ indicates the CSI amplitude response, and $\angle\hat{h}_{t,s,k}$ corresponds to the CSI phase response. Note that only the amplitude response of CSI is considered in this work since the phase information has been studied to be indescribable \cite{DeepFi}.
Hence, the estimated CSI matrix at the receiver at time $t$ can be obtained as
\begin{align}\label{csi_matrix}
\hat{\boldsymbol{H}}_t = \begin{bmatrix} \hat{h}_{t,1,1}&\dots&\hat{h}_{t,1,k}&\dots&\hat{h}_{t,1,K}
\\
\hat{h}_{t,2,1}&\dots&\hat{h}_{t,2,k}&\dots&\hat{h}_{t,2,K}
\\
\vdots&\ddots&\vdots&\ddots&\vdots
\\
\hat{h}_{t,S,1}&\dots&\hat{h}_{t,S,k}&\dots&\hat{h}_{t,S,K} \end{bmatrix}.
\end{align}
Moreover, the normalization over antenna pairs is conducted to the amplitude response of CSI. The normalized CSI with $s^{th}$ subcarrier and $k^{th}$ antenna pair at $t^{th}$ packet can be expressed as
\begin{align}\label{eq:normalization}
    \bar{h}_{t, s, k} = norm\left(\hat{h}_{t,s,k}\right)=\frac{\left|\hat{h}_{t, s, k}\right|-\mathop{\min}\limits_{\tilde{s}}\left|\hat{h}_{t, \tilde{s}, k}\right|}{\mathop{\max}\limits_{\tilde{s}}\left|\hat{h}_{t, \tilde{s}, k}\right|-\mathop{\min}\limits_{\tilde{s}}\left|\hat{h}_{t, \tilde{s}, k}\right|},
\end{align}
where $\tilde{s}=\{1,2, \ldots, S\}$, and $|\hat{h}_{t, s, k}|$ represents the amplitude response of raw CSI. In order to perform pair-wise normalization in $\eqref{eq:normalization}$, the minimum and maximum values of the amplitude response of raw CSI from all $S$ subcarriers at the $k^{th}$ antenna pair are selected. Fig. \ref{fig:raw_norm_CSI} highlights the difference between the amplitude response of raw CSI and that of normalized CSI. Observed from Figs. \ref{fig:raw_CSI} and \ref{fig:normalized_CSI}, normalization is necessary to remove undesirable on-device mechanisms that are unrelated to CSI, such as automatic gain control, in order to obtain appropriate CSI features.

\subsection{Preliminary CSI Observation}
\label{Observation}

\subsubsection{Human Presence in Adjoining Room}\label{Observation1}

\begin{figure}[t]
     \centering
     \begin{subfigure}{0.24\textwidth}
         \centering
         \includegraphics[width=1\linewidth]{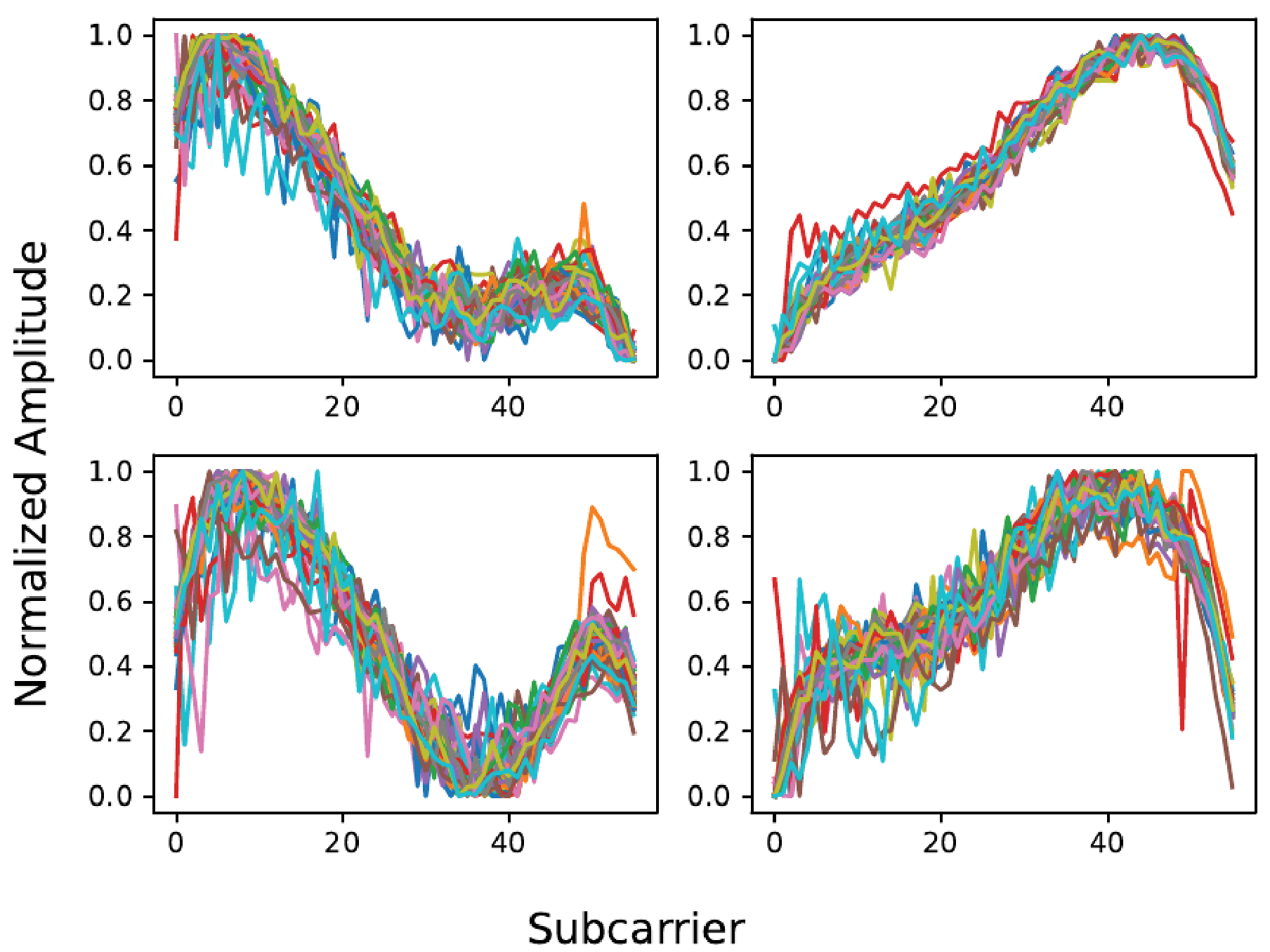}
         \caption{}
         \label{fig:case0}
     \end{subfigure}
     \begin{subfigure}{0.24\textwidth}
         \centering
         \includegraphics[width=1\linewidth]{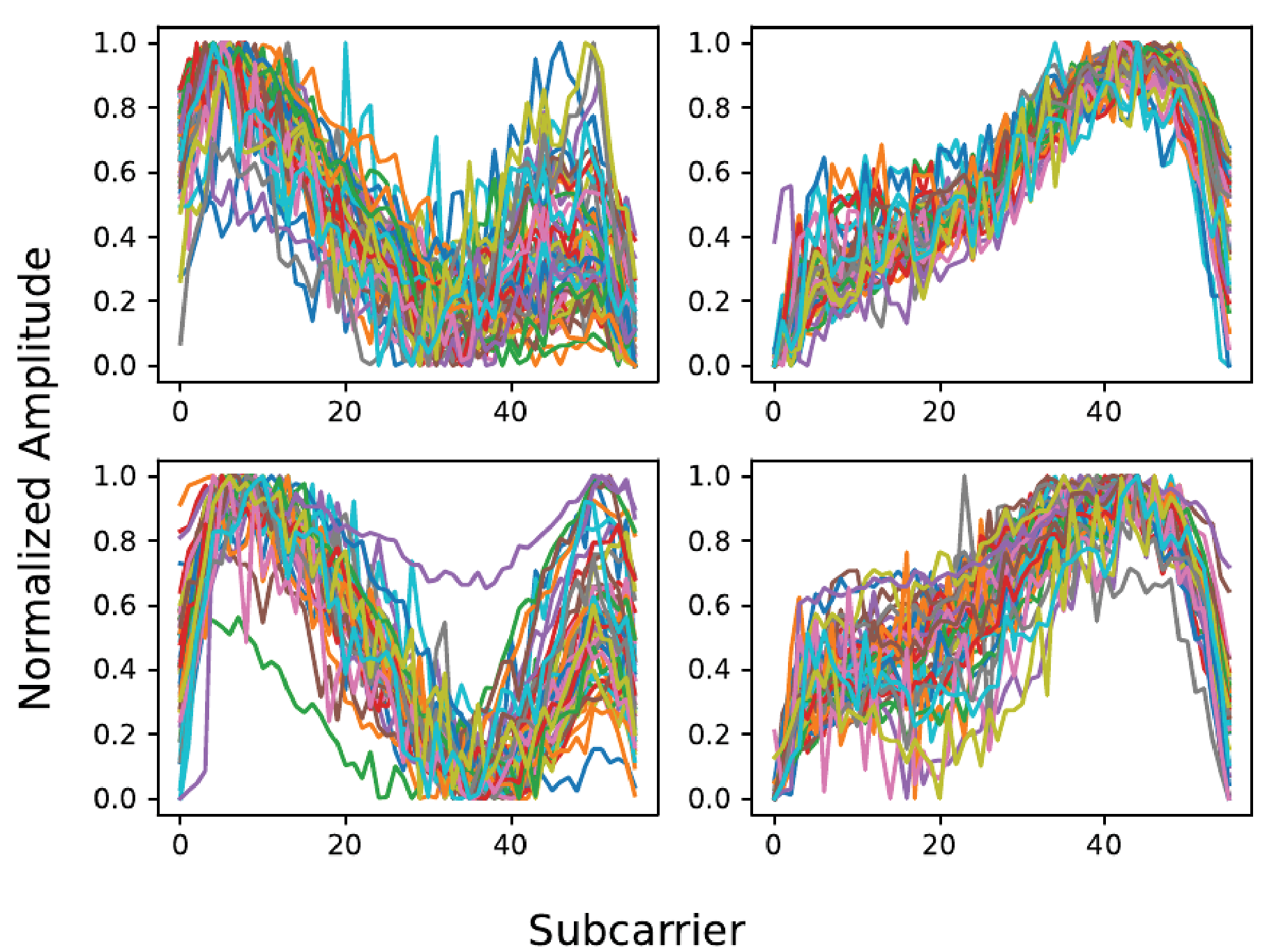}
         \caption{}
         \label{fig:case1}
     \end{subfigure}
     
     \begin{subfigure}{0.24\textwidth}
         \centering
         \includegraphics[width=1\linewidth]{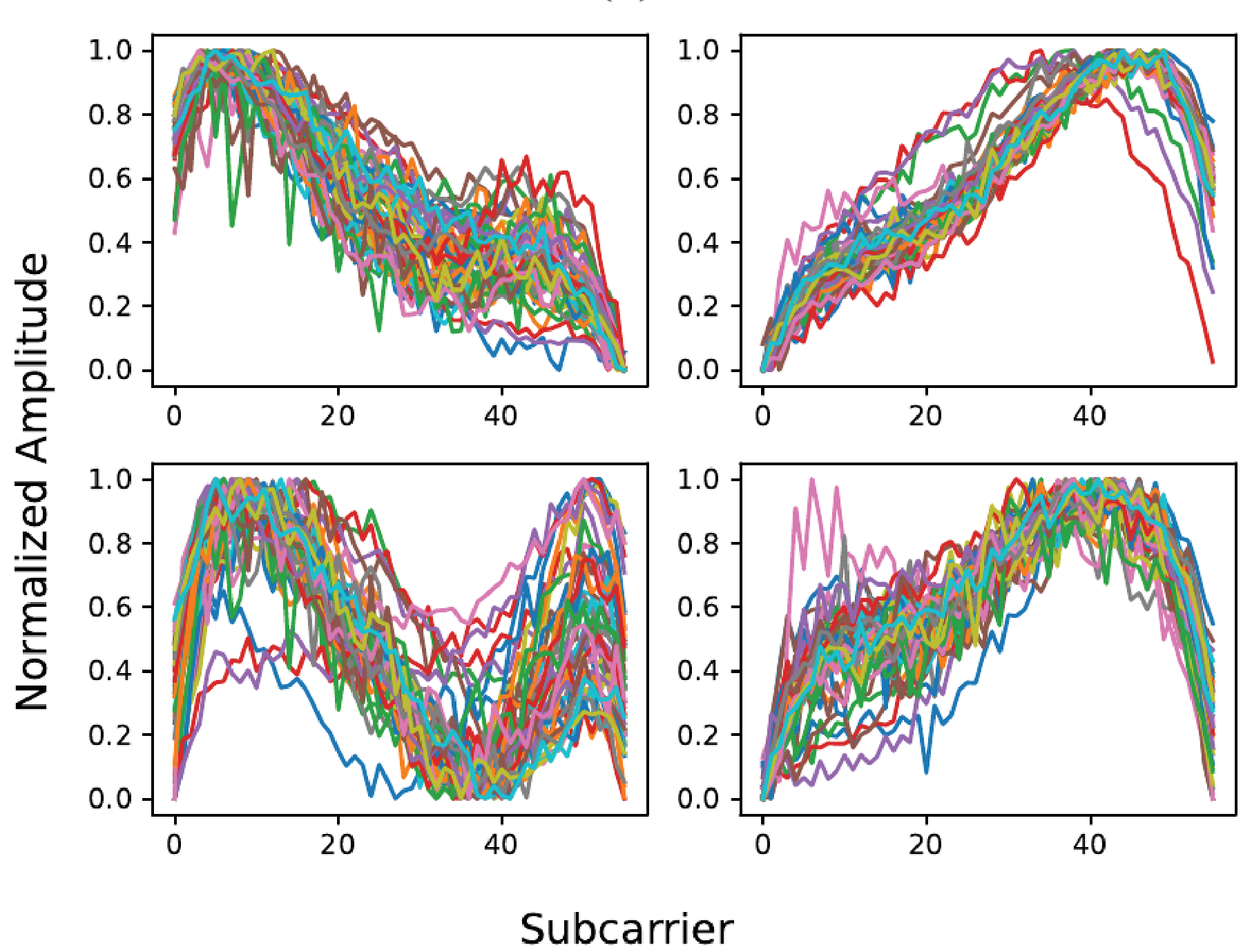}
         \caption{}
         \label{fig:case2}
     \end{subfigure}
     \begin{subfigure}{0.24\textwidth}
         \centering
         \includegraphics[width=1\linewidth]{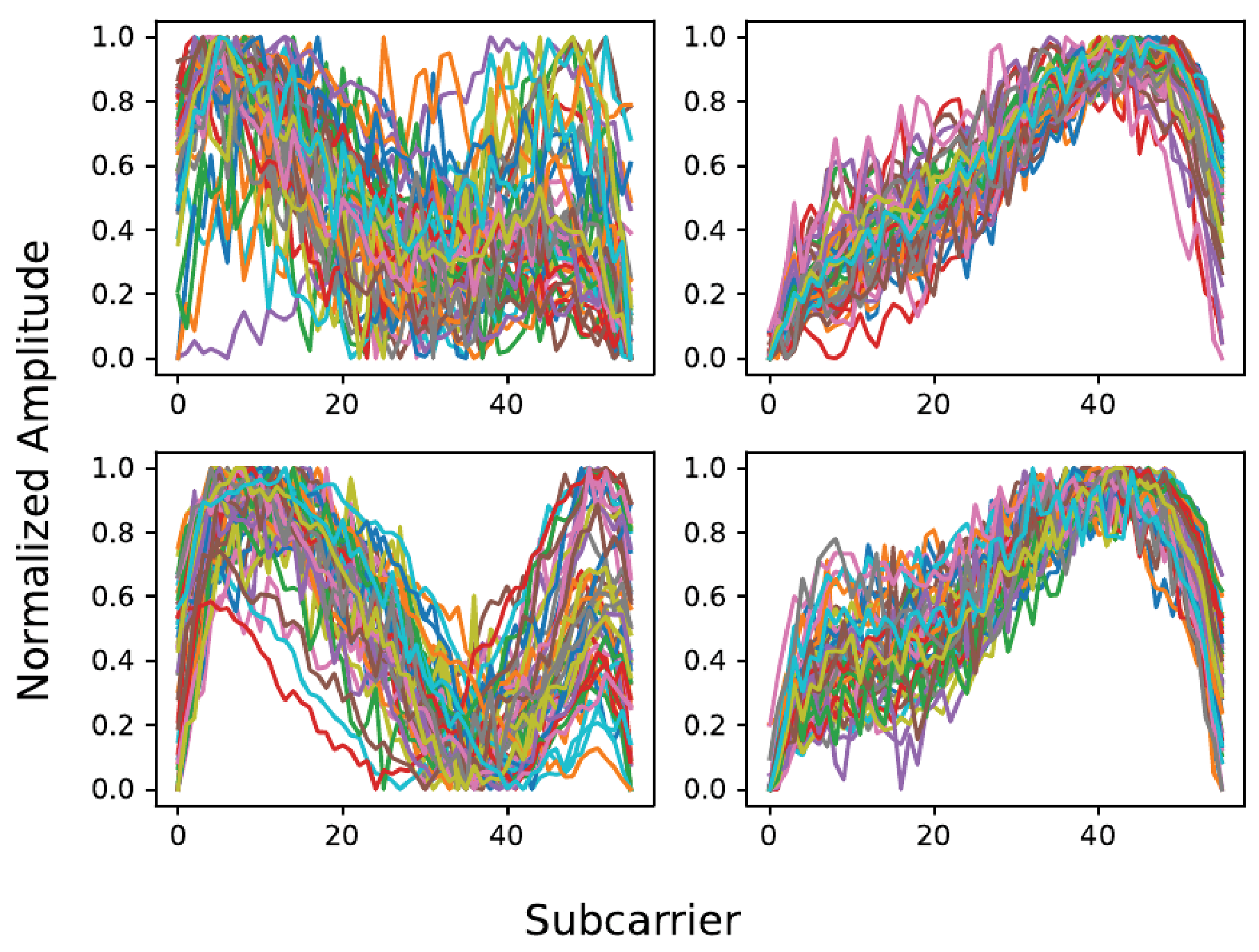}
         \caption{}
         \label{fig:case3}
     \end{subfigure}
        \caption{\small CSI observation of four cases of human presence. (a) Case 1: Empty in both rooms (b) Case 2: A person in room A and empty in room B (c) Case 3: Empty in room A and a person in room B (d) Case 4: people in both rooms. Note that we consider two TX and two RX antennas generating $K=4$ transmission pairs, labeled as $k\in\{1,2,3,4\}$ for respective subplots at top-left, top-right, bottom-left and bottom-right. The x-axis indicates the index of $S=52$ subcarrier, whilst y-axis is value of CSI amplitude.}
        \label{fig:all_cases}
\end{figure}

In an adjoining room scenario, the amplitude response of normalized CSI is fluctuating and follows different patterns among cases. In \fig \ref{fig:all_cases}, we consider 4 transmission pairs with 56 subcarriers. The amplitude response of normalized CSI when human presence in Figs. \ref{fig:case1} to \ref{fig:case3} exhibits significant fluctuations than that of empty case in \fig \ref{fig:case0}. To elaborate a little further, from Figs. \ref{fig:case1} and \ref{fig:case2} with moderate fluctuations,  we can infer that human present on either side of adjoining room reacts in a similar shape of curves. However, there exist subtle differences between them. Although they are both affected by multipath effects, the human presence in different rooms can cause divergent fading and variation. For example, some large-scale patterns can be distinguishable when considering subcarriers as representative features, e.g., CSI amplitude at pair $k=1$ with subcarriers indexed from 40 in cases 2 and 3 respectively in Figs. \ref{fig:case1} and \ref{fig:case2} reveals different shapes of curves. Based on the multipath fading effect, we can leverage these small-scale variations to establish a relationship between the CSI and spatial features.

\begin{figure}
     \centering
     \begin{subfigure}{0.24\textwidth}
         \centering
         \includegraphics[width=1\textwidth]{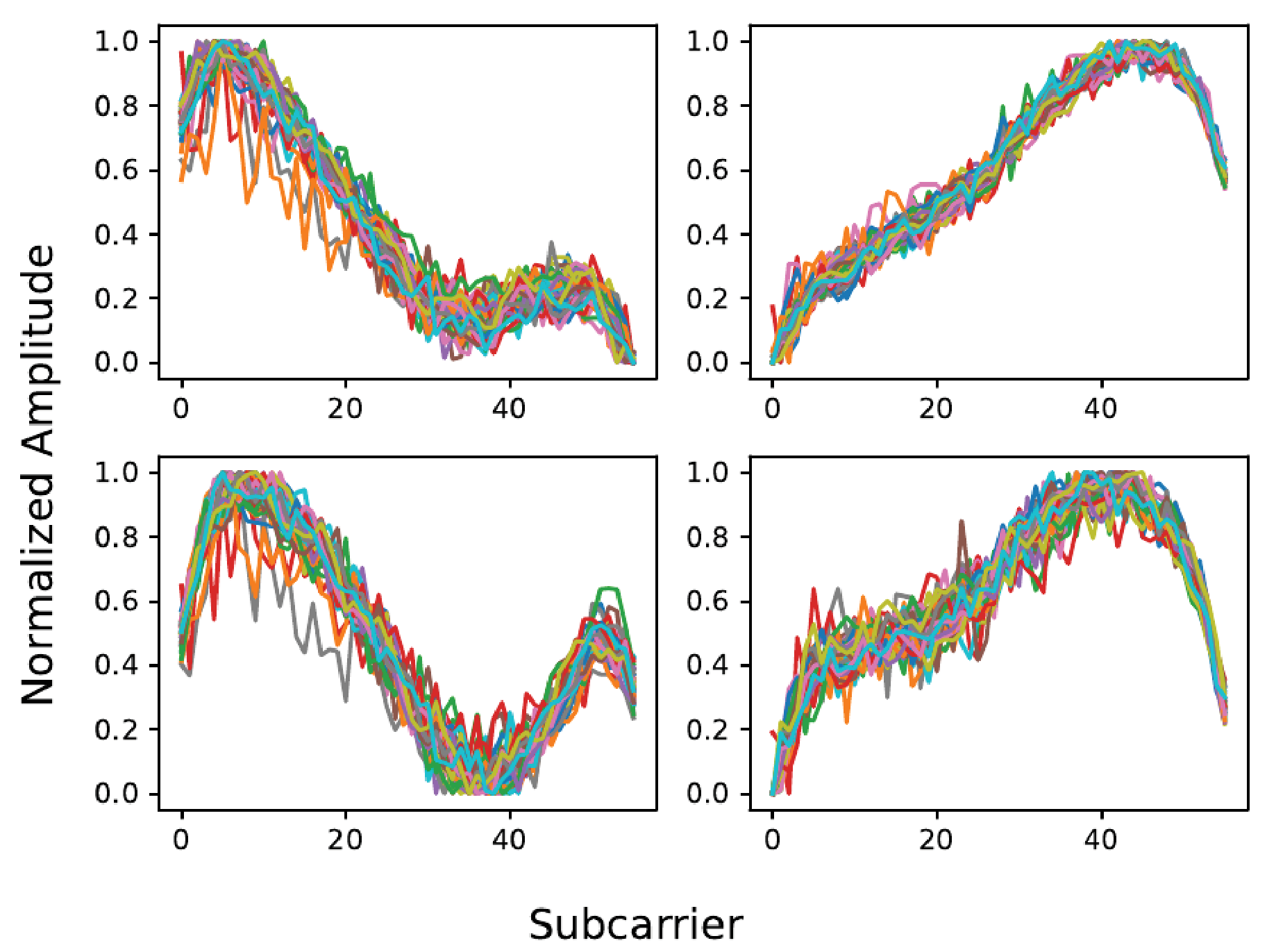}
         \caption{}
         \label{fig:round1}
     \end{subfigure}
     \begin{subfigure}{0.24\textwidth}
         \centering
         \includegraphics[width=1\textwidth]{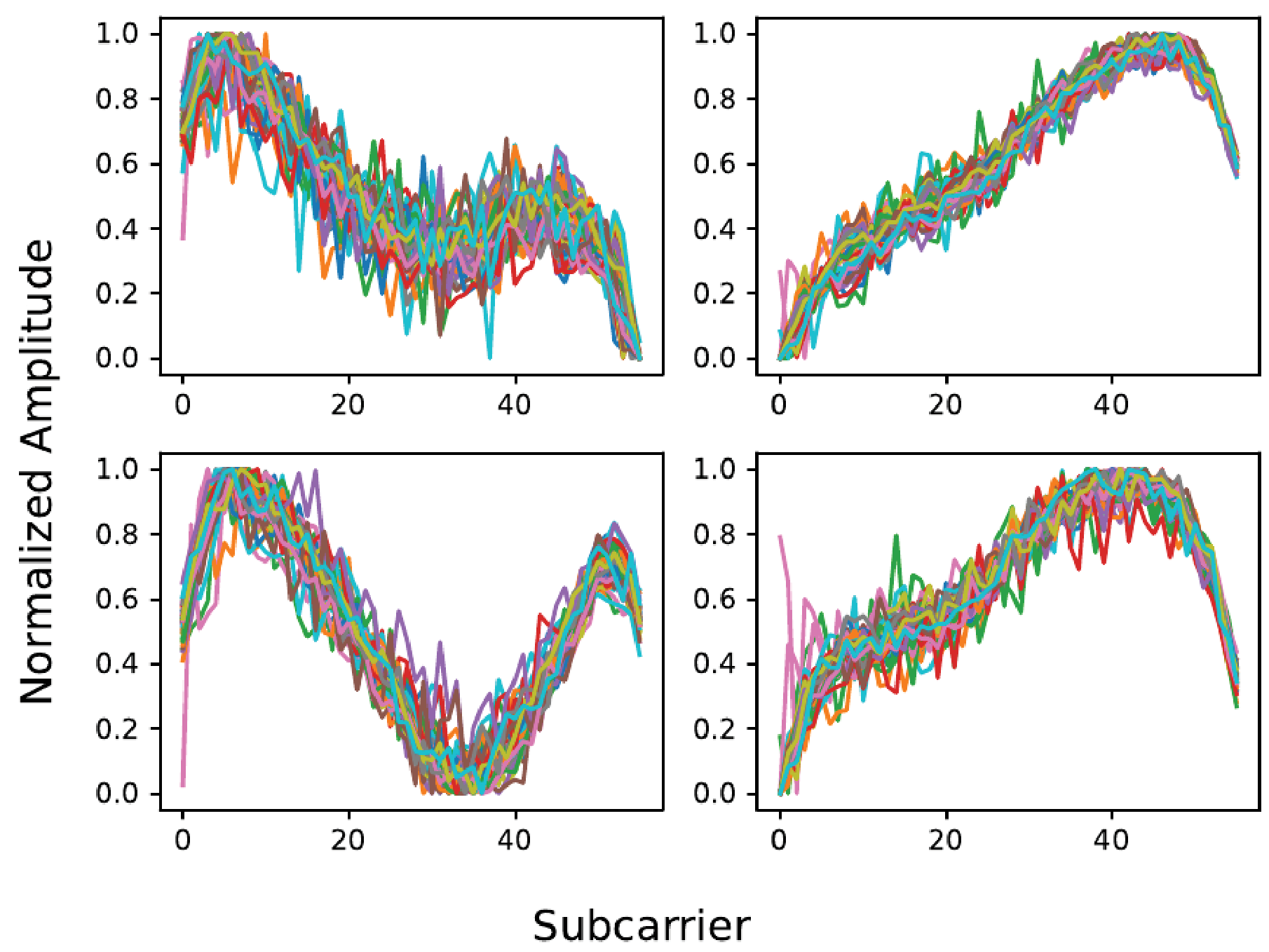}
         \caption{}
         \label{fig:round2}
     \end{subfigure}
     
     \begin{subfigure}{0.24\textwidth}
         \centering
         \includegraphics[width=1\textwidth]{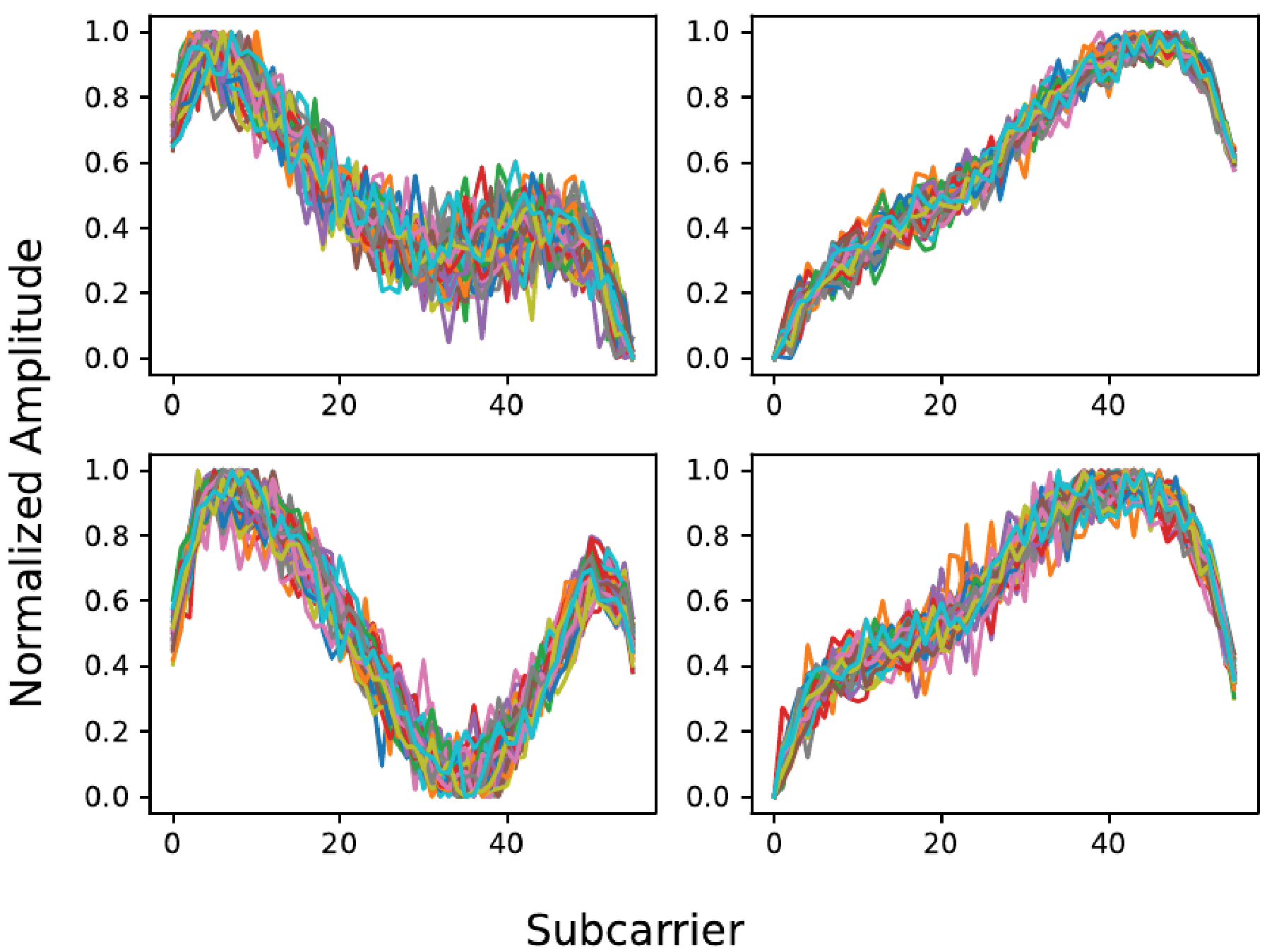}
         \caption{}
         \label{fig:round3}
     \end{subfigure}
     \begin{subfigure}{0.24\textwidth}
         \centering
         \includegraphics[width=1\textwidth]{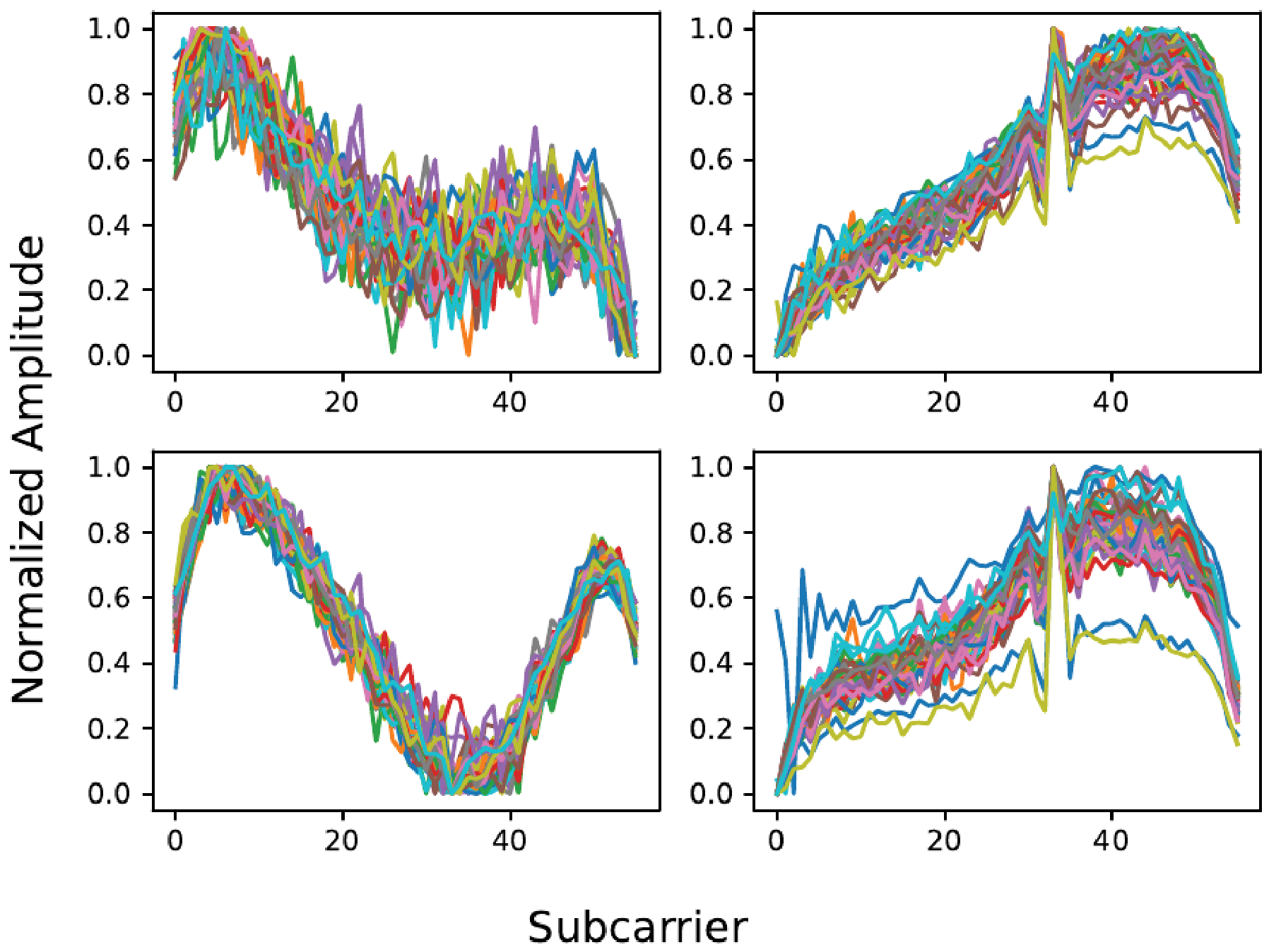}
         \caption{}
         \label{fig:round4}
     \end{subfigure}
     
     \begin{subfigure}{0.24\textwidth}
         \centering
         \includegraphics[width=1\textwidth]{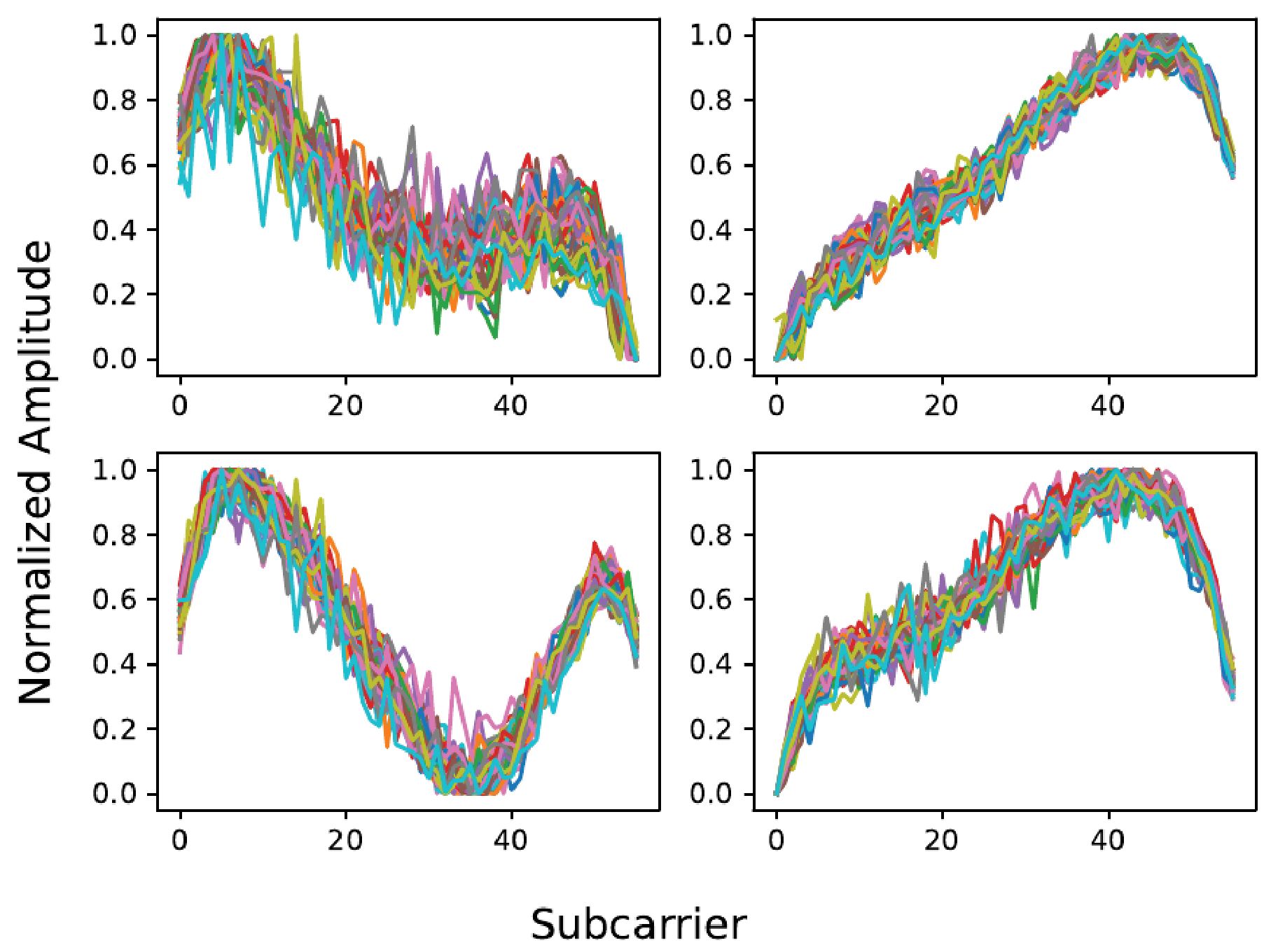}
         \caption{}
         \label{fig:round5}
     \end{subfigure}
     \begin{subfigure}{0.24\textwidth}
         \centering
         \includegraphics[width=1\textwidth]{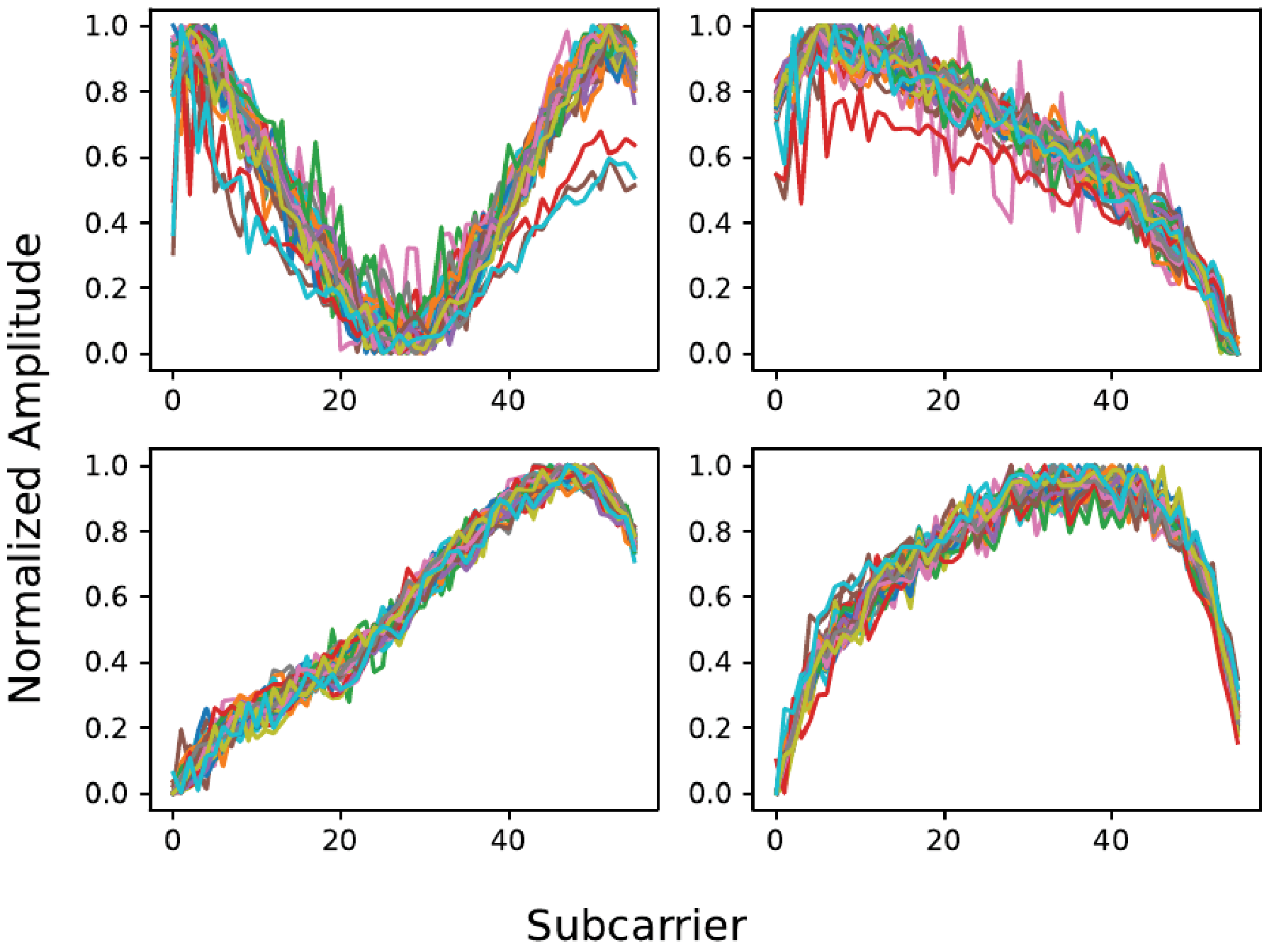}
         \caption{}
         \label{fig:round6}
     \end{subfigure}
     
        \caption{\small Amplitude response of normalized CSI in six different timestamps: (a) Round 1, (b) Round 2, (c) Round 3, (d) Round 4 with heavy rain, (e) Round 5, and (f) Round 6 with different antenna orientation. We consider Case 1 of empty room as an example. Note that we consider two TX and two RX antennas generating $K=4$ transmission pairs, labeled as $k\in\{1,2,3,4\}$ for respective subplots at top-left, top-right, bottom-left and bottom-right. The x-axis indicates the index of $S=52$ subcarrier, whilst y-axis is value of CSI amplitude.}
        \label{fig:all_rounds}
\end{figure}

\subsubsection{Time-Varying Effect}
\label{Observation2}

We observe the amplitude response of normalized CSI when the time-varying effect takes place in Fig. \ref{fig:all_rounds}. We consider Case 1 of empty room. These datasets are collected over six rounds at different times with details described in Table \ref{rounddetail}. Note that the collection interval between each round is an hour. There was heavy rain in the outdoor environment during the fourth round, which can be viewed as an example when the time-varying effect occurred. Additionally, during the sixth round, we manually change the angle of the equipped antennas of APs to emulate the scenario where individuals inadvertently touch the AP causing the displacement. Figs. \ref{fig:round1}, \ref{fig:round2}, and \ref{fig:round3} represent the first three rounds of data with a relatively stable scenario, whereas \fig \ref{fig:round4} reveals drifted signals caused by changes in the environment. Moreover, the channel after the environment in the fourth round is displayed in \fig \ref{fig:round5}. Finally, \fig \ref{fig:round6} demonstrates the severe situation where CSI is no longer similar to all previous rounds due to different deployment, changing the spatial feature caused by multipath. However, it is impractical to label all potential situations in the indoor environment over different timestamps, which is time-consuming and laborious. Therefore, we propose a BTS scheme to flexibly and dynamically learn for presence detection, which is demonstrated in the following section.

\section{Proposed Bifold Teacher-Student (BTS) Learning for Presence Detection} \label{BTS}

\begin{figure*}[!t]
\centering
\includegraphics[width=4.5in]{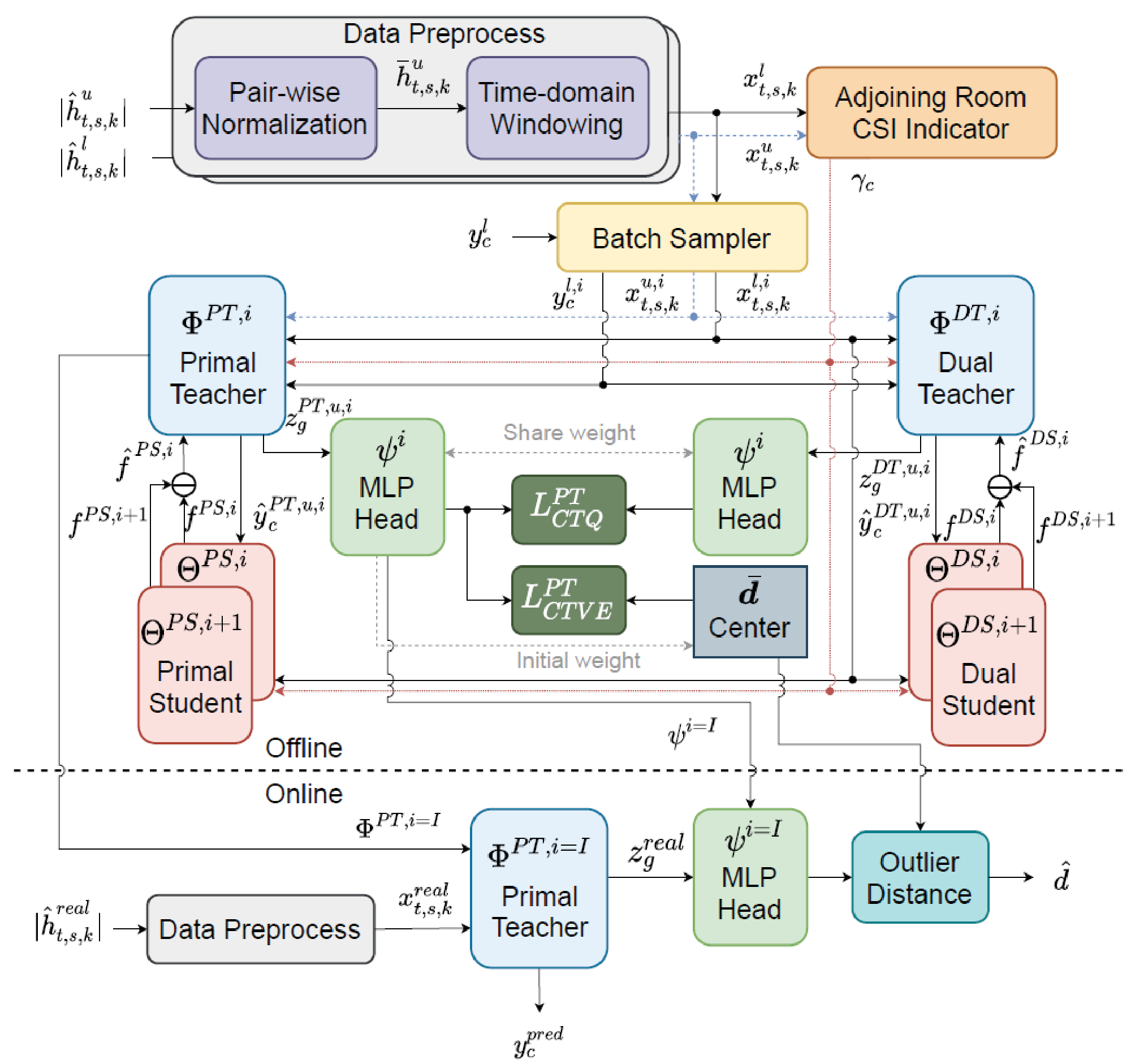}
\caption{\small Schematic diagram of the proposed BTS system.}
\label{fig:Schematicdiagram}
\end{figure*}

In practical scenarios, the process of labeling CSI data can be laborious and time-consuming. To address this issue, we propose a semi-supervised learning approach that leverages both labeled and unlabeled data to learn common characteristics and reduce the need for labeled data. Furthermore, in Section \ref{SYS_PRE}, we have observed the impact of human presence on CSI in an adjoining room scenario, as well as the influence of time-varying effects caused by external factors. To achieve accurate human presence detection under these conditions, we propose a deep learning-based algorithm that exploits these physical phenomena. Fig. \ref{fig:Schematicdiagram} shows the block diagram of the proposed BTS system, which uses two teacher-student (TS) networks, each trained using a similar approach to MPL \cite{MPL} but with different feedback mechanisms. In the training process, labeled and unlabeled CSI data are provided to the teacher network, while only unlabeled data is given to the student network for neural network model updating. The teacher network guides the student network by providing pseudo-labels to the unlabeled data as it learns from the labeled data.

\subsection{Data Preprocessing}

As depicted in \fig \ref{fig:Schematicdiagram}, the data preprocess function block manages the preprocessing pipeline, which comprises of pair-wise normalization and time-domain windowing. The pair-wise normalization normalizes raw CSI data based on the subcarriers in each antenna pair to enhance the spatial feature. The time-domain windowing, on the other hand, conducts sliding window mechanism with a fixed window size to store varying features over a period of time.

\subsubsection{Pair-wise Normalization}

Due to the non-line-of-sight (NLoS) propagation paths from TX to RX, CSI experiences severe multipath effects. Therefore, pair-wise normalization is required as it provides informative spatial feature for each antenna pair. The normalization is initialized at first step in data preprocessing, which takes the CSI amplitude response $|\hat{h}_{t,s,k}|$ as input. We define $q \in Q=\{l, u\}$, where labeled data is denoted as $l$ and unlabeled data is indicated by $u$. Then, we adopt the normalization scheme in $\eqref{eq:normalization}$ as $\bar{h}^{q}_{t,s,k}=norm(\hat{h}^{q}_{t, s, k})$. Therefore, the normalized CSI data based on this equation, including both labeled and unlabeled data, provides more spatial information about human presence and the environment in each antenna pair than the raw data, which are illustrated in \fig \ref{fig:raw_norm_CSI}.

\subsubsection{Time-domain Windowing}

After conducting pair-wise normalization, we utilize time-domain windowing to capture temporal variations in CSI. Specifically, we treat the CSI data at $t^{th}$ packet as a frame with a window size of $\tau$, where each frame stores the varying features in each subcarrier within the windowing period. It is capable of capturing temporal changes in the environment and human presence. The windowing process for the amplitude response of normalized CSI can be formulated as
\begin{align} \label{htox}
x^{q}_{t,s,k}=\left[\bar{h}^{q}_{t-\tau,s,k}, \bar{h}^{q}_{t-\tau+1,s,k}, \ldots, \bar{h}^{q}_{t,s,k}\right],
\end{align}
where the labeled and unlabeled training data will be in the form of $x^{l}_{t,s,k}$ and $x^{u}_{t,s,k}$, respectively. By having these normalized frames, we can leverage different types of training strategies to learn both spatial and time-domain features.

\subsubsection{Adjoining Room CSI Indicator}\label{ARCI}

 	According to CSI observations in Section \ref{SYS_PRE}, we have found that each round of CSI dataset has specific common characteristics that remain stable despite time-varying effects. Discovering these common features is essential as there may be a large difference between labeled and unlabeled data due to time-varying effects, making it difficult to adopt SSL-based methods for training. We define the \textit{disarray} term $\rho$ as
\begin{align}
	\rho^{q}&=\prod_{k=1}^K\left(\frac{1}{S} \sum_{s=1}^S \tilde{w}_{s, k}^q\right), \label{indicator1} 
\end{align}
where
\begin{align}	
	\tilde{w}_{s, k}^q &=-\sum_{t=1}^\tau \frac{x_{t, s, k}^q}{\sum_{t^{'}=1}^{\tau} x_{t^{'}, s, k}^q} \log \left(\frac{x_{t, s, k}^q}{\sum_{t^{''}=1}^{\tau} x_{t^{''}, s, k}^q}\right) \nonumber\\
& \qquad-\frac{\alpha}{\tau}\sum_{t=1}^\tau \left|x_{t, s, k}^q-\frac{1}{\tau} \sum_{t^{'}=1}^{\tau} x_{t^{'}, s, k}^q\right|^\beta. \label{indicator2}
\end{align}
The parameter of $\alpha$ indicates fine-step and $\beta$ denotes fine-order aiming for weighting and rescaling the distance between the values of sampled data and the average data. In $\eqref{indicator1}$, $\rho^{q}$ represents the level of data variance within a packet frame. We calculate the average of $\tilde{w}^{q}_{s,k}$ over subcarriers to obtain a representative value of each antenna pair $k$, and then multiply the values of each pair from $k=1$ to $k=K$ to amplify the influence of each pair. Furthermore, in $\eqref{indicator2}$, $\tilde{w}^{q}_{s,k}$ is composed of subcarrier-wise entropy over time and the average \textit{discrepancy} from the data point $x_{t, s, k}^q$ over a time window. By measuring the subcarrier-wise entropy, we can analyze the instability of CSI frame. It is anticipated that when computing the subcarrier-wise entropy in time, it will have a larger range of values for empty case, and vice versa for people present in rooms. This is because in the empty room, the variation is comparatively smaller, provoking asymptotic trends to that of the averaged value. By contrast, in human presence case, CSI fluctuates over time with larger variance in each subcarrier resulting in smaller entropy.

After calculating the disarray of the labeled data $\rho^{l}, \forall l=\{1,2...,M\}$ and unlabeled data $\rho^{u}, \forall u=\{1,2...,N\}$, we proceed to compute the average disarray. Note that $M$ and $N$ are total data amount of labeled and unlabeled data, respectively. We then subtract the average disarray of the unlabeled dataset from the average disarray of the labeled dataset to obtain the \textit{disparity} as
\begin{align}
	\delta=\frac{1}{M} \sum_{l=1}^M \rho^{l}-\frac{1}{N} \sum_{u=1}^N \rho^{u},
\end{align}
which reflects the differences between the labeled and unlabeled datasets. Note that $\delta$ has smaller impact on the empty room case. Therefore, we can obtain the \textit{indicator} $\gamma_{c}$ as prior knowledge in each case $c\in\{1,2,3,4\}$ as
\begin{align}\label{indicator}
	\gamma_c = \left\{\begin{array}{l}
\frac{1}{M_c} \sum\limits_{l=1}^{M_c} \rho^{l}\left( c \right), \qquad\qquad\quad \mbox{ if } c=1, \\
	\frac{1}{M-M_{c'}} \sum\limits_{l=1}^{M-M_{c'}} \rho^{l}\left( c \right)+\delta,  \, \mbox{ if } c'=1, c\in\{2,3,4\},
\end{array} \right.
\end{align}
where $\rho^{l}(c)$ and $M_c$ represent the CSI data and the total amount of data for case $c$, respectively. Note that $\sum_{c=1}^C M_c = M$. Since the empty room case involves human absence affecting CSI less, $\delta$ is not calculated for this case. After obtaining the indicator $\gamma_{c}$, we use them as reference values for batch data during the training phase.

\begin{figure*}[!th]
     \centering
     \begin{subfigure}[b]{0.45\textwidth}
         \centering
         \includegraphics[width=1\textwidth]{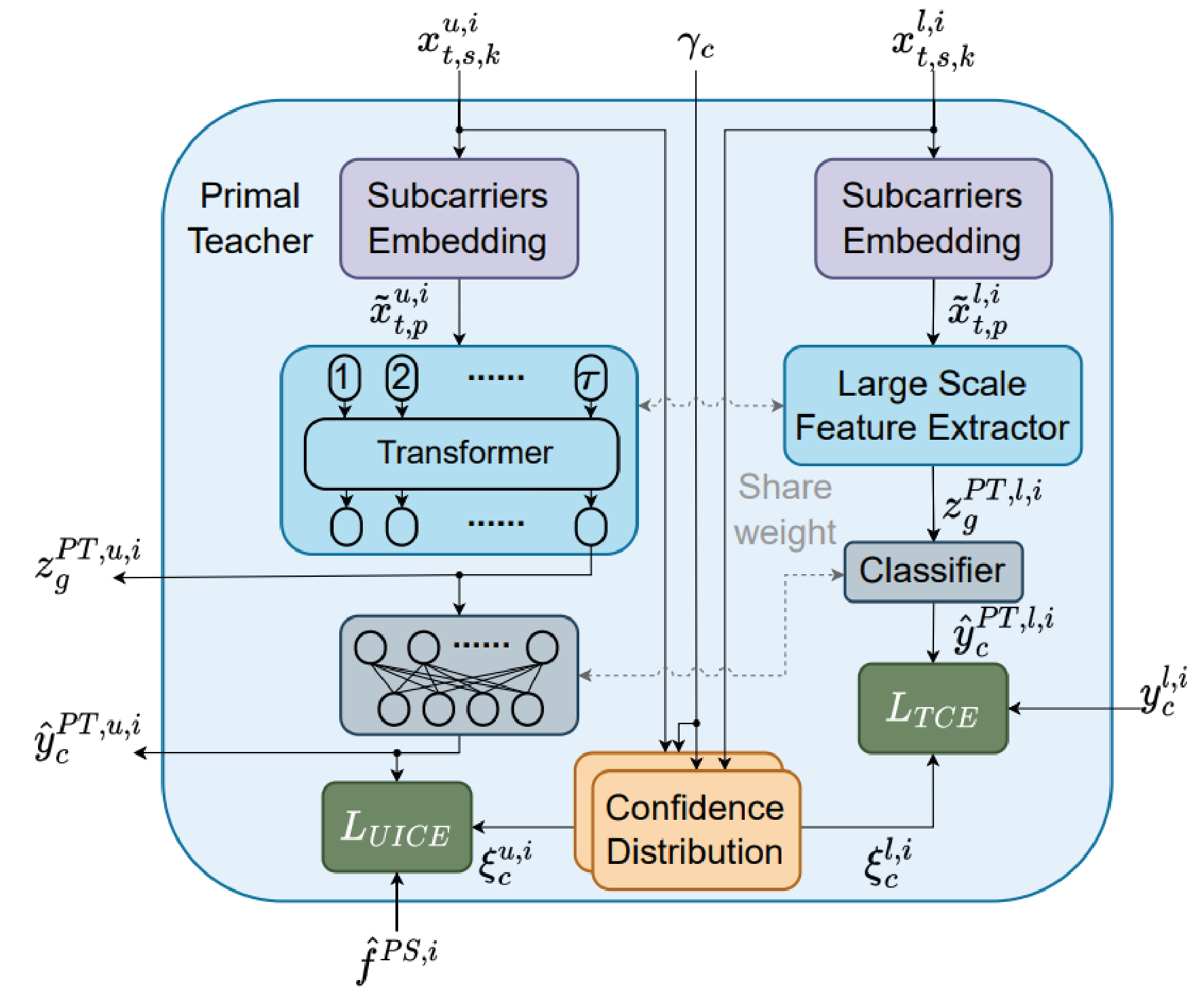}
         \caption{}
         \label{fig:P_T}
     \end{subfigure}
     \quad
     \begin{subfigure}[b]{0.35\textwidth}
         \centering
         \includegraphics[width=1\textwidth]{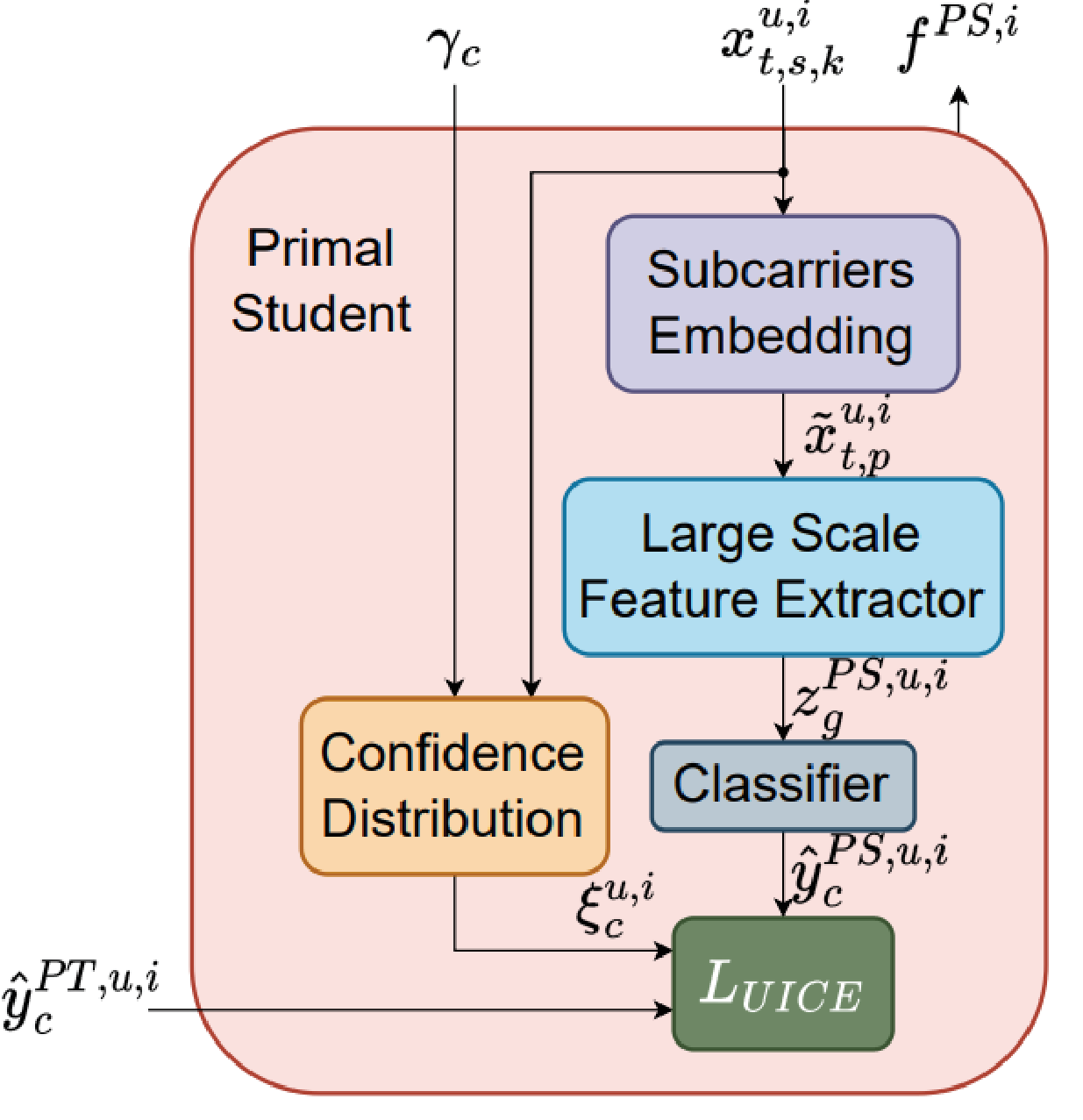}
         \caption{}
         \label{fig:P_S}
     \end{subfigure}
        \caption{\small Architecture of primal teacher and student: (a) Primal teacher network and (b) primal student network.}
        \label{fig:P_TS}
\end{figure*}

\begin{figure*}[!th]
     \centering
     \begin{subfigure}[b]{0.45\textwidth}
         \centering
         \includegraphics[width=1\textwidth]{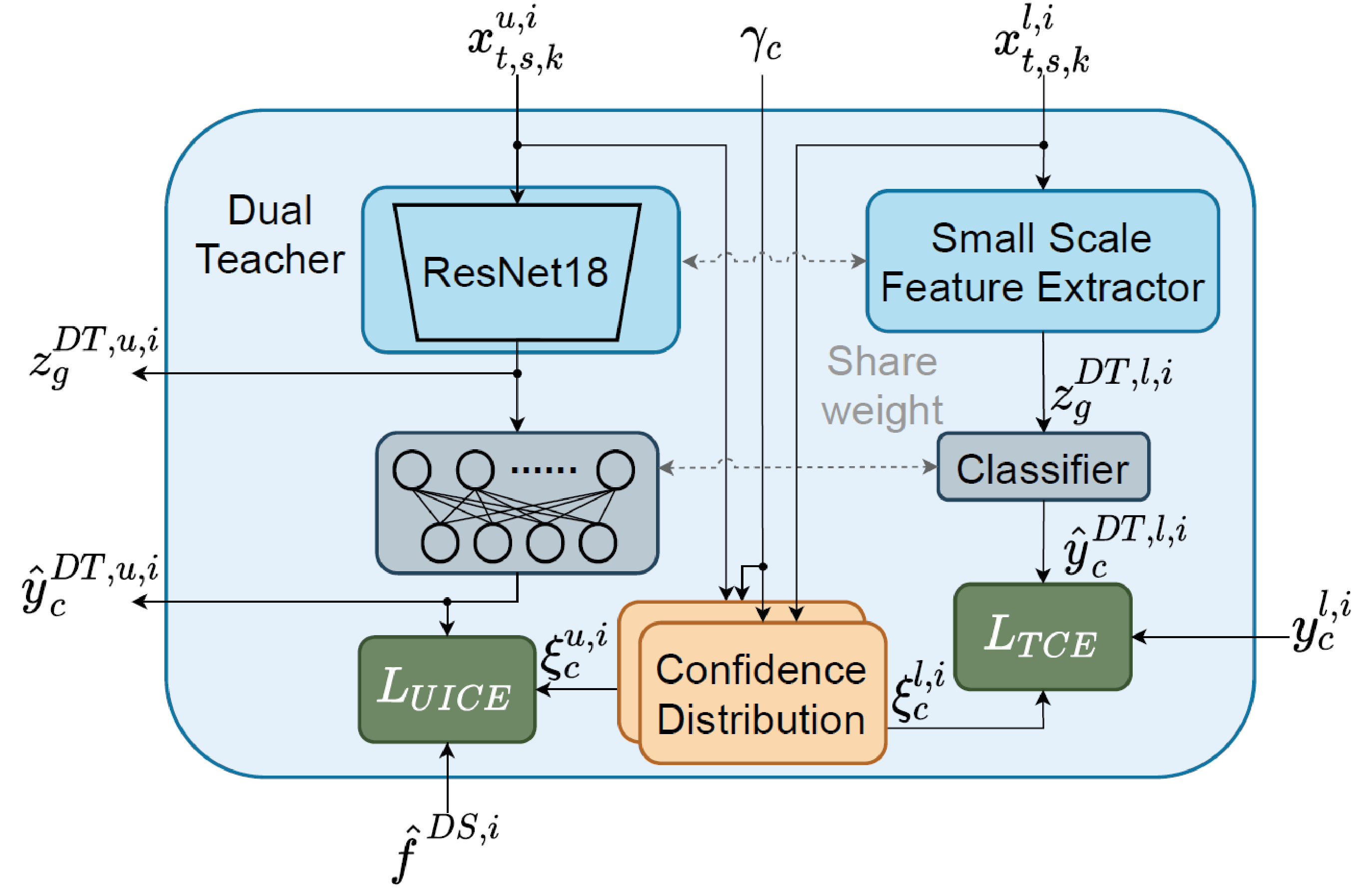}
         \caption{}
         \label{fig:D_T}
     \end{subfigure}
	\quad
     \begin{subfigure}[b]{0.35\textwidth}
         \centering
         \includegraphics[width=1\textwidth]{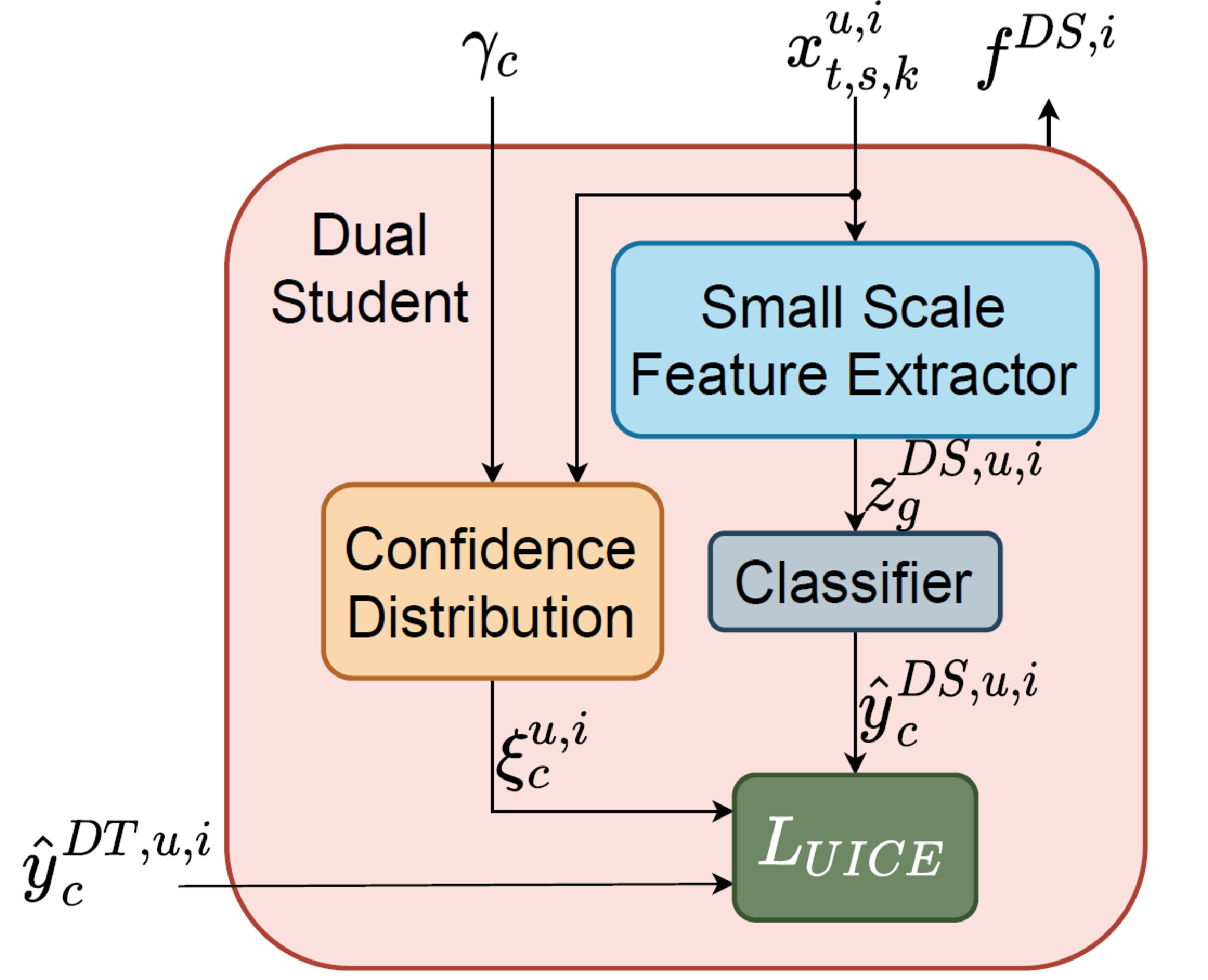}
         \caption{}
         \label{fig:D_S}
     \end{subfigure}
        \caption{\small Architecture of dual teacher and student: (a) Dual teacher network and (b) dual student network.}
        \label{fig:D_TS}
\end{figure*}

\subsection{Deep Neural Network of Primal/Dual Teacher-Students}

In this subsection, we introduce the architecture of the primal TS network as well as the dual TS network. We utilize different network architectures in primal and dual networks to learn different features from the CSI data.

\subsubsection{Primal Teacher-Student}

\fig \ref{fig:P_T} depicts the structure of the primal teacher network $\Phi^{PT,i}$ at iteration $i$, which serves as the final classifier for human presence prediction. First, the primal teacher utilizes the labeled data batch $x^{l,i}_{t,s,k}$ and unlabeled data batch $x^{u,i}_{t,s,k}$ along with the indicator $\gamma_c$ to conduct training. In the subcarriers embedding function block, the concatenated data is defined as $x^{q,i}_{t,p}$, which conducts reshaping in the subcarrier dimension of the original CSI from 3D with a dimension of $\tau \times S \times K$ to 2D with that of $\tau \times P$, where $p\in \{1,...,P\}$ and $P={S \times K}$. Additionally, we consider additional sinusoidal diversity information to each time $t$ of $x^{q,i}_{t,p}$ to diversify the features, which can be obtained as
\begin{align}
\tilde{x}^{q,i}_{t, p} =
\begin{cases}
{x}^{q,i}_{t, p}+\sin\left(\frac{t}{\eta^{\frac{p-2}{P}}}\right), & \text{if}  \text{ mod}(p,2)=0, \\
{x}^{q,i}_{t, p}+\cos\left(\frac{t}{\eta^{\frac{p-1}{P}}}\right), & \text{otherwise},
\end{cases}
\end{align}
where $\eta$ is a diversity constant, whilst $\text{mod}(\cdot, \cdot)$ is the modulo operation. After completing the diversity information, we input $\tilde{x}^{q,i}_{t, p}$ into transformer network. We aim to extract the large-scale correlation among the time sequences through the transformer encoder due to variation in time of all the subcarriers within a frame. We notice that we consider the last feature vector output $z_{g}^{PT,q,i}$ for classification since we pay more attention to the latest timestamp than the expired time sequences.

 Furthermore, a function block of confidence distribution can be calculated based on $\eqref{indicator1}$ and $\eqref{indicator}$ as
\begin{align}\label{CD}
	\xi^{q, i}_{c}=-\left|\rho^{q, i}-\gamma_c\right|^2,
\end{align}
which measures the euclidean distance between the disarray of batch data $x_{t, s, k}^{q, i}$ at iteration $i$. If the disarray of a batch data $\rho_{t}^{q, i}$ is closer to its corresponding indicator $\gamma_c$, the data is more likely to belong to that case and will be given higher confidence. Conversely, if it is far away from its indicator, the data will be given lower confidence. Note that we normalize $\xi^{q,i}_{c}$ to a range of $[1,-1]$ as the final confidence distribution. Finally, we calculate the loss functions $L_{UICE}$ and $L_{TCE}$, which will be elaborated in the latter subsection. Moreover, as shown in \fig \ref{fig:P_S}, the primal student network considers unlabeled data $ x^{u,i}_{t,s,k}$ and pseudo labels $\hat{y}^{PT,u,i}_c$ provided by the primal teacher for training. The labeled data is only adopted to evaluate the two pseudo labels from the primal student networks of ${\Theta}^{PS,i}$ and ${\Theta}^{PS,i+1}$ at different iterations, which is not involved in the network training. Note that the primal student network only calculate loss of $L_{UICE}$ during training. Moreover, as shown in Fig. \ref{fig:Schematicdiagram}, the resulting $f^{PS,i}$ and $f^{PS,i+1}$ generated from consecutive iterations will be fed back to the primal teacher's network. The values of $f^{PS,i}$ and $f^{PS,i+1}$ will be explained and obtained in the latter subsection.

\subsubsection{Dual Teacher-Student}
In \fig \ref{fig:D_TS}, the dual TS network of $\Phi^{DT,i}$ at iteration $i$ is presented, which is designed to assist the training of the primal TS network. The dual teacher network shown in \fig \ref{fig:D_T} includes a confidence distribution, small-scale feature extractor and classifier. Similar to the primal teacher, the dual teacher network also considers labeled and unlabeled batches of data with indicators for training. Alternatively, ResNet\textsuperscript{\ref{note2}}\footnotetext[2]{Compared to other visual feature extractors, such as VGG16 or VGG19, ResNet provides a deeper neural network architecture as well as residual mechanism benefited from batch normalization. The VGG16 model was based on the ImageNet dataset which includes several object categories. Therefore, the existence of a domain gap between ImageNet and the CSI dataset employed in this study may severely occur. \label{note2}} is adopted as a simpler feature extractor to capture small-scale temporal and spectral information within each antenna pair. After extracting features of $z_{g}^{DT,u,i}$ and $z_{g}^{DT,l,i}$ through ResNet, we perform prediction outputs of $\hat{y}^{DT,u,i}$ and $\hat{y}^{DT,l,i}$ via the classifier. To calculate the loss functions $L_{UICE}$ and $L_{TCE}$, we use the confidence distribution variables $\xi^{u, i}_{c}$ and $\xi^{l, i}_{c}$ obtained by $\eqref{CD}$ along with the model predictions and ground truth labeled data $y^{l,i}_{c}$. To elaborate a little further, the dual student network in Fig. \ref{fig:D_S} is similar to the dual teacher network. The dual student network only utilizes the unlabeled data with the pseudo label $\hat{y}^{DT,u,i}$ from dual teacher during training. The labeled data is considered as an evaluation of the two pseudo labels output from the dual student networks of ${\Theta}^{DS,i}$ and ${\Theta}^{DS,i+1}$ at different iterations, to provide a feedback values $f^{DS,i}$ and $f^{DS,i+1}$ to the dual teacher as show in Fig. \ref{fig:Schematicdiagram}.

\subsection{Loss Functions for Training} \label{Loss_Function}

We have designed several loss functions to be optimized for primal/dual TS networks, including transformative cross entropy (TCE), unlabeled indication cross entropy loss, cross-teacher quadratic loss and CSI time-varying Euclidean loss, which are respectively elaborated below. We firstly define a set of teacher network $a\in\{PT,DT\}$ and that of student network $b\in \{PS,DS\}$, where $P$, $D$, $T$, and $S$ stand for "Primal", "Dual", "Teacher", and "Student", respectively.

\subsubsection{Transformative Cross Entropy Loss}
As shown in Figs. \ref{fig:P_T} and \ref{fig:D_T}, the transformative cross entropy loss is formulated as
\begin{align} \label{TCE}
L_{TCE}^a=-\sum_{c=1}^{C}{Softmax\left(y_c^{l,i}+\xi^{l, i}_{c}\right)\log{\left(\Phi^{a,i}\left(x_{t,s,k}^{l,i}\right)\right)}},
\end{align}
where $y^{l,i}_{c}$ is the ground truth of batch labeled data $x_{t,s,k}^{l,i}$ in case $c$ at $i^{th}$ iteration, $\xi^{l,i}_{c}$ is the confidence distribution of batch labeled data, and $\Phi^{a,i}(x_{t,s,k}^{l,i})$ predicts the batch labeled data in iteration $i$ from teacher network $a$. The design concept is described as follows. We incorporate the ground truth of the labeled batch into its corresponding confidence distribution, converting it into a probability distribution with a $softmax$ function. This process increases the probability of the label with the highest confidence when the distributions of the labeled and time-varying unlabeled datasets become similar. However, when the distribution is highly dissimilar due to severe time-varying effects, the lowest  confidence value will then confuse the ground truth, leading to inaccurate estimation. Note that the summation of $y_c^{l,i}$ and $\xi^{l, i}_{c}$ aims for improving the similarity between the distributions of the labeled and unlabeled datasets. Moreover, as mentioned previously, the teacher networks employ the labeled dataset to guide the student networks in learning the unlabeled dataset. While, the knowledge of the unlabeled dataset is acquired through the feedback signals from student networks. The confidence distribution is capable of compromising the difference between the labeled and unlabeled datasets by leveraging the prior knowledge of the CSI data.

\subsubsection{Unlabeled Indication Cross Entropy Loss}

Inspired by MPL, we have conceived two types of unlabeled indication cross entropy (UICE) losses that belong to either teacher or student networks as shown in both Figs. \ref{fig:P_TS} and \ref{fig:D_TS}, which are respectively expressed as $L_{UICE}^a$ in $\eqref{UICE_total}$ and $L_{UICE}^b$ in $\eqref{UICE_student}$ as follows. 
\begingroup
\allowdisplaybreaks
\begin{subequations} \label{UICE_total}
\begin{align}\label{UICE}
L_{UICE}^a&=\hat{f}^{b,i}\cdot \left[-\sum_{c=1}^{C}Softmax\left({\hat{y}}_c^{a,u,i}+\xi^{u, i}_{c}\right) \right. \notag \\
&\qquad\qquad\qquad\qquad\qquad
\left. \log{ \left(\Phi^{a,i}\left(x_{t,s,k}^{u,i}\right)\right)} \right],
\\
\hat{f}^{b,i}&=ReLU\left(\sum_{c=1}^{C}y_c^{l,i}\log{\left(\Theta^{b,i+1}\left(x_{t,s,k}^{l,i}\right)\right)} \right. \notag \\
&\qquad\qquad\quad
\left. - \sum_{c=1}^{C}{y_c^{l,i}\log{\left(\Theta^{b,i}\left(x_{t,s,k}^{l,i}\right)\right)}}\right), \label{UICE2}
\end{align}
\end{subequations}
\begin{align} \label{UICE_student}
L_{UICE}^b=-\sum_{c=1}^{C}{Softmax\left({\hat{y}}_c^{a,u,i}+\xi^{u, i}_{c}\right)\log{\left(\mathrm{\Theta}^{b,i}\left(x_{t,s,k}^{u,i}\right)\right)}}.
\end{align}
\endgroup
The loss function $L^{a}_{UICE}$ in $\eqref{UICE}$ is adopted in the two teacher networks with unlabeled dataset and its corresponding pseudo label as well as confidence distribution. The cross entropy loss is multiplied by the feedback signal $\hat{f}^{b,i}$ in $\eqref{UICE2}$. Moreover, since both student networks are trained with the unlabeled dataset and its pseudo labels without $\hat{f}^{b,i}$, we mainly calculate their cross entropy loss, as expressed in $L^{b}_{UICE}$ in $\eqref{UICE_student}$. Note that $\Phi^{a,i}(x_{t,s,k}^{u,i})$, $\Theta^{b,i}(x_{t,s,k}^{l,i})$, and $\Theta^{b,i}(x_{t,s,k}^{u,i})$ represent the neural network output of prediction distribution of pseudo labels from the unlabeled data through teacher networks, labeled as well as unlabeled one through the student networks, respectively. The benefit leveraging confidence distribution is the same as that in $L^{a}_{TCE}$ in $\eqref{TCE}$. To elaborate a little further, a rectified linear unit (ReLU) function is used in $\eqref{UICE2}$ in order to constrain the feedback and stabilize the training of the TS network. A negative value of the feedback indicates that the loss of iteration $i+1$ is lower than that of iteration $i$, suggesting that training on unlabeled data is not necessary for the teacher network. However, it is not desirable for the teacher network to train in the opposite direction with a negative feedback. Accordingly, the ReLU function helps to avoid this issue and ensure the stability of the training process.

\subsubsection{Cross-Teacher Quadratic Loss}
Inspired by contrastive learning \cite{contrastive} and the bootstrap mechanism in \cite{BYOL}, we propose a self-supervised learning loss function that utilizes the same input data for generating different representations in different neural network. Therefore, as shown in Fig. \ref{fig:Schematicdiagram}, a cross-teacher quadratic (CTQ) loss is designed between the primal and dual teachers, which can be expressed as
\begin{align} \label{CTQ}
L_{CTQ}^{PT}=\left \|\psi^{i}\left(z_{g}^{PT,u,i}\right) - \psi^{i}\left(z_{g}^{DT,u,i}\right)\right \| ^2,
\end{align}
where $\psi^{i}(\cdot)$ refers to the multilayer perceptron (MLP), whilst $z_{g}^{PT,u,i}$ and $z_{g}^{DT,u,i}$ denote the latent features obtained after performing transformer of the primal and dual teacher, respectively. Note that the loss function is denoted by $PT$ due to the final human presence prediction in the primal teacher. We consider transformer encoder of primal teacher to discover temporal correlations, whereas ResNet of dual teacher is employed to extract spatial features. Subsequently, the mean square error between the two representation vectors in $\eqref{CTQ}$ is computed to establish communication between the two networks and generate an enhanced representation vector.

\begin{algorithm}[!t]
\caption{Proposed BTS Scheme}
\begin{algorithmic}[1] \label{Alg}

	\STATE \textbf{(Offline Phase)}
	\STATE  Preprocess raw CSI data $|\hat{h}^{l}_{t,s,k}|$ and $| \hat{h}^{u}_{t,s,k}|$ to obtain $x^{l}_{t,s,k}$ and $x^{u}_{t,s,k}$ based on $\eqref{htox}$
	
	\STATE  Generate adjoining room CSI indicators $\gamma_c$ according to $\eqref{indicator} $ as input of primal/dual TS function blocks

	\STATE Initialize iteration $i=0$ and upper bound of iteration $I$

	\STATE Initialize primal/dual TS networks $\Phi^{PT, i}$, $\Phi^{DT, i}$, $\Theta^{PS,i}$ and $\Theta^{DS,i}$, MLP head $\psi^{i}$, as well as center point $\bar{\boldsymbol{d}}$
	
	\WHILE{$i<I$}
	
	\STATE Sample labeled and unlabeled data in batches: $x^{l,i}_{t,s,k}$ and $x^{u,i}_{t,s,k}$ with the corresponding ground truth of labeled data: $y^{l,i}_{c}$

	\STATE Obtain fixed vector for drift detection: $\bar{\boldsymbol{d}}$ based on $\eqref{center}$

	\STATE Calculate representation vectors: $z^{PT,u,i}_{g}$ and $z^{DT,u,i}_{g}$ as input of MLP head $\psi^{i}$

	\STATE Update the pseudo labels of $\hat{y}^{PT,u,i}_{c}$ and $\hat{y}^{DT,u,i}_{c}$ for students

	\STATE Calculate primal/dual student feedbacks as $f^{PS,i}$ and $f^{DS,i}$ based on $\eqref{UICE2}$, which are fedback to their primal/dual teachers

	\STATE Obtain confidence distribution of labeled and unlabeled data respectively as $\xi^{l,i}_c$ and $\xi^{u,i}_c$ based on $\eqref{CD}$

	\STATE Conduct back-propagation for updating primal/dual TS networks $\Phi^{PT,i}$, $\Phi^{DT,i}$, $\Theta^{PS,i}$, and $\Theta^{DS,i}$ and MLP head $\psi^i$ based on $L^{PT}$, $L^{DT}$, $L^{PS}$ and $L^{DS}$
	
	\STATE Next iteration $i\leftarrow i+1$
	
	\ENDWHILE

	\STATE \textbf{(Online Phase)}
	\STATE Process real-time raw CSI data as $|\hat{h}^{real}_{t,s,k}|$ $\rightarrow$ $x^{real}_{t,s,k}$
		
	\STATE Predict human presence based on primal teacher network $\Phi^{PT,i=I}(x^{real}_{t,s,k})$ $\rightarrow$ $y^{pred}_{c}$
	
	\STATE \textbf{(Retraining)}
	\STATE Calculate online outlier distance $\hat{d}$ based on $\eqref{outlier}$ using offline obtained $\bar{\boldsymbol{d}}$	
	\IF{$\hat{d} \geq d_{th}$}
		\STATE Conduct neural network retraining in the offline phase
	\ENDIF

\end{algorithmic}
\end{algorithm}

\subsubsection{CSI Time-Varying Euclidean Loss}
The CSI time-varying Euclidean loss (CTVE) as shown in Fig. \ref{fig:Schematicdiagram} is leveraged to detect whenever data drift and variation occurs, which is attained as
\begin{align} \label{CTVE}
	L_{CTVE}^{PT}=\left \|\psi^{i}\left(z_{g}^{PT,u,i}\right) - \bar{\boldsymbol{d}}\right \|^2,
\end{align}
where the center point of the unlabeled dataset is expressed as
\begin{align} \label{center}
	\bar{\boldsymbol{d}}=\frac{1}{N}\sum_{u=1}^{N}{\psi^{i=0}\left(z_{g}^{PT,u,i}\right)}.
\end{align}
Note that initial weight of $\eqref{center}$ is given by the MLP head coming from primal teacher.

As a result, we can constrain all unlabeled data to a certain range by minimizing the distance between the representation vector and the center point $\bar{\boldsymbol{d}}$. As illustrated at the bottom part of Fig. \ref{fig:Schematicdiagram}, this mechanism allows us to evaluate real-time streaming data in the online phase through the outlier distance as
\begin{align}\label{outlier}
	\hat{d}=\left \|\psi^{i=I}\left(z_{g}^{real}\right)-\bar{\boldsymbol{d}}\right \|^2,
\end{align}
where $z_{g}^{real}$ represents the latent feature of real-time CSI, and $\psi^{i=I}$ indicates model of MLP at the last iteration $I$. When deploying our model in real-time scenarios, we need to consider the time-varying nature of CSI caused by environmental changes. Serious time-varying effects can lead to data drift, resulting in further model drift with degraded performances. Therefore, we continuously monitor the outlier distance of real-time CSI data by comparing it with the outlier distance of the training data. If the distance between them exceeds a certain threshold, i.e., $\hat{d} \geq d_{th}$, we conclude it as data drift and collect new data to retrain the network. This process ensures that our model remains robust and performs well in real-time scenarios.

\subsubsection{Total Loss and Training Process}

In our proposed BTS system, we have designed five loss functions for optimizing the models in primal/dual TS networks, including $L^{a}_{TCE}$ in $\eqref{TCE}$, $L^{a}_{UICE}$ in $\eqref{UICE_total}$ , $L^{b}_{UICE}$ in $\eqref{UICE_student}$, $L^{PT}_{CTQ}$ in $\eqref{CTQ}$, and $L^{PT}_{CTVE}$ in $\eqref{CTVE}$. Therefore, the total loss of primal and dual teacher networks are respectively designed as
\begin{subequations}
\begin{align}\label{totalloss1}
	L^{PT}& \!=\!\lambda_1^{PT} L_{TCE}^{PT} \!+\! \lambda_2^{PT} L_{UICE}^{PT} \!+\!\lambda_3^{PT} L_{CTQ}^{PT}\!+\!\lambda_4^{PT} L_{CTVE}^{PT},
\end{align}
\begin{align}\label{totalloss2}
	L^{DT}=\lambda_1^{DT} L_{TCE}^{DT}+\lambda_2^{DT} L_{UICE}^{DT}+\lambda_3^{DT} L_{CTQ}^{PT},
\end{align}
\end{subequations}
where $\lambda$'s are respective parameters for striking a balance between losses. Furthermore, the total loss of primal and dual student networks can be acquired as $L^{PS}=L_{UICE}^{PS}$ and $L^{DS}=L_{UICE}^{DS}$, respectively. The overall training pipeline can be demonstrated in Algorithm \ref{Alg}, which follows the procedures in Fig. \ref{fig:Schematicdiagram}.

\begin{figure*}[!ht]
     \centering
      \begin{subfigure}{0.37\textwidth}
         \centering
         \includegraphics[width=1\textwidth]{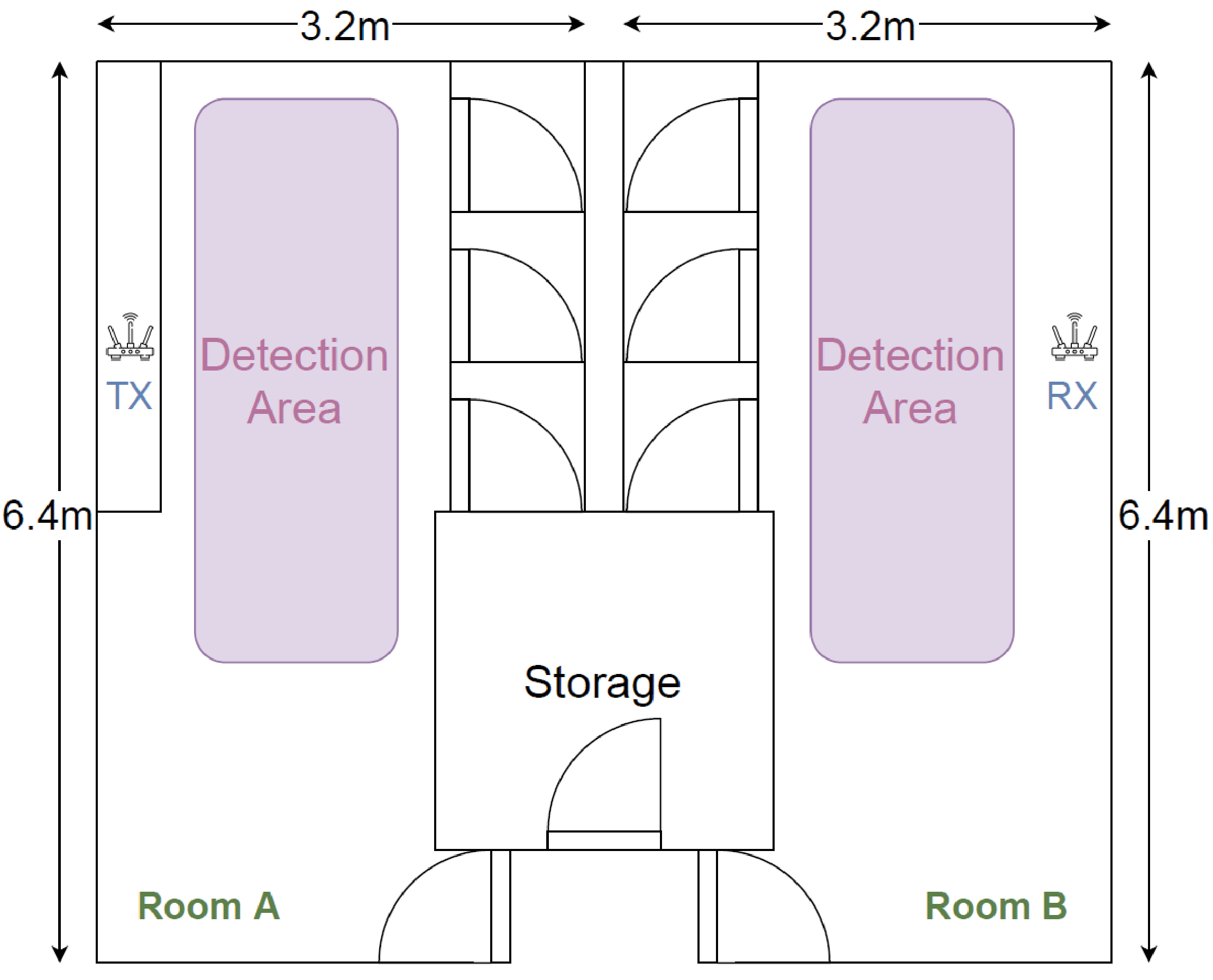}
         \caption{}
         \label{fig:Schematic}
     \end{subfigure}
     \hfill
     \begin{subfigure}{0.3\textwidth}
         \centering
         \includegraphics[width=1\textwidth]{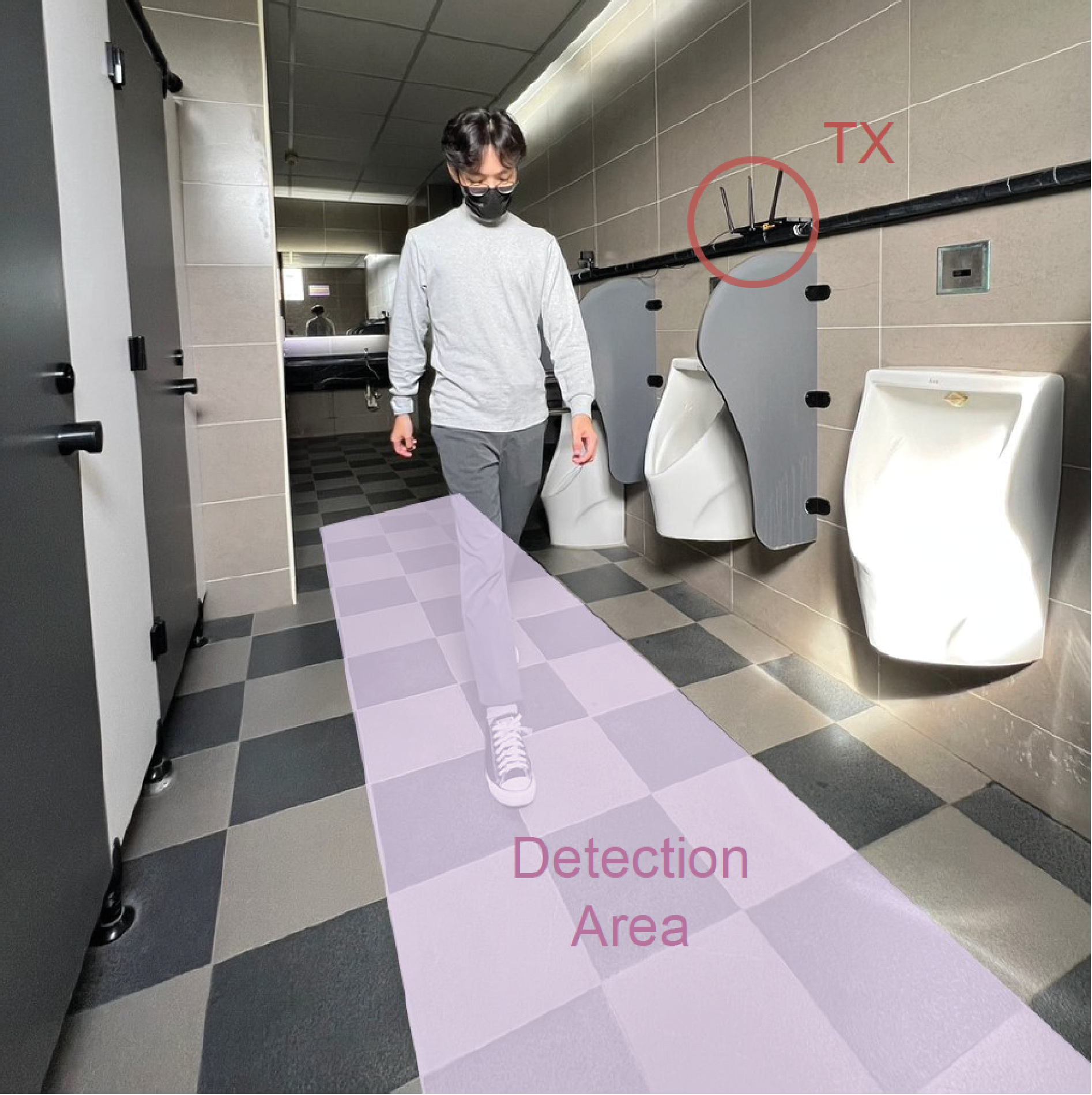}
         \caption{}
         \label{fig:RoomA}
     \end{subfigure}
      \hfill
     \begin{subfigure}{0.3\textwidth}
         \centering
         \includegraphics[width=1\textwidth]{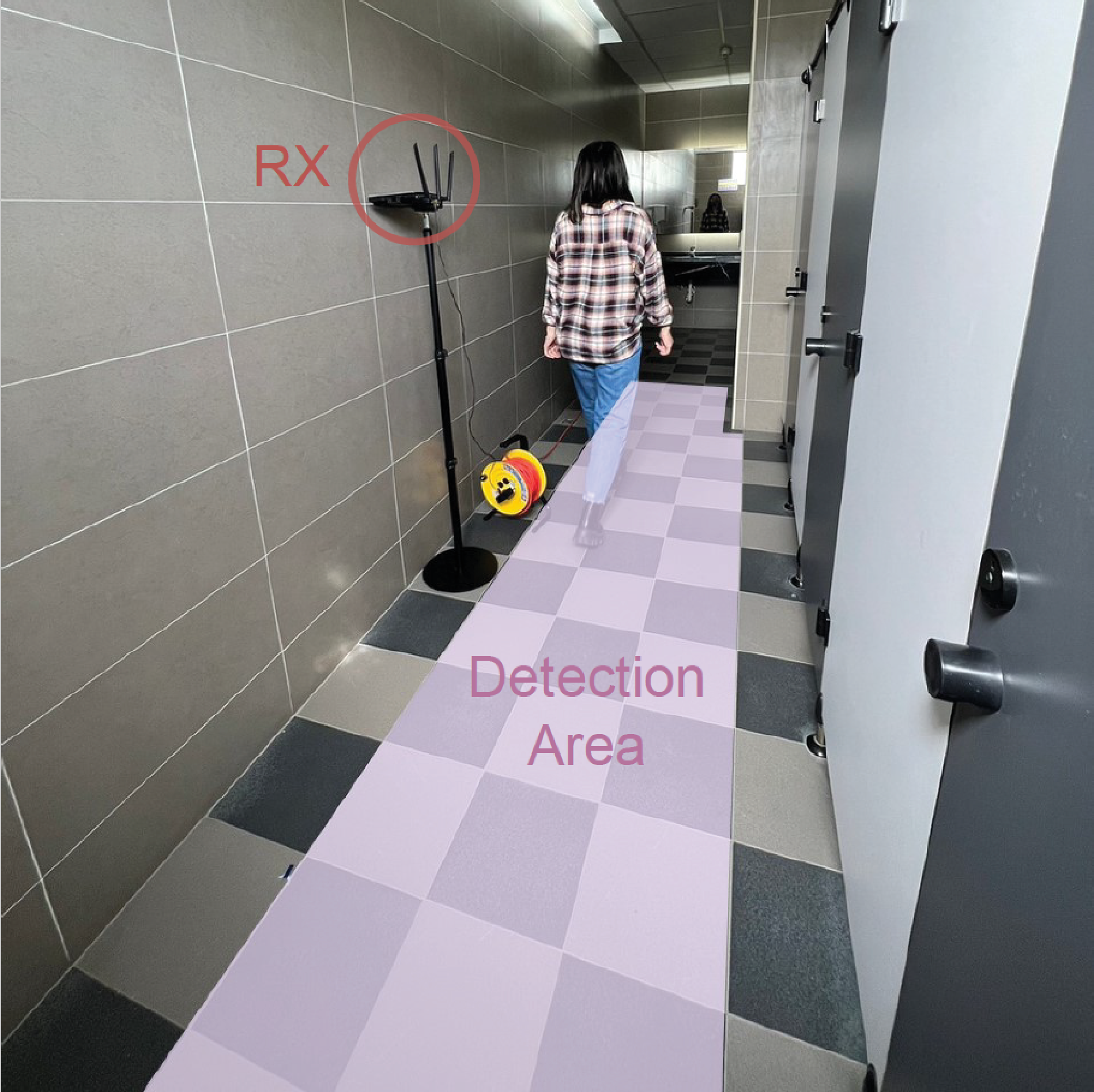}
         \caption{}
         \label{fig:RoomB}
     \end{subfigure}
        \caption{\small Schematic diagram of experimental scenarios for adjoining room human presence detection: (a) Experimental scenario and settings of adjoining room, (b) experimental scenario in room A, and (c) experimental scenario in room B.}
        \label{fig:experiment1}
\end{figure*}

\section{Performance Evaluation} \label{EXP}
\subsection{Experimental Settings}

Our experiments are conducted in an adjoining room setup, as illustrated in Fig. \ref{fig:Schematic}. The configuration consists of two neighboring restrooms, specifically room A in Fig. \ref{fig:RoomA} and room B in Fig. \ref{fig:RoomB}, with each of a dimension of $3.2\times 6.4\ m^2$. Note that a $1.5 \times 1.5\ m^2$ storage space with walls are separating the two restrooms. We deploy two TP-Link TL-WDR4300 wireless routers, which support IEEE 802.11n Wi-Fi protocol providing CSI information based on \cite{csitool}. The routers are configured in AP mode, with TX located in room A and RX in room B. The APs transmit at a center frequency of 2.447 GHz with a bandwidth of 20 MHz. The RX is operated under a dual-band mode, which is additionally connected to an edge server at a center frequency of 5.3 GHz. The edge server performing training and prediction receives 10 estimated CSI packets per second from the RX. As previously mentioned, data collection takes place for six rounds, with an one-hour interval between each round. We also record the environmental conditions during collection. The system parameters are presented in Table \ref{systemPar}, while the details of each round can be found in Table \ref{rounddetail}. We collect the data with a size of 2000 for each case in each round. Note that round 4 is conducted in a rainy weather, while round 6 is conducted with a different antenna orientation. The information of the involved participants is listed as follows: 3 women with heights (cm)/weights (kg) as 158/50, 155/46, 162/55, and 7 men with heights (cm)/weights (kg) as 174/67, 178/62, 171/62, 184/73, 185/71, 176/70, 172/56. The first two women and two men participate the two-room scenarios, whereas the remaining people involve the three-/four-room cases in Subsection \ref{newsub}. Note that all participants provide around a data volume of 500 for each case in all rounds. At most one person can randomly walk in either one of the two detection areas, which is regarded as a more difficult task than that more than two people. This is because more people will fluctuate the signal more significantly, making it more distinguishable than that with a single person in the room \cite{new_cronos}.

\begin{table}[t]
\small
\centering
\caption{Parameter Setting}
\begin{tabular}{ll}
\hline
Parameters of system                      & Value              \\ \hline
Number of rooms                           & 2                  \\
Number of cases                           & 4                  \\
Number of APs per room                     & 1                  \\
Number of antennas in TX                  & 2                  \\
Number of antennas in RX                  & 2                  \\
Number of antenna pairs                   & 4                  \\
Number of subcarriers per antenna pair    & 56                 \\
Carrier frequency between TX and RX       & 2.447 GHz          \\
Carrier frequency between RX and edge     & 5.3 GHz \\
Channel bandwidth                         & 20 MHz             \\
CSI data sampling rate                       & 10 packets/sec     \\
Number of collection rounds               & 6                  \\\hline
\end{tabular}
\label{systemPar}
\end{table}

\begin{table*}[t]
\small
\centering
\caption{Information of Data Collection Rounds}
\begin{tabular}{lcccl}
\hline
Rounds  & Collection time & Data volume   & Antenna angle & Environments \\ \hline
1       & 12:00-12:20     & 2000/case     & $90^o$        &    Sunny   \\
2       & 13:20-13:40     & 2000/case     & $90^o$        &   Sunny with layout same as round 1   \\
3       & 14:40-15:00     & 2000/case     & $90^o$        &   Sunny with layout same as round 1   \\ 
4       & 16:00-16:20     & 2000/case     & $90^o$        &   Rainy with layout changed  \\ 
5       & 17:20-17:40     & 2000/case     & $90^o$        &   Sunny with layout same as round 4    \\ 
6       & 18:40-19:00     & 2000/case     & $45^o$        &   \tabincell{l}{Sunny with layout same as round 4 \\ but different antenna orientation} \\ \hline
\end{tabular}
\label{rounddetail}
\end{table*}

\subsection{Selection of Different Hyperparameters}
In our system, there are several adjustable parameters that are crucial in stabilizing the training process and achieving feasible performance, including the fine-step $\alpha$ and fine-order $\beta$ in $\eqref{indicator2}$, as well as the value of $\lambda$'s adopted in the total loss functions $\eqref{totalloss1}$ and $\eqref{totalloss2}$.

\subsubsection{Fine-Step $\alpha$ and Fine-Order $\beta$ in Disarray Function}

\begin{figure}[!t]
     \centering
     \begin{subfigure}{0.35\textwidth}
         \centering 
         \includegraphics[width=1\textwidth]{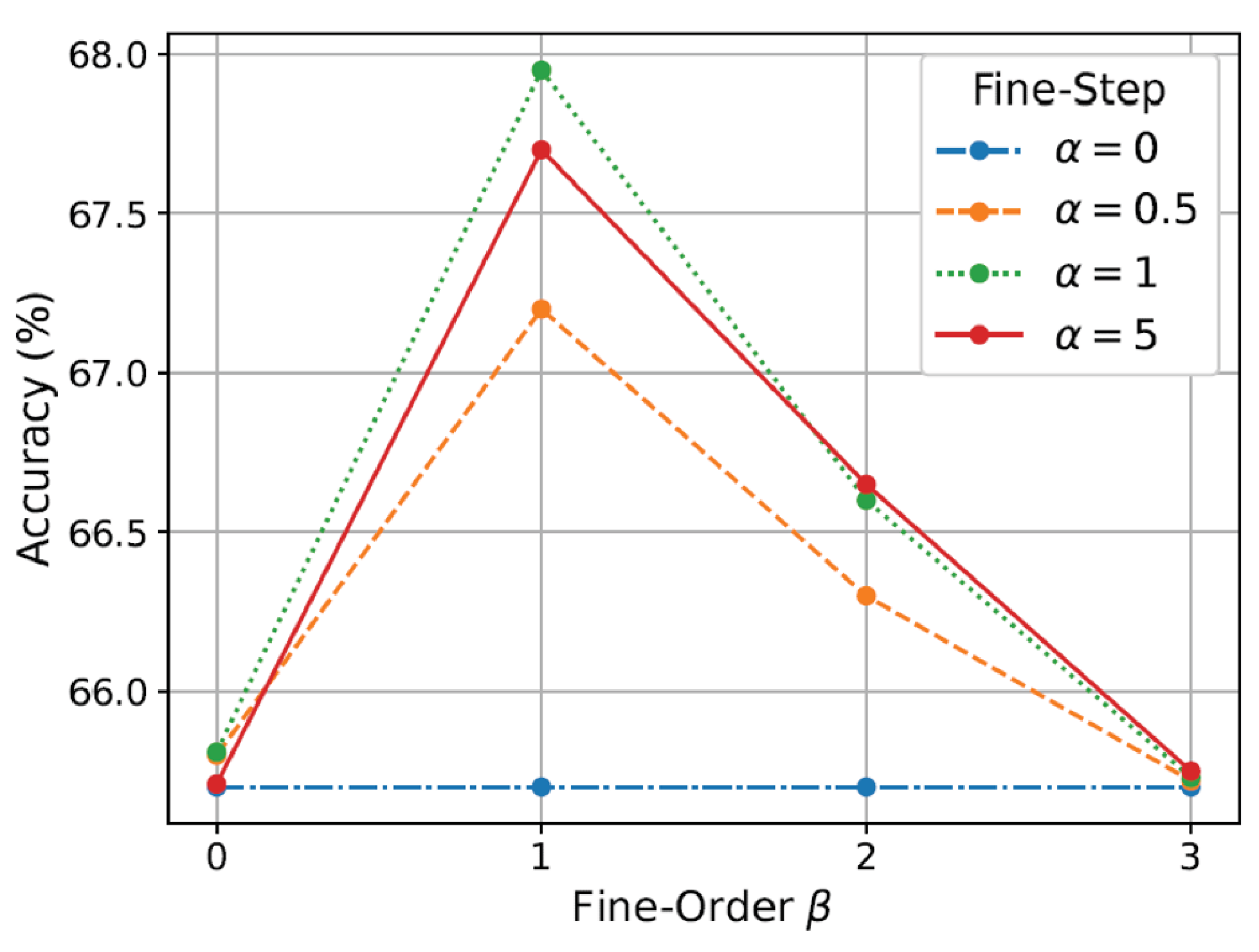}
         \caption{}
         \label{fig:f1_1}
     \end{subfigure}
     \begin{subfigure}{0.35\textwidth}
         \centering
         \includegraphics[width=1\textwidth]{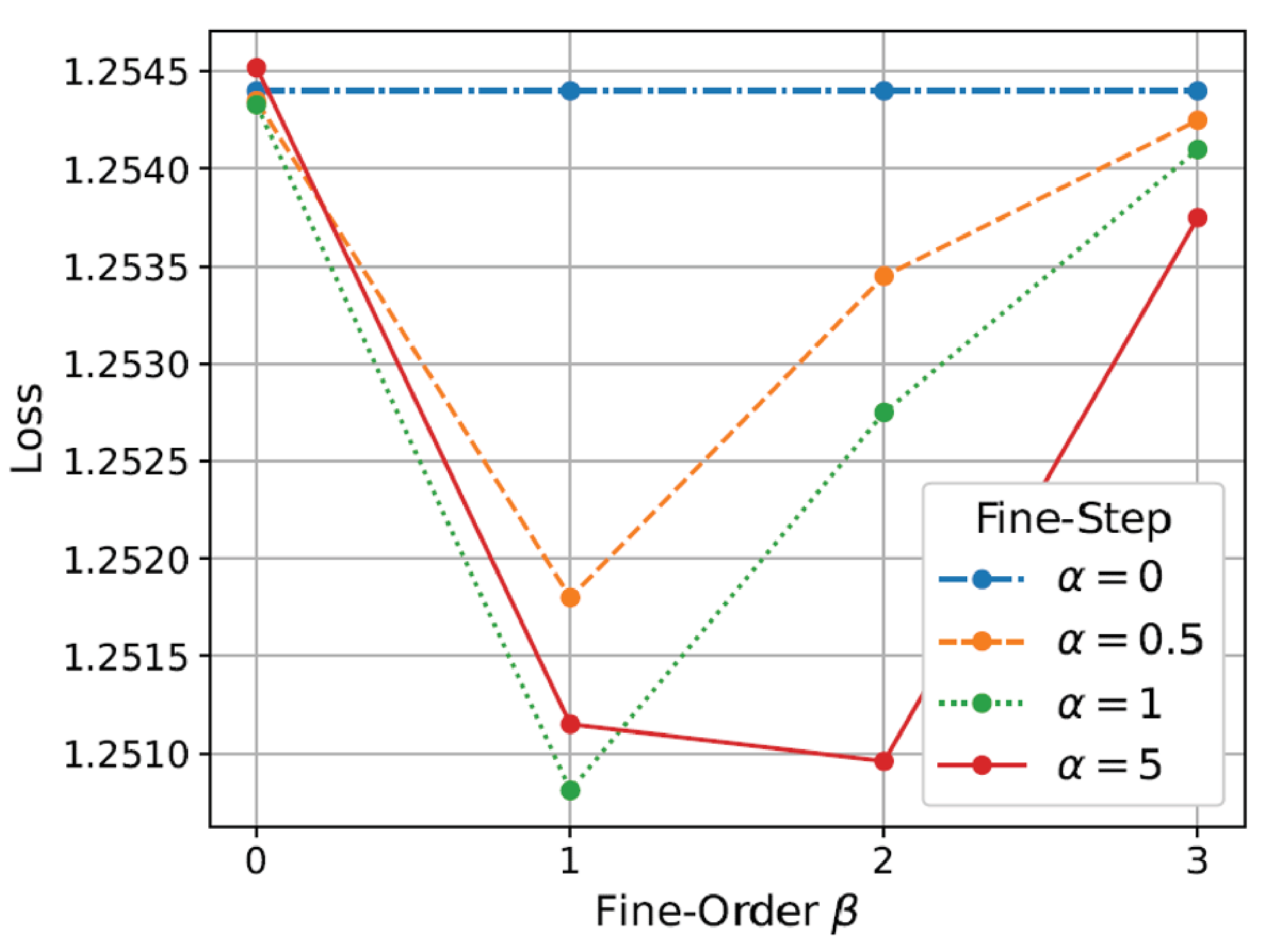}
         \caption{}
         \label{fig:f1_2}
     \end{subfigure}
        \caption{\small Performance of accuracy and loss with different parameters of fine-step $\alpha$ and fine-order $\beta$ in terms of (a) accuracy and (b) loss.}
        \label{fig:f1}
\end{figure}

Note that the adjoining room CSI indicator and confidence distribution that we have designed can be performed as a training-free classifier. Accordingly, we apply this classifier to evaluate parameters of $\alpha$ and $\beta$ as illustrated in \fig \ref{fig:f1}. We consider round 1 as the labeled dataset and round 6 as the unlabeled one. Thereafter, the unlabeled dataset is sampled in a size of 256 for calculating confidence distribution, which aims for emulating the batches of training data. We can observe from \fig \ref{fig:f1_1} that $\alpha=0$ leads to the lowest accuracy of $65\%$ since it considers only subcarrier-wise entropy, which did not take time-varying factors into account. However, larger values of $\alpha$ will lead to over-estimation according to only time-varying effect. By adjusting fine-order, smaller $\beta$ leads to insignificance of second term in $\eqref{indicator2}$, whilst larger one lead to pure cross entropy classification result. The asymptotic result can also be observed in \fig \ref{fig:f1_2}. Although the loss of $\alpha=5$ is lower than that of $\alpha=1$ due to the second term in $\eqref{indicator2}$, it dominates the result inducing a lower accuracy as explained previously. Therefore, we select $\alpha=1$ and $\beta=1$ as the parameters in the following simulations.

\begin{figure}[!t]
     \centering
     \includegraphics[width=0.4\textwidth]{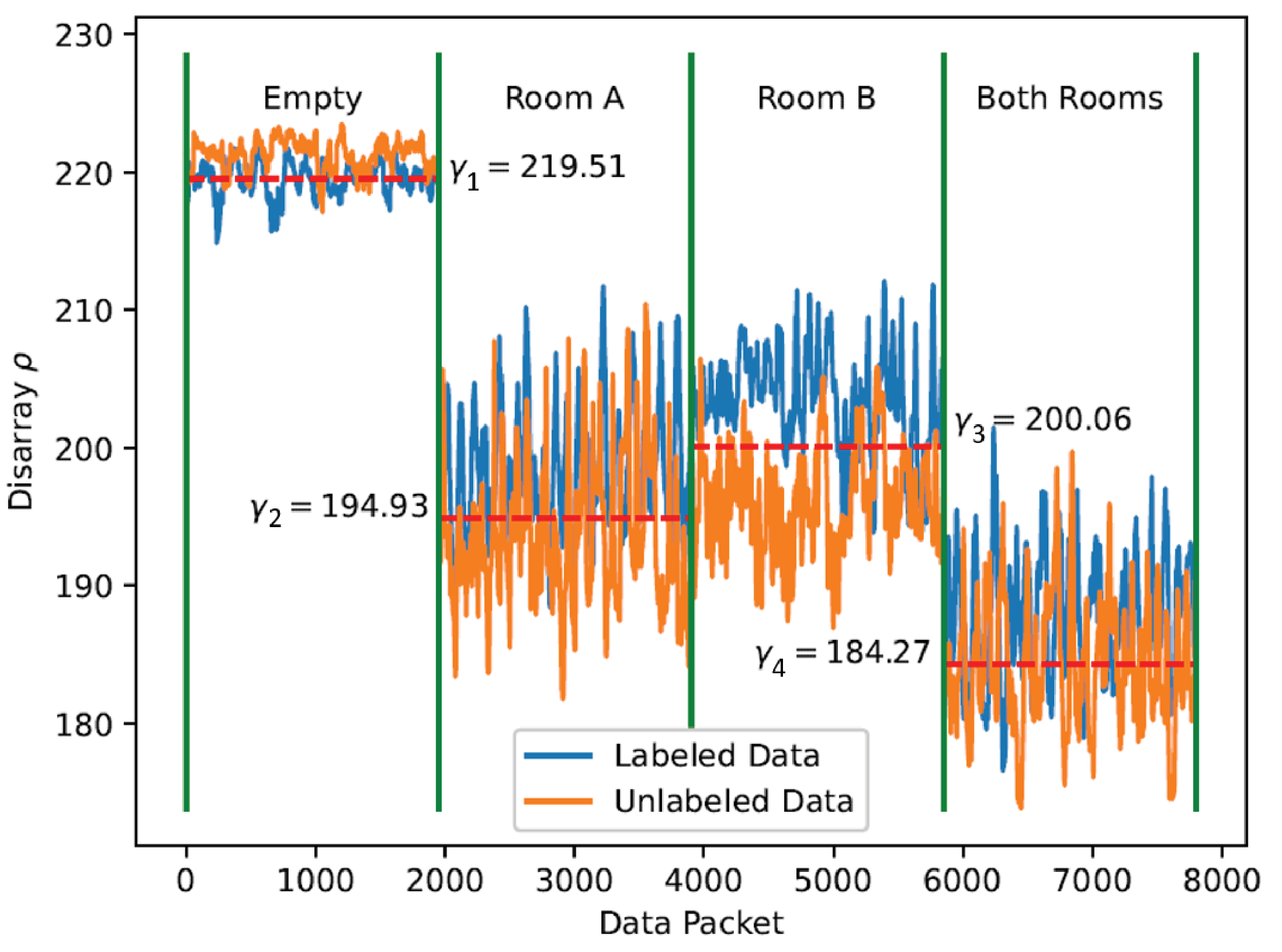}
     \caption{\small Disarray and indicator of the labeled and unlabeled datasets in round 1 and round 6, respectively.}
     \label{fig:disarray}
\end{figure}

As shown in \fig \ref{fig:disarray}, we visualize the trend of disarray $\rho^q$ in $\eqref{indicator2}$ and indicator $\gamma_{c}$ in $\eqref{indicator}$, where $c\in\{1,2,3,4\}$ indicate 4 cases of presence detection. Note that $\rho^q$ presents the variance of labeled/unlabeled data among subcarriers, whilst $\gamma_{c}$ additionally considers disparity of average disarray. As explained previously, a larger range of values for empty case is observed due to the comparatively smaller variation than human presence. On the contrary, for human presence case, CSI fluctuates over time with larger variance in each subcarrier resulting in a smaller value of $\eqref{indicator2}$. We can infer that empty case has the widest disarray range and largest indicator of $\gamma_1=219.51$, whereas the case of both room with human presence has the lowest indicator of $\gamma_1=184.27$. A moderate performance is inferred in cases of presence in either room A or B. Moreover, we can observe different trends between labeled and unlabeled datasets in the case of empty and human presence in room B, which potentially improves the prediction accuracy.

\subsubsection{Different Weights of $\lambda$ in Teacher Loss Functions}

\begin{figure}[!t]
     \centering
     \begin{subfigure}{0.24\textwidth}
         \centering
         \includegraphics[width=1\textwidth]{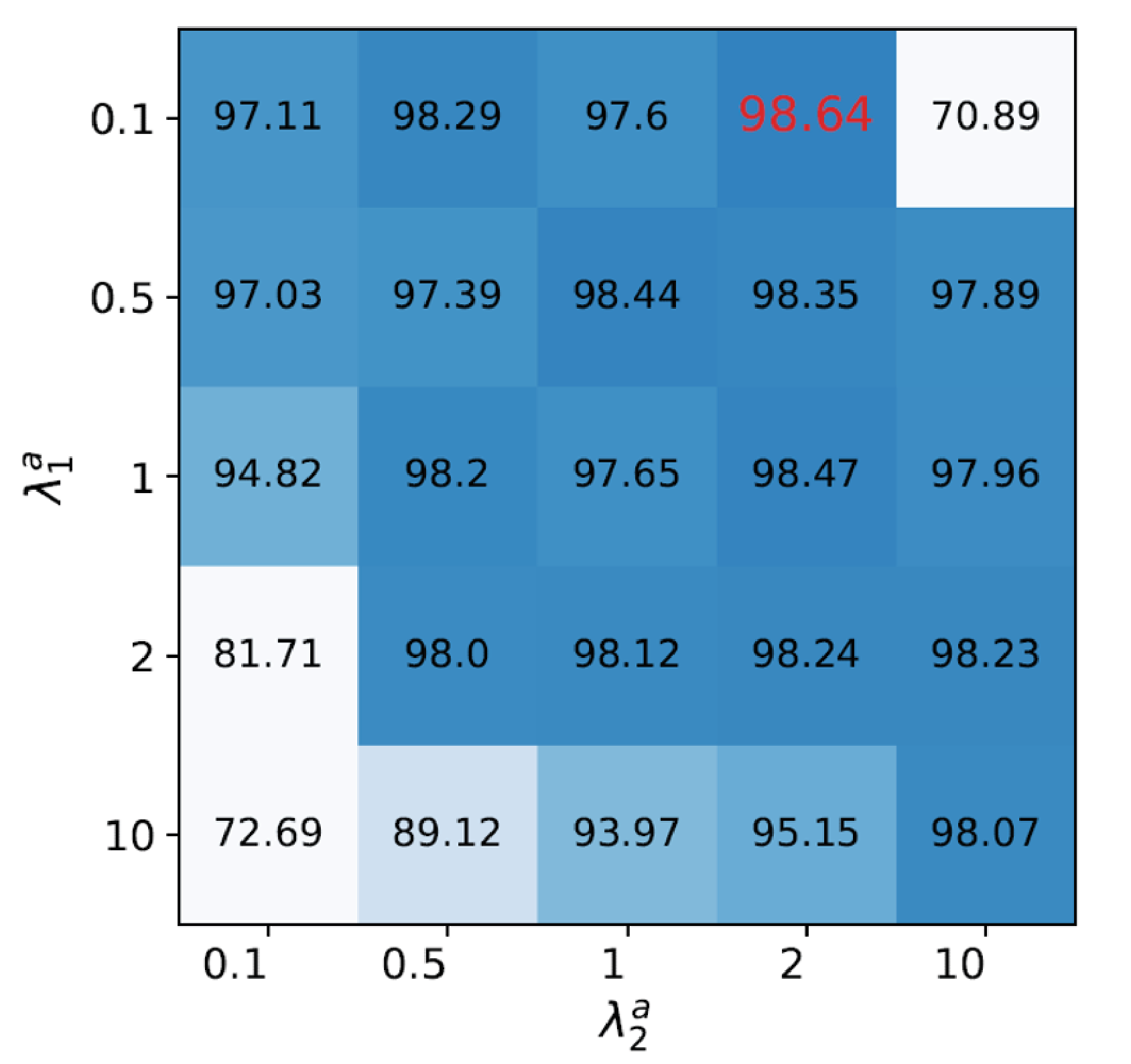}
         \caption{}
         \label{fig:f2_1}
     \end{subfigure}
     \begin{subfigure}{0.24\textwidth}
         \centering
         \includegraphics[width=1\textwidth]{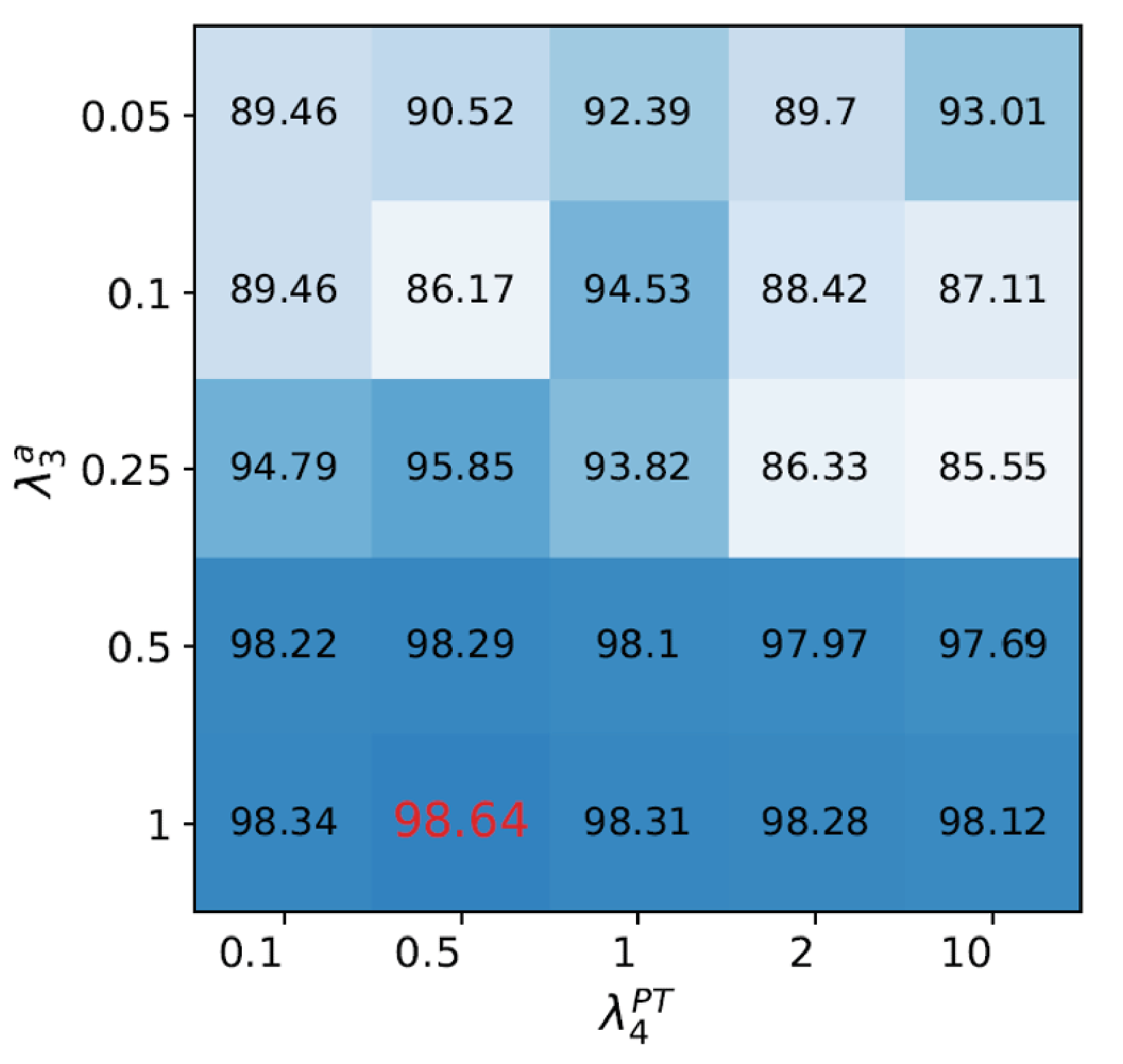}
         \caption{}
         \label{fig:f2_2}
     \end{subfigure}
        \caption{\small Accuracy with different hyperparameters in total loss. (a) Accuracy with different value of $\lambda^{a}_1$ and $\lambda^{a}_2$. (b) Accuracy with different value of $\lambda^{a}_3$ and $\lambda^{PT}_4$.}
        \label{fig:alpha_beta_disarray}
\end{figure}

In our system, we design by combining different loss functions for optimizing primal/dual teacher networks in $\eqref{totalloss1}$ and $\eqref{totalloss2}$, respectively. In Figs. \ref{fig:f2_1} and \ref{fig:f2_2}, we evaluate accuracy with different values of $\lambda$'s. Owning to compellingly high computational complexity, we only provide 2D comparison for $\{ \lambda_1^a, \lambda_2^a\}$ and for $\{ \lambda_3^a, \lambda_4^{PT}\}$, where $a\in\{PT, DT\}$ indicates primal/dual teacher networks, respectively. Note that we use the same $\lambda$ for teacher networks in each case, i.e., $\lambda^{PT}_c = \lambda^{DT}_c, \forall c\in\{1,2,3\}$. We notice that $\lambda_{2}^{a}$ has the highest weight because of the significance of feedbacks obtained from the student networks. While, fundamental cross entropy is considered less required with the smallest $\lambda_1^a$ since the labeled data is deemed to be simpler to be trained compared to that with unlabeled dataset. In this context, we select $\lambda^{a}_1=0.1$, $\lambda^{a}_2=2$, $\lambda^{a}_3=1$, and $\lambda^{PT}_4=0.5$ as the hyperparameters for the loss functions in the remaining simulations, which is summarized in Table \ref{Hyperparameters}.

\begin{table}[!t]
\small
\centering
\caption{Hyperparameters Setting}
\begin{tabular}{lcl}
\hline
Hyperparameters & Notation  & Value\\ \hline
Window size & $\tau$    & 50   \\
Diversity constant   & $\eta$    & 10000\\
Fine-step    & $\alpha$  & 1  \\ 
Fine-order   & $\beta$   & 1  \\ 
Weight of $L^{a}_{TCE}$ & $\lambda^{a}_1$   & 0.1  \\ 
Weight of $L^{a}_{UICE}$    & $\lambda^{a}_2$   & 2  \\ 
Weight of $L^{a}_{CTQ}$ & $\lambda^{a}_3$   & 1  \\ 
Weight of $L^{PT}_{CTVE}$    & $\lambda^{PT}_4$   & 0.5  \\
Data drift threshold & $d_{th}$ & 50 \\ \hline
\end{tabular}
\label{Hyperparameters}
\end{table}

\subsection{Effects of Loss Functions}

\subsubsection{Cross-Teacher Quadratic Loss}

\begin{figure}[!t]
\centering
\includegraphics[width=3in]{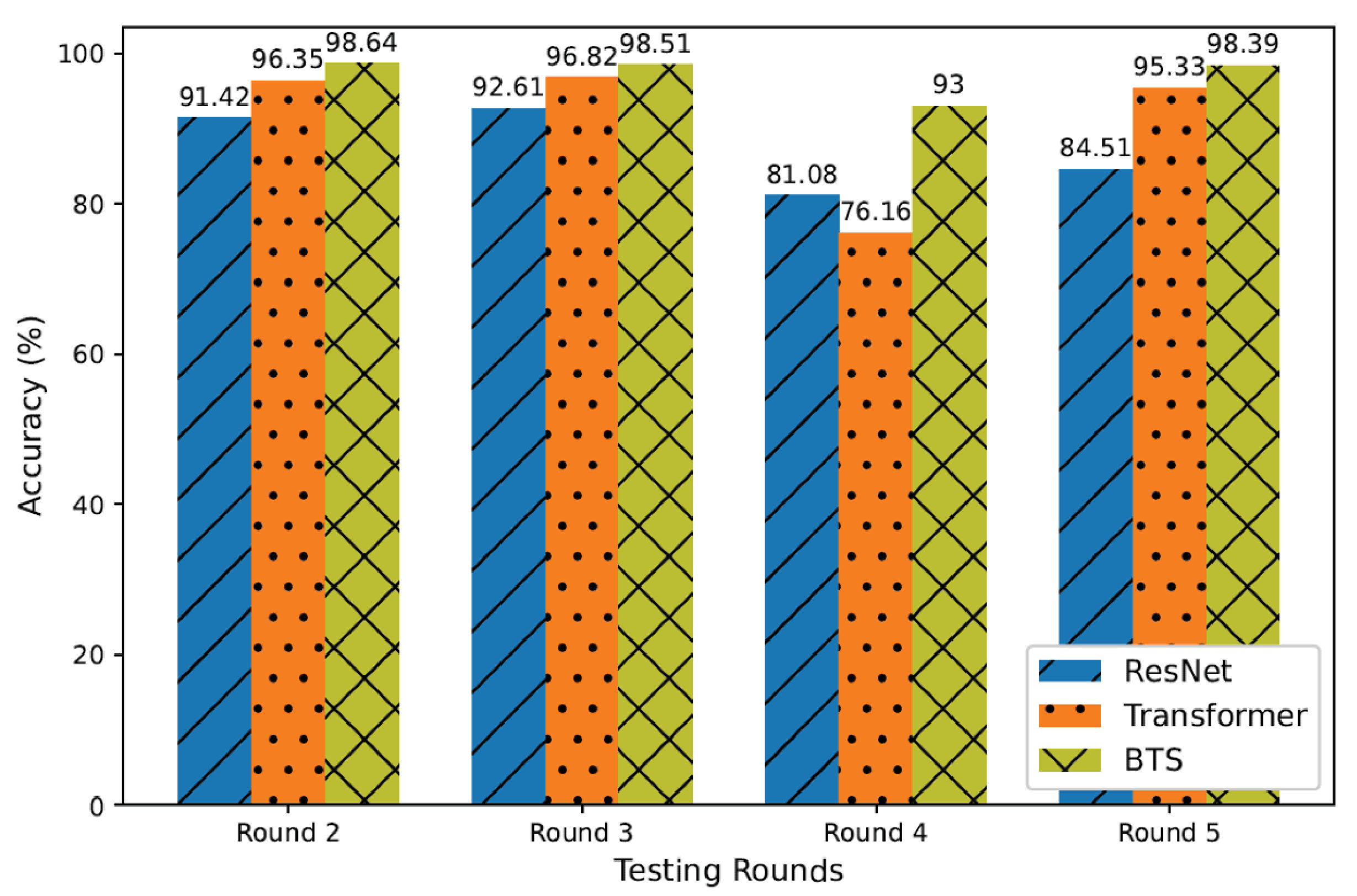}
\caption{\small Performance of accuracy comparing between BTS and a single network using either ResNet or transformer architecture.}
\label{fig:f3}
\end{figure}

In this subsection, we will discuss the cases with and without $L^{PT}_{CTQ}$, whilst the case without $L^{PT}_{CTQ}$ is evaluated based on two different networks. We train these cases with round 1 as labeled data and round 2 as unlabeled data, and evaluated the model from rounds 2 to 5. \fig \ref{fig:f3} shows the model performance in three mechanisms, i.e., ResNet, transformer and BTS. The blue slashed bars represent the ResNet-based single TS network, and the orange dotted bars represent the transformer-based single TS network. Furthermore, the bars with olive green crossed lines represent our proposed BTS system that leverages $L^{PT}_{CTQ}$ and combination of both ResNet and Transformer. As shown in \fig \ref{fig:f3}, transformer achieves a performance of $96.35\%$, $96.82\%$, and $95.33\%$ in round 2, round 3, and round 5 respectively, but only $76.16\%$ in round 4. We can infer from Section \ref{SYS_PRE} that some subcarriers perform abnormal behavior in round 4 due to changes in the outdoor environment. However, subcarriers possess significant information for transformer. When subcarriers have different patterns such as those in round 4, the results of embedding in transformer will be degraded. Alternatively, we consider spatial information to improve additional performance of the model. It can be noticed that ResNet outperforms transformer in round 4 due to feature extraction of spatial information. ResNet prefers to focus on common features learned from small regions, rather than directly referencing the entire subcarriers like transformer. Leveraging both advantages in ResNet and transformer architectures, the proposed BTS system is capable of employing different latent features in primal/dual networks. Accordingly, the design of loss in $L^{PT}_{CTQ}$ achieves the highest accuracy in all testing rounds, i.e., the evaluation of BTS in round 4 performs $12\%$ higher accuracy than ResNet and about $17\%$ higher than transformer.

\begin{figure}[!t]
\centering
\includegraphics[width=3in]{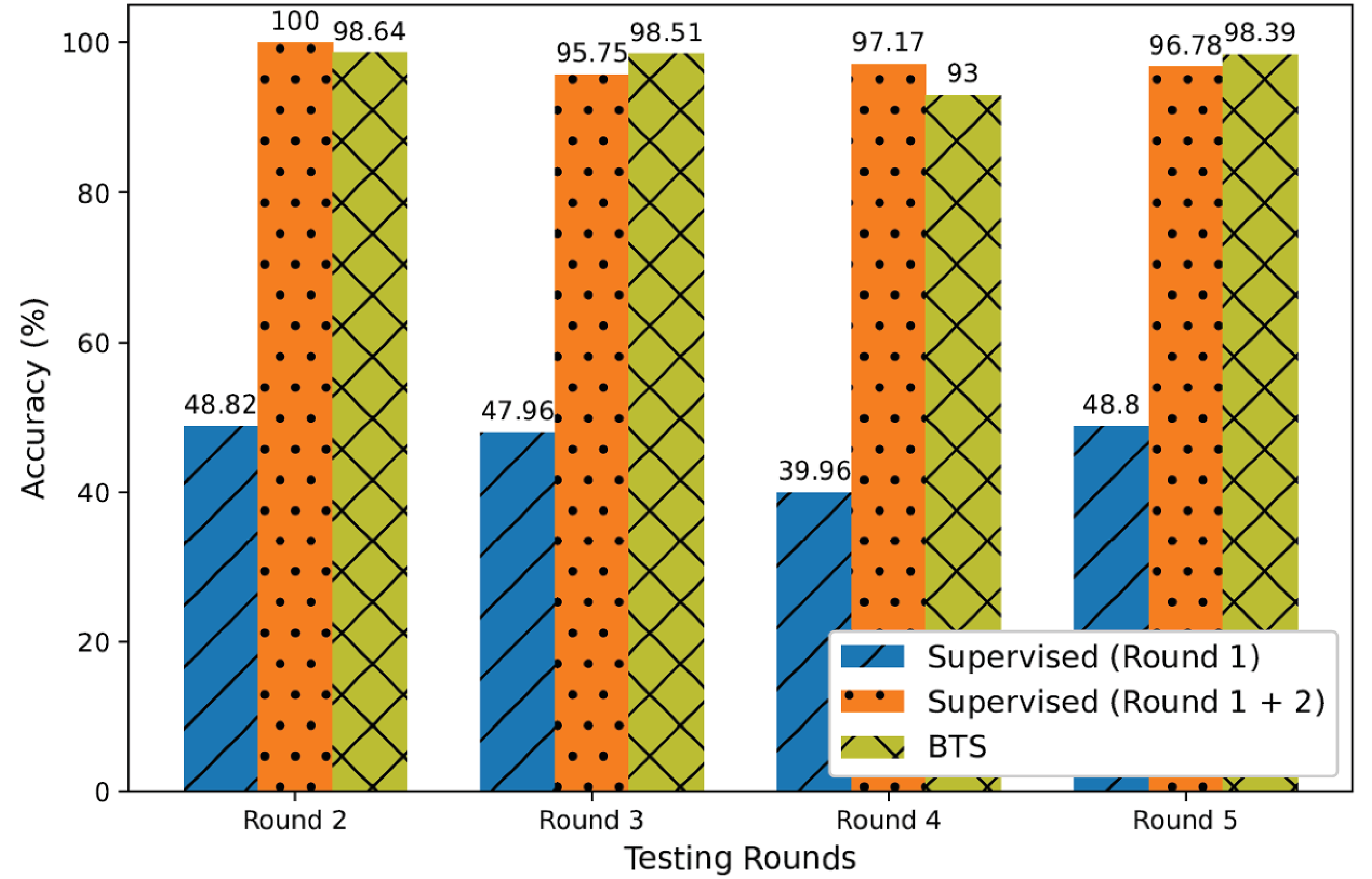}
\caption{\small Performance of accuracy  of proposed BTS system with supervised learning using either labeled dataset of round 1 or both rounds 1 and 2.}
\label{fig:f4}
\end{figure}

\begin{figure}[!t]
     \centering
     \includegraphics[width=3.5in]{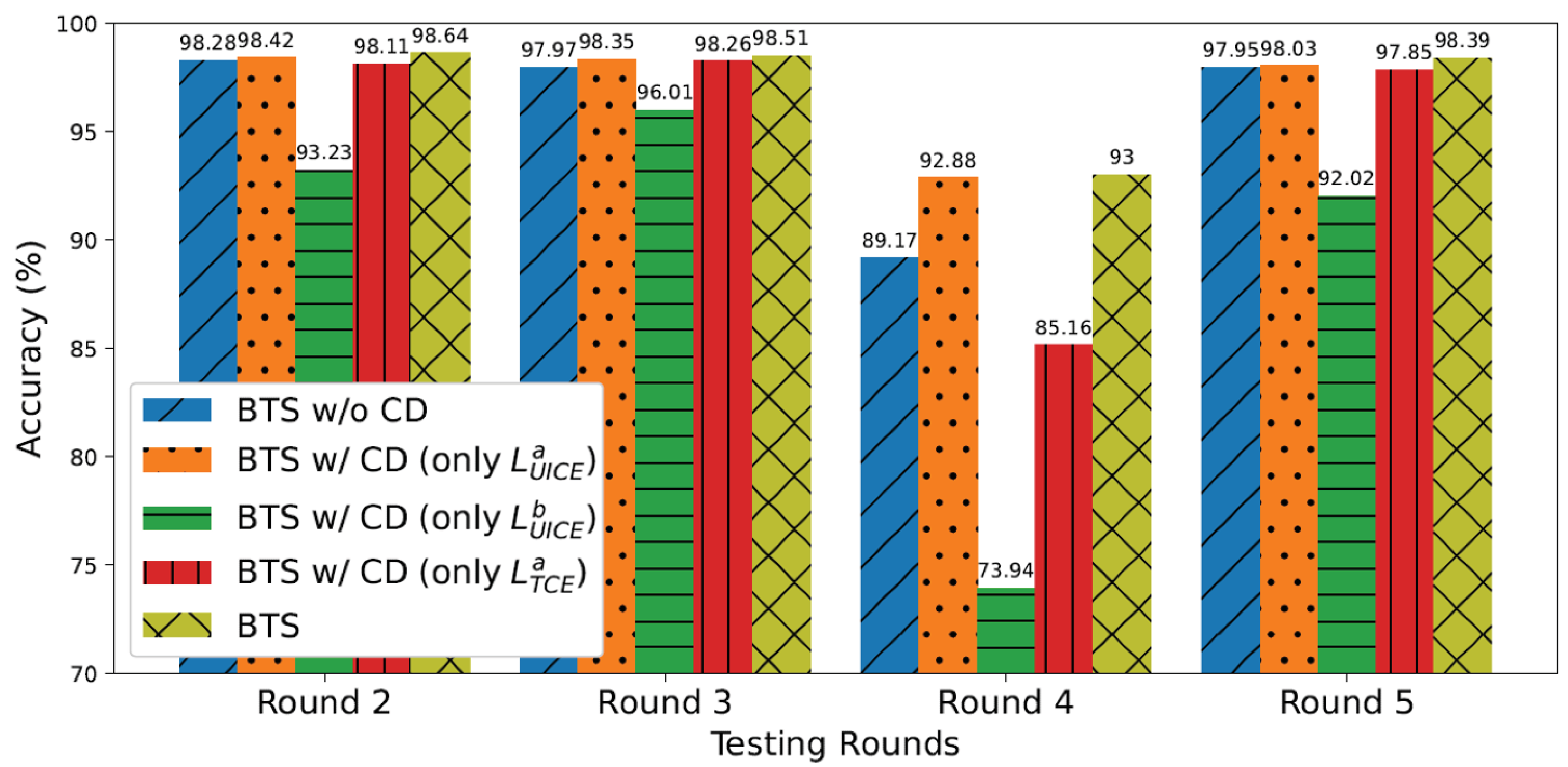}
     \caption{\small Performance comparison of proposed BTS system with and without confidence distribution mechanisms employing either UICE or TCE losses.}
     \label{fig:f6_1}
\end{figure}

\subsubsection{SSL in Unlabeled Indication Cross Entropy and Transformative Cross Entropy Losses} \label{subUICE}

We recognize that the UICE and TCE loss functions are closely related to TS in SSL. The proposed BTS system is designed to reduce the effort required to label collected CSI data and address the time-varying effect caused by environmental changes. However, there is often concern about whether the unlabeled dataset is effectively learned, which is validated through the evaluations in this experiment in \fig \ref{fig:f4}. Firstly, we train on round 1 as a labeled dataset using a transformer-based network, which is considered a benchmark, as indicated by the blue slashed bars. Secondly, we adopt both rounds 1 and 2 as labeled datasets for supervised-based training, which employs the same neural network as that in the first evaluation, as shown in orange dotted bars. Finally, we evaluate BTS by treating round 1 as labeled and round 2 as unlabeled dataset, as depicted in olive green crossed bars. We can observe from Fig. \ref{fig:f4} that utilizing a single round potentially leads to ineffective detection with time-varying CSI, i.e., it has an accuracy lower than $50\%$ from rounds 2 to 5. On the other hand, over $90\%$ accuracy is accomplished thanks to multi-timestamp data applied in supervised learning. While existing supervised learning methods require labeled data from both rounds 1 and 2, the proposed BTS system is able to leverage unlabeled data from round 2, which still achieves asymptotic accuracy to supervised learning using both labeled datasets.

In \fig \ref{fig:f6_1}, we compare five cases, including the case without utilization of confidence distribution (CD), usage of CD but with only either $L_{UICE}^a$, $L_{UICE}^b$, or $L_{TCE}^a$, and the proposed BTS system, which are shown in blue slashes, orange dots, green horizontal lines, red vertical lines, and olive green crossed lines, respectively. Note that CD is a non-training-based weak classifier that assists the neural network to learn from prior knowledge. As shown in \fig \ref{fig:f6_1}, we can observe that in the case of rounds 2, 3, and 5, the participation of CD has little impact on the accuracy performance due to similar data features. However, larger differences can be found in the case of environmental changes in round 4 due to the attained prior knowledge relevant to environmental change. Notice that the case of BTS with CD and only the loss function of $L^{b}_{UICE}$ has the lowest accuracy in all testing rounds. This is mainly because when the student network is the only one that knows the prior knowledge, there will be a biased answer towards itself, which further misleads the teachers with confusing feedback. Such a circumstance also impinges the teacher network with no faith to provide correct pseudo-labels for the students to learn. By considering all loss functions, the proposed BTS can achieve the highest accuracy among other baselines.

\subsubsection{Data Drift and CSI Time-Varying Euclidean Loss}

\begin{figure}
     \centering
     \begin{subfigure}{1.6in}
         \centering
         \includegraphics[width=1\textwidth]{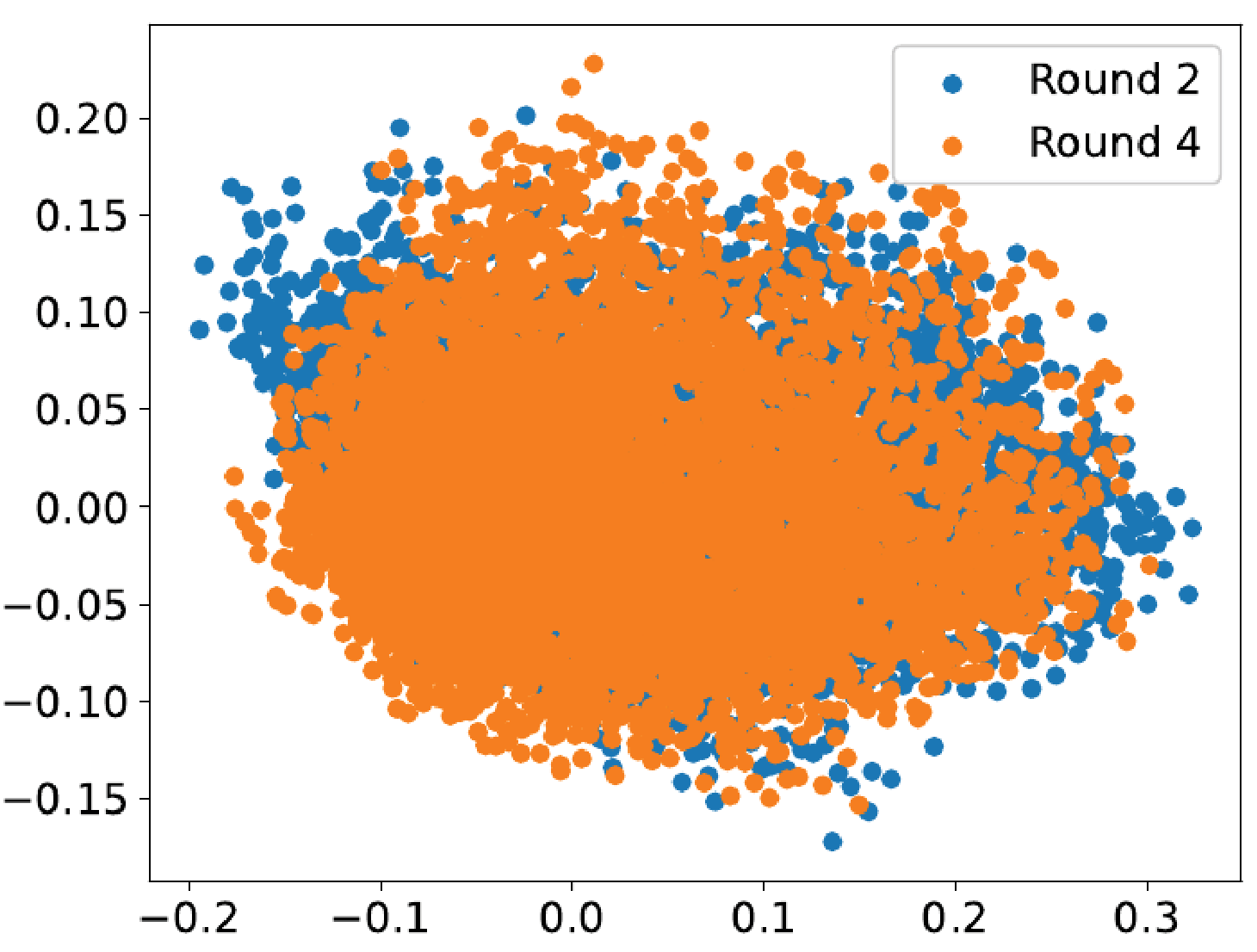}
         \caption{}
         \label{fig:f5_before_4}
     \end{subfigure}
     \begin{subfigure}{1.6in}
         \centering
         \includegraphics[width=1\textwidth]{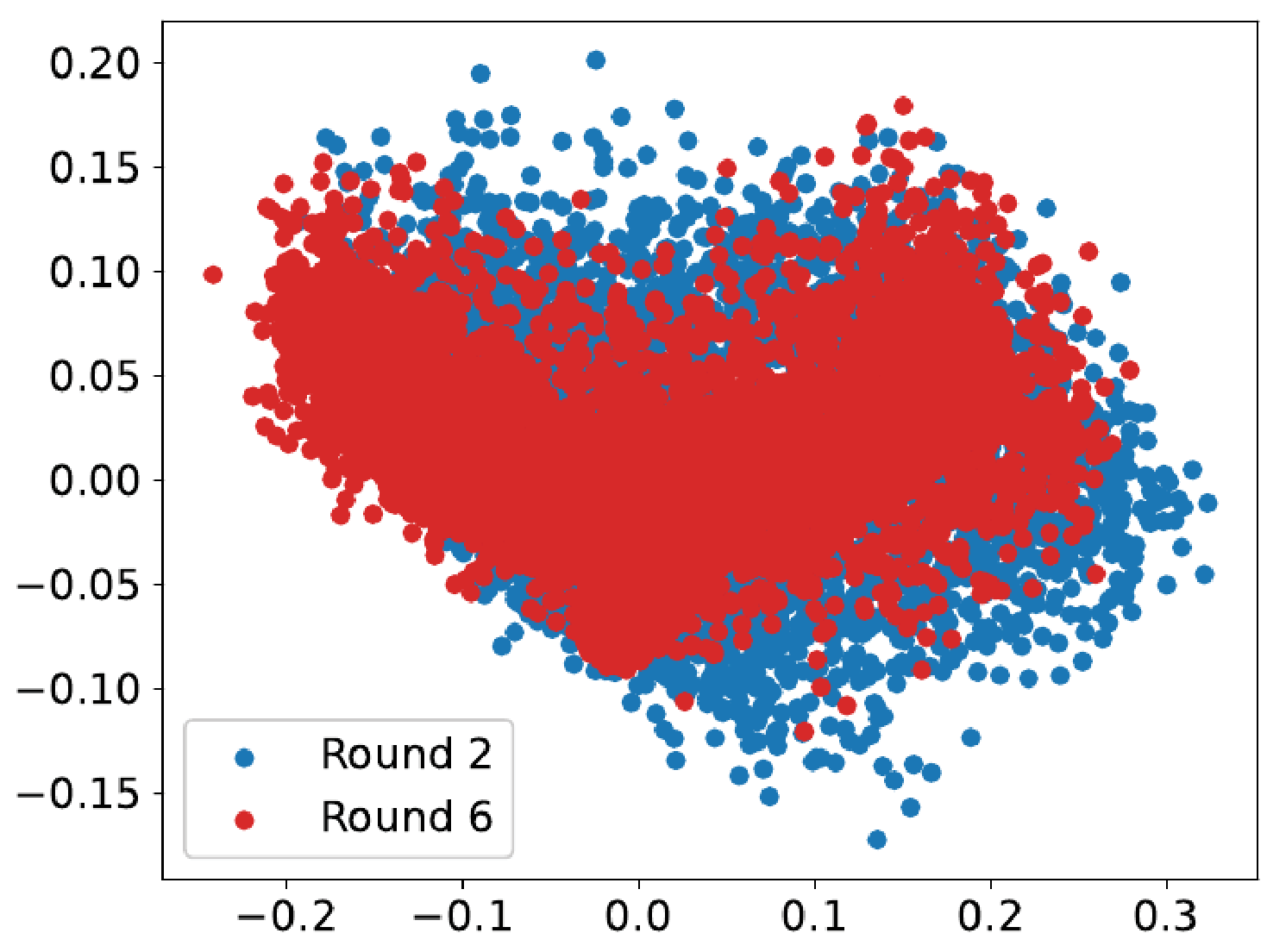}
         \caption{}
         \label{fig:f5_before_6}
     \end{subfigure}

     \begin{subfigure}{1.6in}
         \centering
         \includegraphics[width=1\textwidth]{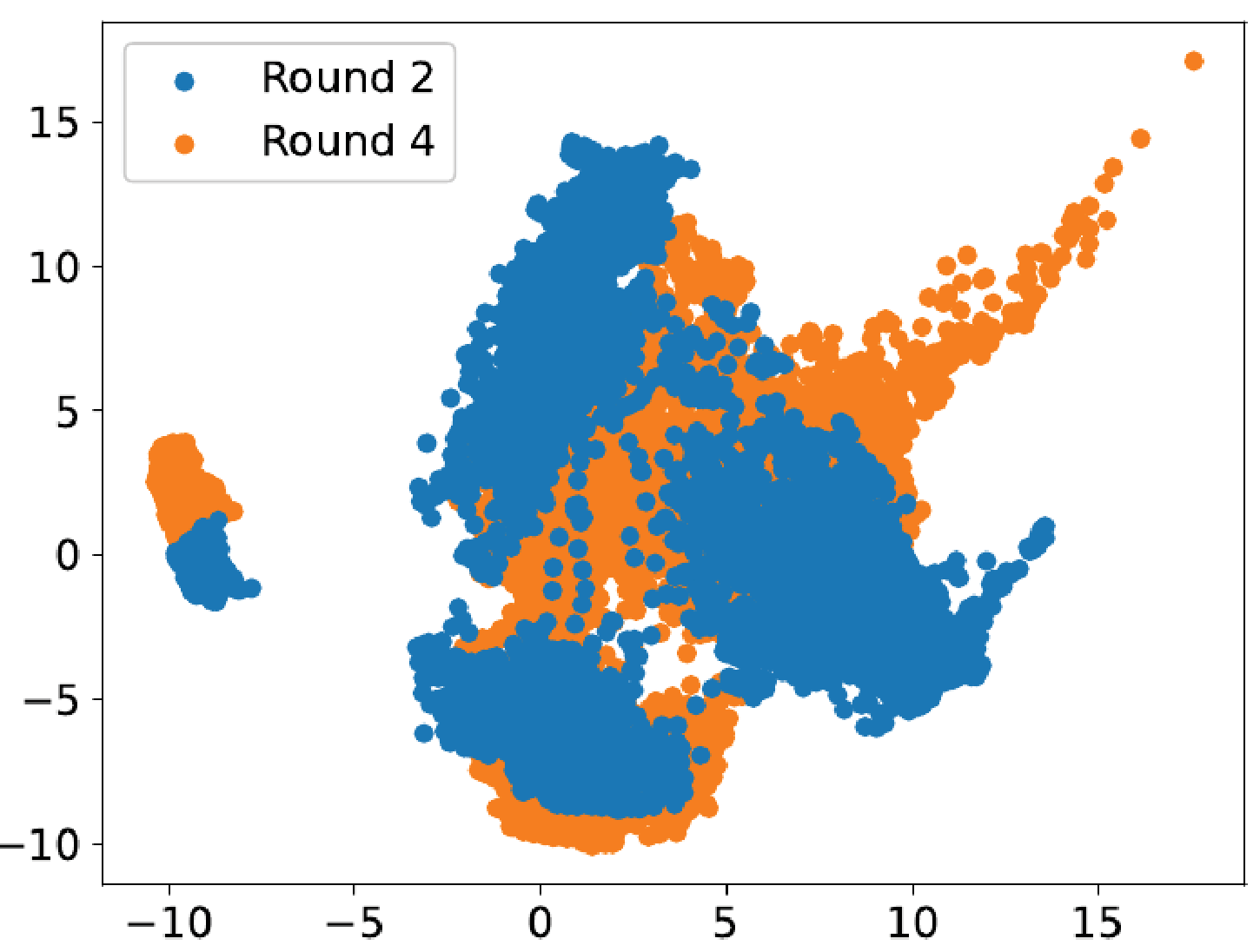}
         \caption{}
         \label{fig:f5_after_4}
     \end{subfigure}
     \begin{subfigure}{1.6in}
         \centering
         \includegraphics[width=1\textwidth]{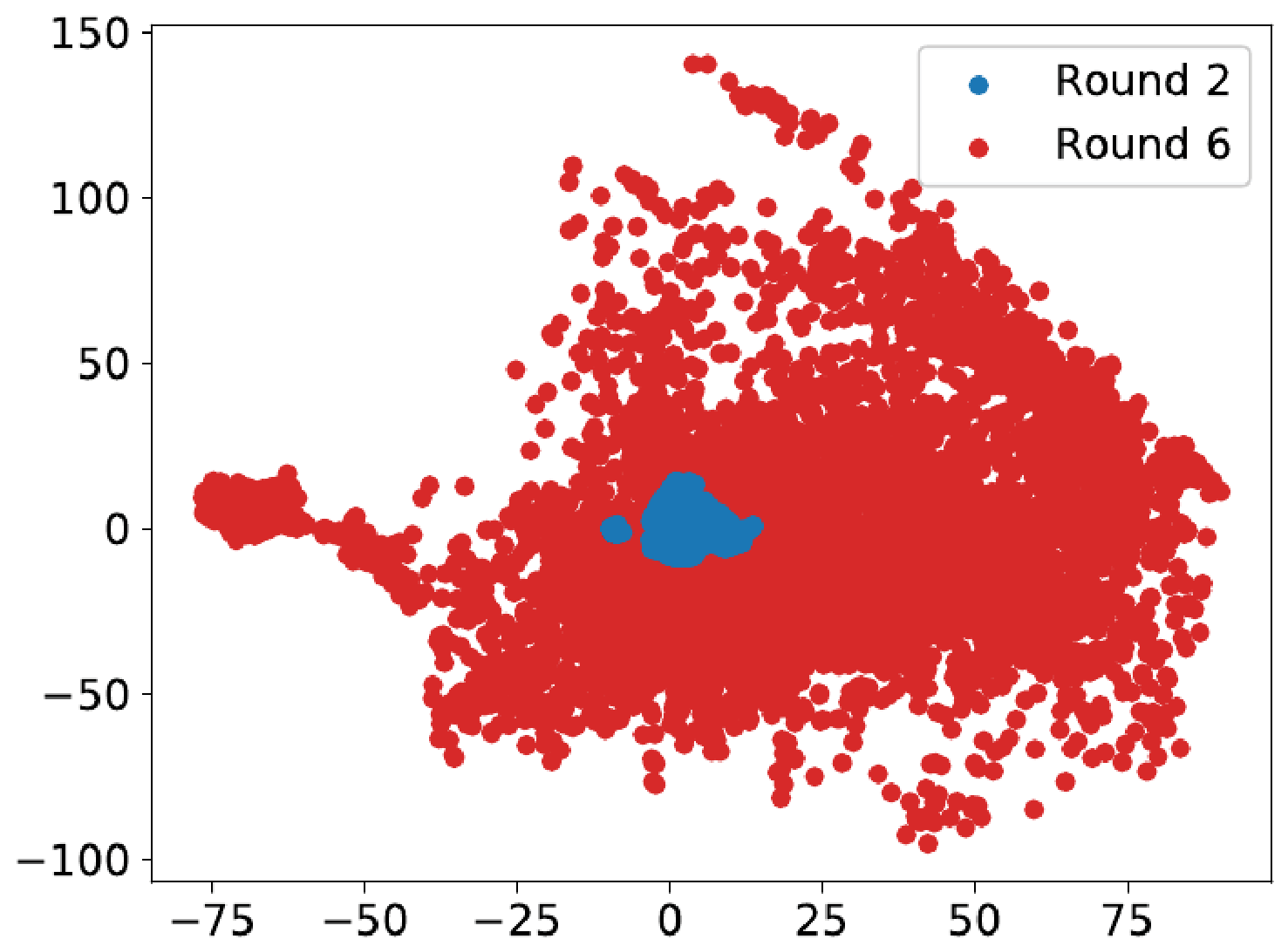}
         \caption{}
         \label{fig:f5_after_6}
     \end{subfigure}
        \caption{\small Distributions of different rounds before and after training with drift detection. (a) Rounds 2 and 4 before training. (b) Rounds 2 and 6 before training. (c) Rounds 2 and 4 after training. (d) Rounds 2 and 6 after training.}
        \label{fig:f5}
\end{figure}

Here, we demonstrate that by utilizing $L^{PT}_{CTVE}$ during the training stage, we can limit the similarity of data within a specific range, thereby distinguishing drifted data from outlier distances. In \fig \ref{fig:f5}, we present the dataset distribution before and after training with blue, orange and red dots indicating the representation vectors of rounds 2, 4, and 6, respectively. We utilize dataset of round 1 as the labeled dataset, whereas rounds 2, 4, 6 are unlabeled ones. Note that we visualize their relationships by employing dimension reduction from high-dimension to 2D plots. In \ref{fig:f5_before_4}, we can observe almost identical distributions even with changes of outdoor environments, whereas little difference can seen in \fig \ref{fig:f5_before_6} owing to different deployment of antennas. It is challenging to predict and detect the difference under these distributions even with occurrence of data drift. Leveraging the benefits of the designed loss functions $L_{CTVE}^{PT}$ can alleviate the mentioned problem. As can be readily inferred from Figs. \ref{fig:f5_after_4} and \ref{fig:f5_after_6} that the proposed BTS scheme can generate different distributions based on the designed loss functions. To elaborate a little further, due to severe time-varying effects, rounds 2 and 6 have substantially distinct distributions of representations.

Subsequently, in Table \ref{outlier_distance}, we have applied $\eqref{outlier}$ to calculate the outlier distance of every data point in each round, which shows the maximum and minimum distances of the representation vectors from rounds 2 to 6. Note that round 6 possesses the severe data drift owing to distinct deployment. From the mechanism training without $L^{PT}_{CTVE}$, i.e., not considering time-varying effects, we can observe that the data varies substantially from the distance of $7.37$ to $59.42$ from rounds 2 to 5. Moreover, the data with the smallest distance in round 6 overlaps with the range observed in round 5, and is also very close to that in round 4, which leading to a  difficulty to select a threshold of detecing data drift occurrence. On the contrary, the proposed BTS system considering CSI time-varying effect with $L^{PT}_{CTVE}$ can readily determine the threshold of drift detection. We can observe that the data with the same device deployment has a smaller range of data distribution, whilst data in round 6 is significantly separated from those in rounds 2 to 5, potentially improving detection accuracy.

\begin{table}[!t]
\footnotesize
\centering
\caption{Distances among Data Points}
\renewcommand{\arraystretch}{1.15}
\begin{tabular}{lccccc}
\hline
Without $L^{PT}_{CTVE}$  & Round 2 & 3 & 4 &  5 &  6 \\ \hline
Max Distance    & 7.37    & 11.24   & 31.53   & 59.42   & 648.67  \\
Min Distance     & 0.51    & 0.50    & 0.51    & 0.51    & 38.06   \\ \hline
With $L^{PT}_{CTVE}$     & Round 2 &  3 &  4 &  5 &  6 \\ \hline
Max Distance    & 4.58    & 22.22   & 9.72    & 12.18   & 1019.34 \\ 
Min Distance     & 0.58    & 0.59    & 0.58    & 0.57    & 82.52   \\ \hline
\end{tabular}
\label{outlier_distance}
\end{table}

\begin{figure}[!t]
     \centering
     \includegraphics[width=3in]{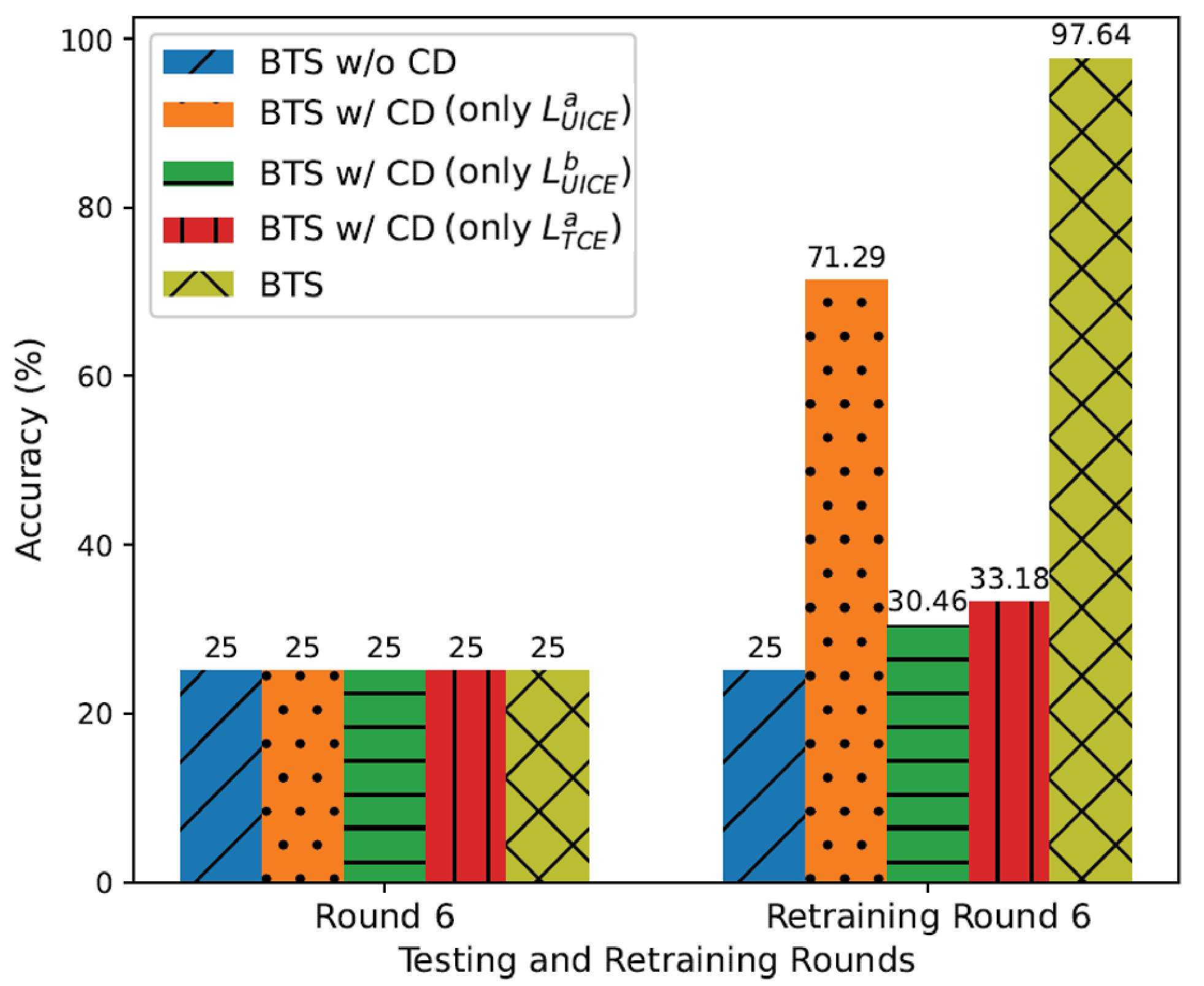}
     \caption{\small Performance of accuracy of BTS under severe time-varying effect in round 6 with and without retraining mechanism. BTS is compared with and without confidence distribution mechanisms employing either UICE or TCE losses.}
     \label{fig:f6_2}
\end{figure}

\begin{figure}[!t]
     \centering
     \includegraphics[width=3in]{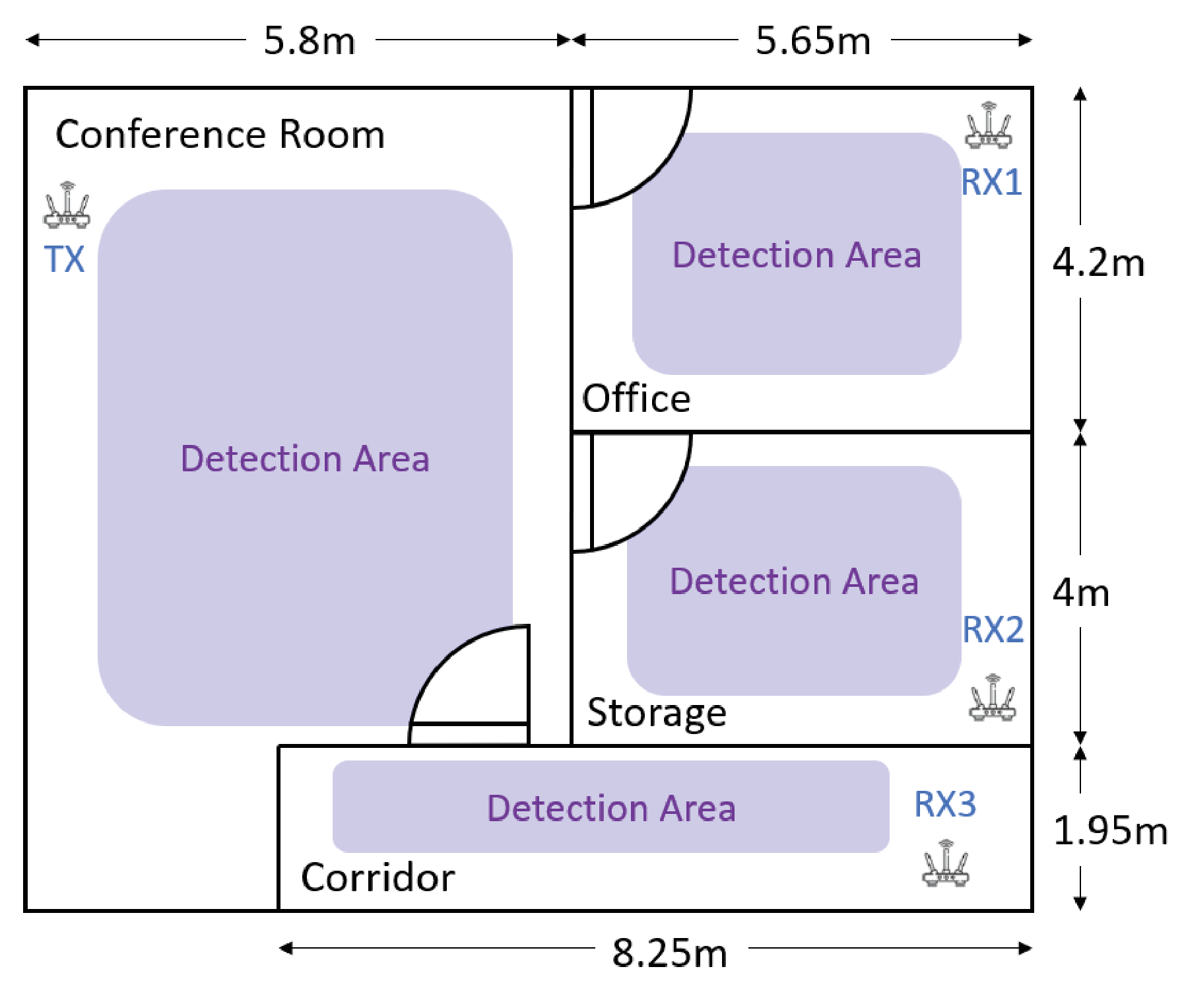}
     \caption{\small Experimental scenario of adjoining four rooms.}
     \label{fig:exp2}
\end{figure}

After detecting data drift as demonstrated in Table \ref{outlier_distance}, we further investigate the scenarios before and after retraining, as depicted in \fig \ref{fig:f6_2}. In the case of severe CSI time-varying effects, significant data drift leads to a random guess with only $25\%$ accuracy. However, with the proposed BTS system, retraining with CD and only loss function of $L^{a}_{UICE}$ can achieve an accuracy of $71.29\%$. This is because the teacher network can provide prior knowledge directly for the unlabeled data during training. Additionally, with the aid of the detection and retraining mechanism, the proposed BTS system can achieve the highest accuracy of $97.64\%$ with the adoption of all loss functions as well as CD.

\begin{table}[!t]
\centering
\scriptsize
\caption{Cases for Multi-Room Human Presence Detection and Activity Detection}
\begin{tabular}{|c|c|c|c|c|}
\hline
\textbf{Presence Case} & \textbf{Conference Room} & \textbf{Office} & \textbf{Storage Room} & \textbf{Corridor} \\ \hline
1  &  &         &         &         \\ \hline
2  & \checkmark &         &         &         \\ \hline
3  &           & \checkmark &         &         \\ \hline
4  &           &           & \checkmark &         \\ \hline
5  &           &           &           & \checkmark \\ \hline
6  & \checkmark & \checkmark &         &         \\ \hline
7  & \checkmark &           & \checkmark &         \\ \hline
8  & \checkmark &           &           & \checkmark \\ \hline
9  &           & \checkmark & \checkmark &         \\ \hline
10  &           & \checkmark &           & \checkmark \\ \hline
11 &           &           & \checkmark & \checkmark \\ \hline
12 & \checkmark & \checkmark & \checkmark &         \\ \hline
13 & \checkmark & \checkmark &           & \checkmark \\ \hline
14 & \checkmark &           & \checkmark & \checkmark \\ \hline
15 &           & \checkmark & \checkmark & \checkmark \\ \hline
16 & \checkmark & \checkmark & \checkmark & \checkmark \\ \hline
\hline
\textbf{Activity Case} & \textbf{Conference Room} & \textbf{Office} & \textbf{Storage Room} & \textbf{Corridor} \\ \hline
1  &     &    &    & - \\ \hline
2  & Walk    &    &    & - \\ \hline
3  & Run     &    &    & - \\ \hline
4  & Jump    &    &    & - \\ \hline
5  &    & Walk    &    & - \\ \hline
6  &    & Run     &    & - \\ \hline
7  &    & Jump    &    & - \\ \hline
8  &    &    & Walk    & - \\ \hline
9  &    &    & Run     & - \\ \hline
10  &    &    & Jump    & - \\ \hline
\end{tabular} \label{human_case}
\end{table}

\begin{figure*}
     \centering
     \begin{subfigure}{2.3in}
         \centering
         \includegraphics[width=1\textwidth]{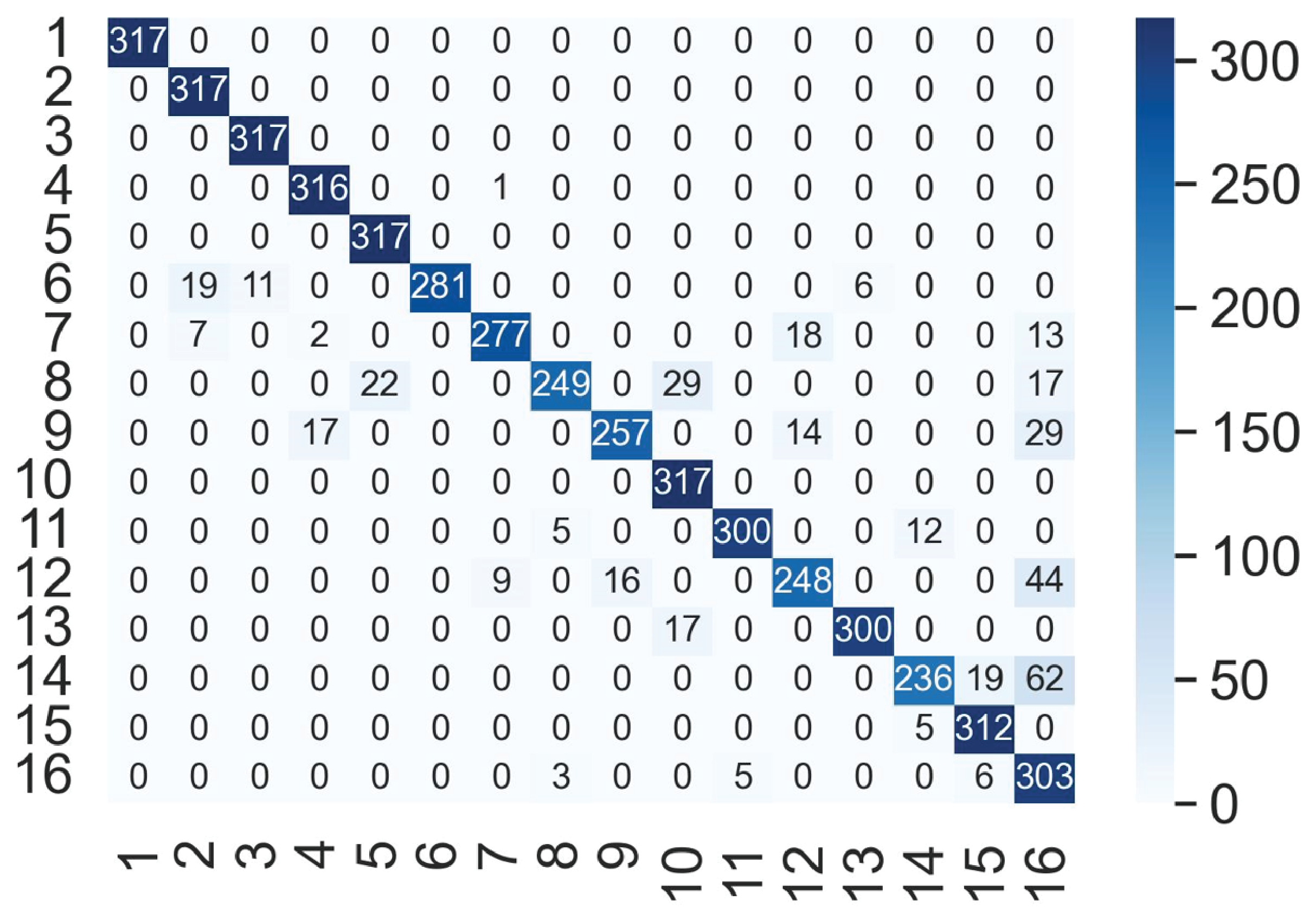}
         \caption{}
         \label{fig:task1_dnn}
     \end{subfigure}
     \begin{subfigure}{2.3in}
         \centering
         \includegraphics[width=1\textwidth]{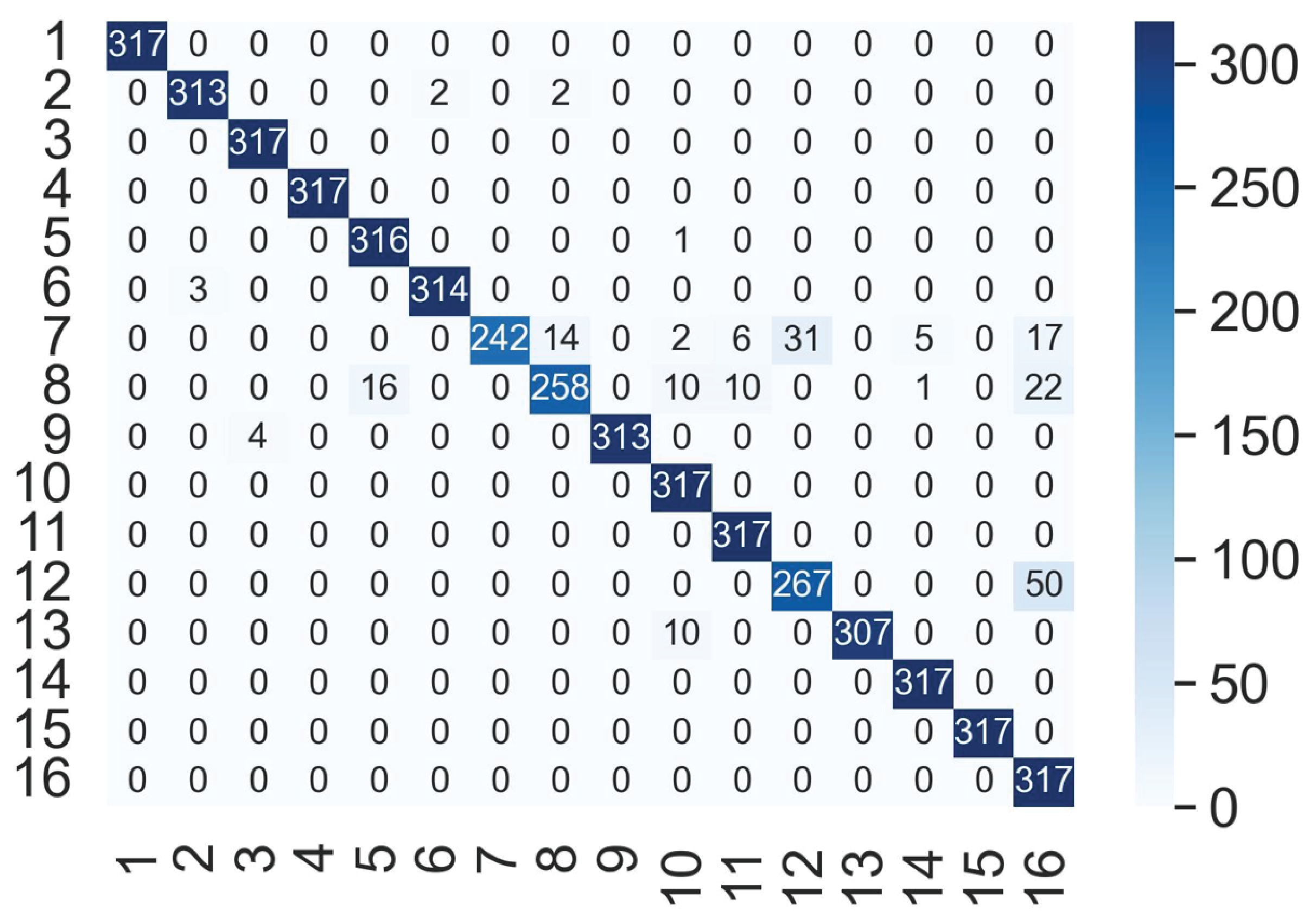}
         \caption{}
         \label{fig:task1_cnn}
     \end{subfigure}
     \begin{subfigure}{2.3in}
         \centering
         \includegraphics[width=1\textwidth]{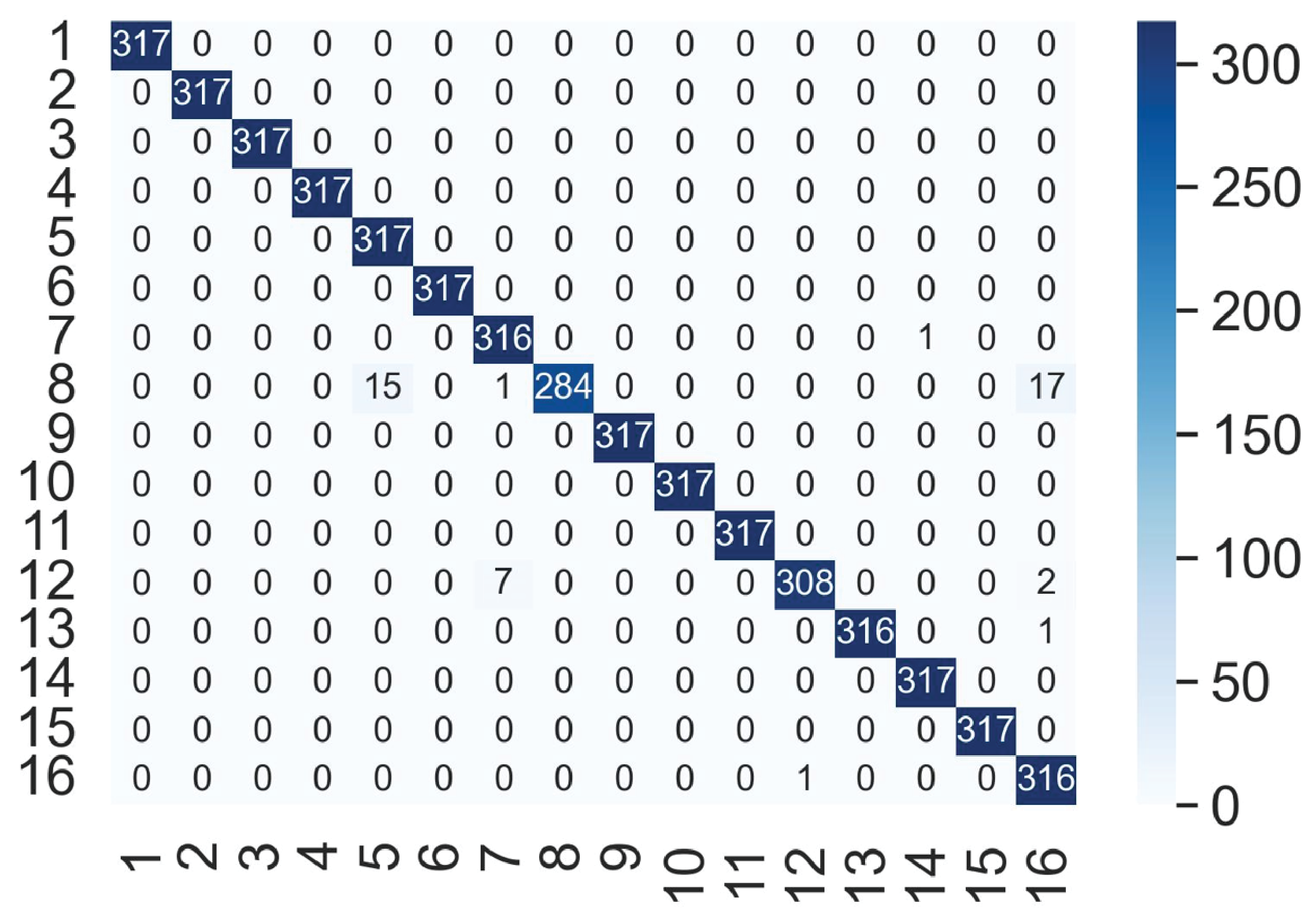}
         \caption{}
         \label{fig:task1_trans}
     \end{subfigure}

     \begin{subfigure}{2.3in}
         \centering
         \includegraphics[width=1\textwidth]{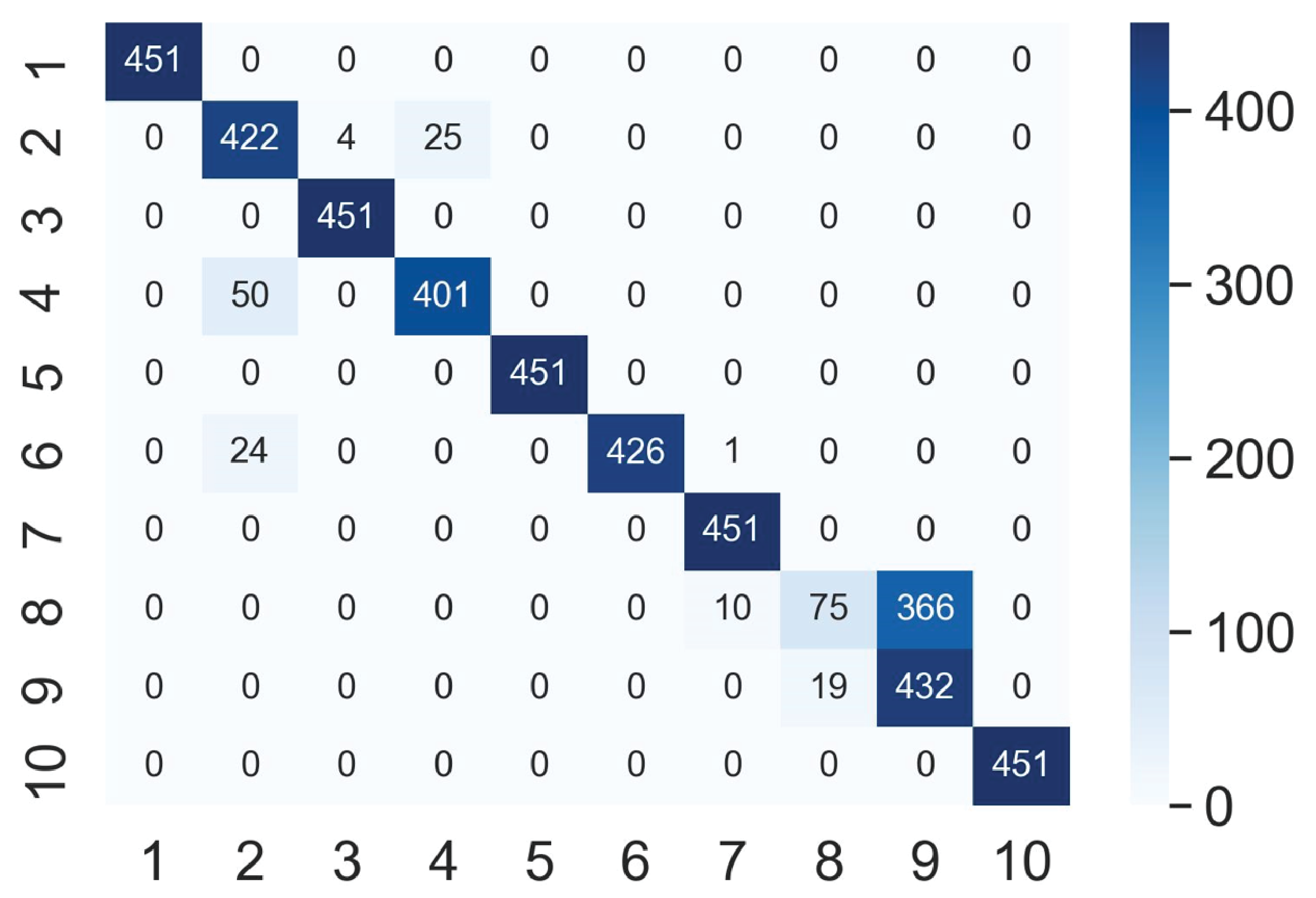}
         \caption{}
         \label{fig:task2_dnn}
     \end{subfigure}
     \begin{subfigure}{2.3in}
         \centering
         \includegraphics[width=1\textwidth]{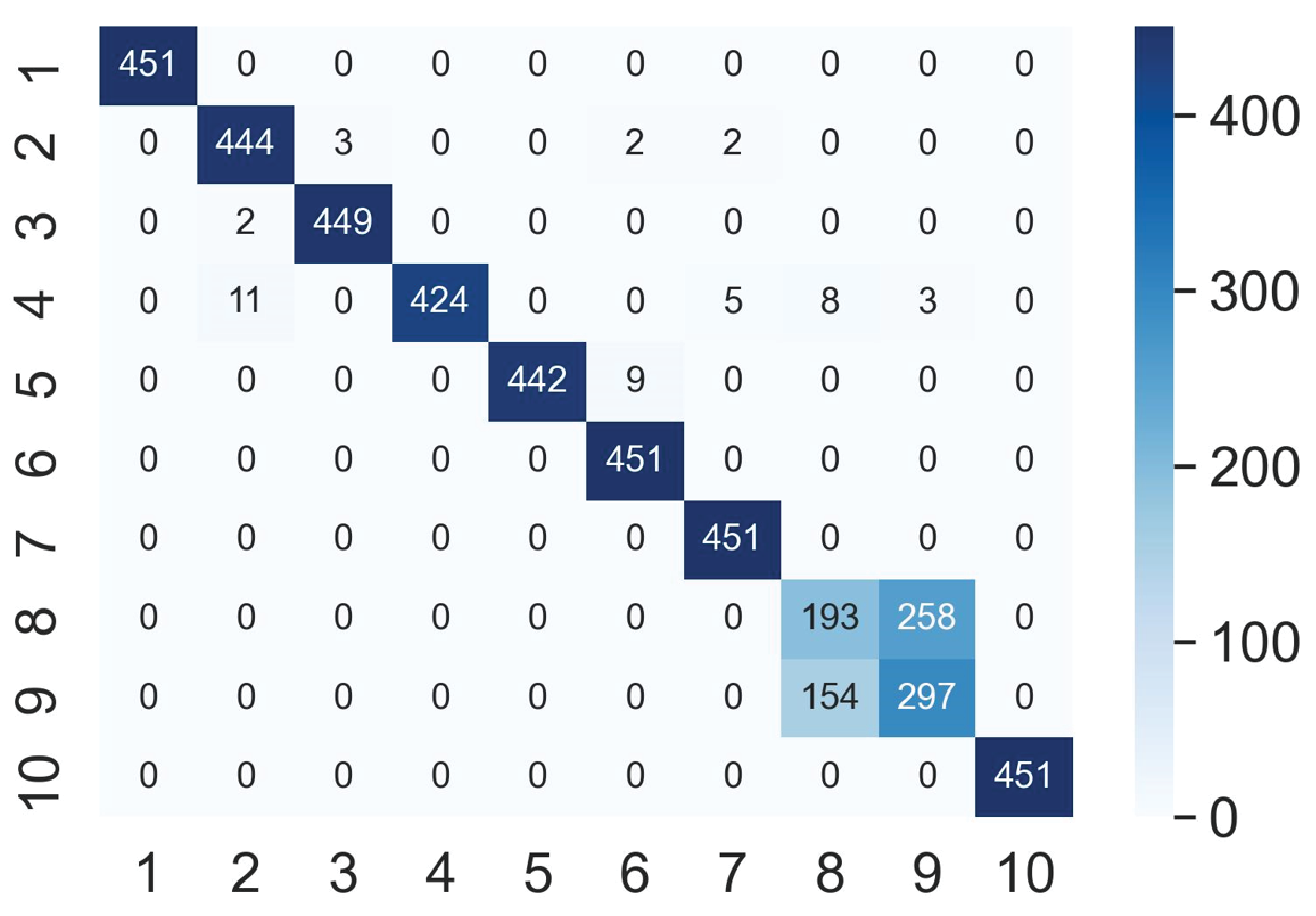}
         \caption{}
         \label{fig:task2_cnn}
     \end{subfigure}
     \begin{subfigure}{2.3in}
         \centering
         \includegraphics[width=1\textwidth]{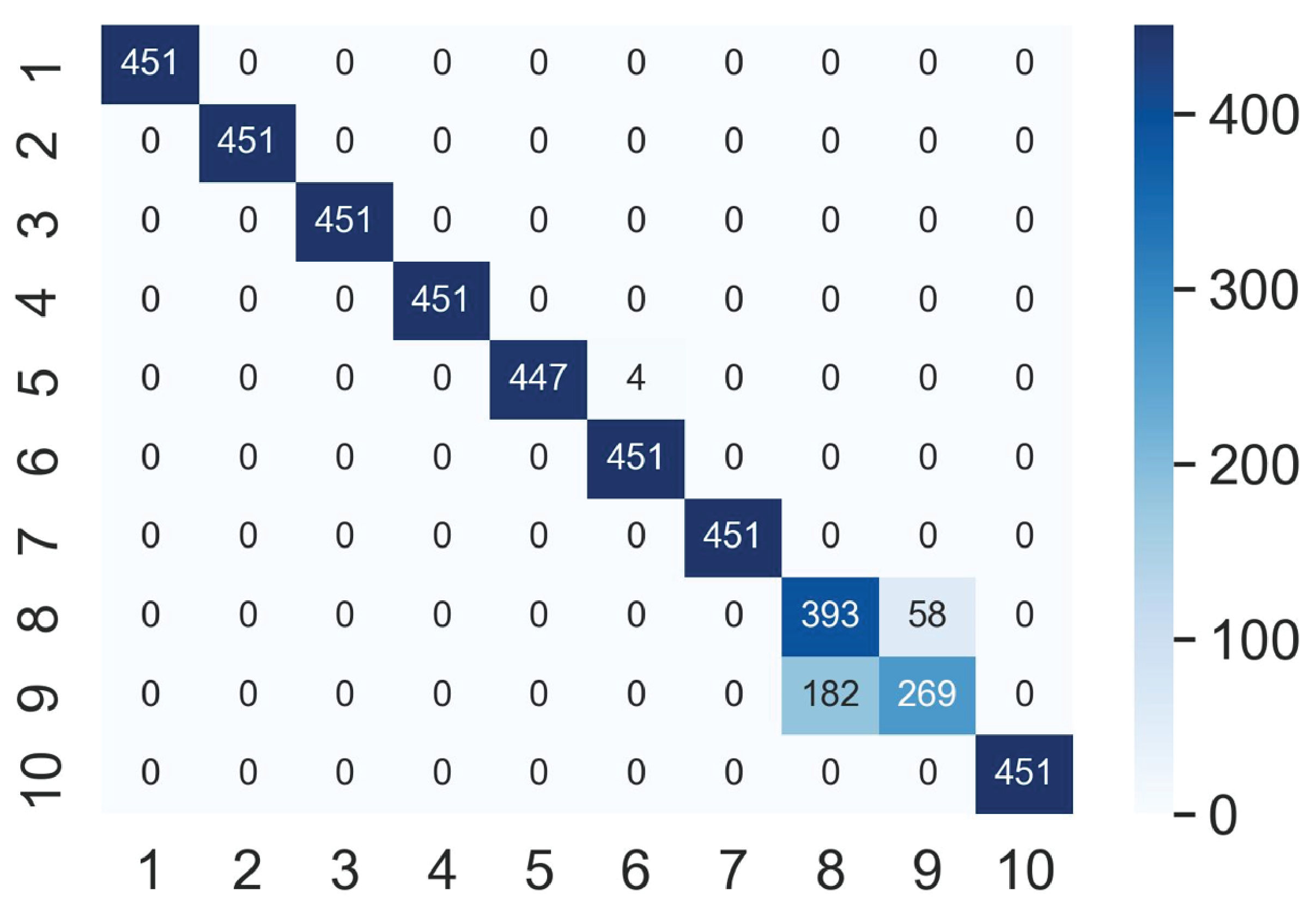}
         \caption{}
         \label{fig:task2_trans}
     \end{subfigure}
        \caption{\small Confusion matrix of multi-room human presence detection using (a) DNN, (b) CNN and (c) transformer, with the respective accuracies of $\{91.9 \%, 95.9\%, 99.1\%\}$. Confusion matrix of human activity detection using (d) DNN, (e) CNN and (f) transformer, with the respective accuracies of $\{88.9 \%, 89.8\%, 94.5\%\}$. }
        \label{fig:task}
\end{figure*}

\subsection{Multi-Room Presence and Activity Detection} \label{newsub}

In Table \ref{human_case}, we have listed the experimental cases for multi-person presence and activity detection. For presence detection, we experiment on a total combination of 16 labels for human presence and absence across a four-room scenario with a conference room, office, storage room and corridor, as shown in Fig. \ref{fig:exp2}. Three TX-RX pairs are required for covering all four rooms based on \cite{new_chu}. Note that case 1 in presence detection indicates the scenario where all rooms empty. While, 10 cases are given for activity detection, including a person walking, running and jumping in a three-room scenario with a conference room, office, and storage room. Note that few cases are considered here due to excessive combinations with a total of $(2 \times 3 )^4 = 1296$ possible cases, leading to difficulties of data collection and training. 

The confusion matrices in Figs. \ref{fig:task1_dnn}--\ref{fig:task1_trans} show the performance of DNN, convolutional neural network (CNN), and transformer models for detecting human presence in four-room settings, where the x-axis indicates the ground-truth labels and the y-axis means the predicted outcome labels. Each case represents a unique configuration where only one or multiple rooms are occupied, while others remain empty. It can be inferred that more misclassifications take place for cases 7--15 in both DNN and CNN owing to the similar features when multi-room presence. We can observe that the transformer model achieves the highest accuracy of $99.1\%$, which is followed by the CNN with an accuracy of $95.9\%$ and the DNN with an accuracy of $91.9\%$, demonstrating the effectiveness of the transformer's attention mechanism in capturing spatial relationships across rooms. Moreover, Figs. \ref{fig:task2_dnn}--\ref{fig:task2_trans} illustrate confusion matrices for human activity detection in a three-room scenario, with labels indicating specific actions of walking, running, and jumping occurring in a particular room while the other rooms remain empty. This setup focuses the models on recognizing specific activities in isolated spaces, presenting a different challenge compared to presence detection. Confusion arises when classifying cases 8 and 9 in the storage room due to the similar features of the environmental items in the room, which create indistinguishable signal reflections. We can observe that the transformer model outperforms both CNN and DNN methods, achieving an accuracy of $94.5\%$ compared to $89.8\%$ and $88.9\%$, respectively. The attention mechanism in transformer enables feature capturing for subtle differences in human activity patterns, leading to more accurate activity classification. The above performance improvement underscores the suitability of the transformer architecture for the tasks requiring temporal and spatial understanding in the more complex environments.

\begin{table*}[!t]
\centering
\caption{Performance of benchmark comparisons}
\begin{tabular}{lccccc}
\hline\hline
Supervised Learning                 & Round 2 & Round 3 & Round 4 & Round 5 & Round 6 \\ \hline
Teacher (round 1)       & 48.82              & 47.96                & 39.96                & 48.80        & -  \\
Teacher (rounds 1 and 2)   & $\boldsymbol{100}$ & 95.75                & $\boldsymbol{97.17}$ & 96.78        & - \\
Parallel CNN \cite{Deep_Learning2}            & 99.60              & 90.44                & 73.20                & 87.07        & 99.80  \\
AE-LRCN \cite{Deep_Learning6}                & $\boldsymbol{100}$ & 93.87                & 96.76                & 98.07        & $\boldsymbol{100}$  \\ 
BTS                     & 98.64              & $\boldsymbol{98.51}$ & 93                  & $\boldsymbol{98.39}$   & 97.64  \\ \hline\hline
Semi-Supervised Learning             & Round 2 & Round 3 & Round 4 & Round 5 & Round 6 \\ \hline
Single ResNet TS        & 91.42   & 92.61   & 81.08   & 84.51   & 31.46  \\
Single Transformer TS   & 96.35   & 96.82   & 76.16   & 95.33   & 35.58  \\
MPL \cite{MPL}          & 89.73   & 82.48   & 32.79   & 85.60   & 48.17  \\
CsiGAN \cite{CsiGAN}    & 85.52   & 86.52   & 79.08   & 84.23   & 83.74  \\
BTS                     & $\boldsymbol{98.64}$   & $\boldsymbol{98.51}$   & $\boldsymbol{93}$      & $\boldsymbol{98.39}$   & $\boldsymbol{97.64}$  \\ \hline\hline
\end{tabular}
\\
\label{summary}
\end{table*}

\subsection{Benchmark Comparison}

We have compared the performance of proposed BTS scheme with existing SL as well as SSL-based mechanisms. For SL comparison, we adopt both rounds 1 and 2 as labeled data and evaluate with round 3. For SSL comparison, we consider round 1 as the labeled dataset, and either round 2 or 6 as the unlabeled dataset. We consider four different cases of presence detection, including an empty room, human presence in either room A or B, and human presence in both rooms. 

\begin{figure}[!t]
\centering
\includegraphics[width=3in]{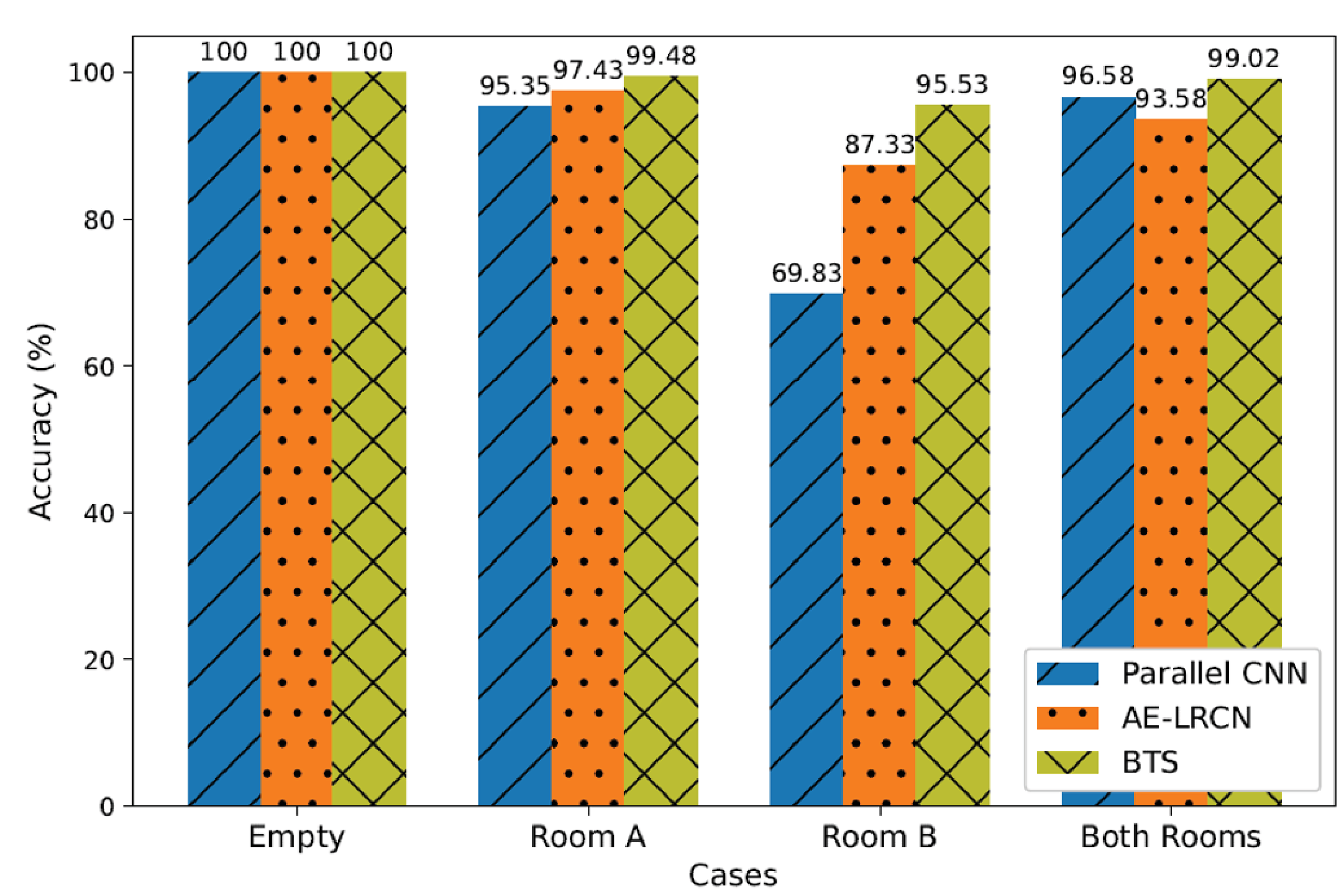}
\caption{\small Performance of accuracy of BTS comparing to different SL methods, i.e., parallel CNN and AE-LRCN. We consider both rounds 1 and 2 as labeled data and evaluate with round 3. Note that BTS still regard data in round 2 as unlabeld dataset.}
\label{fig:f7}
\end{figure}

In \fig \ref{fig:f7}, we compare two different SL methods, i.e., parallel CNN \cite{Deep_Learning2} and AE-LRCN \cite{Deep_Learning6}. Note that both comparison benchmarks require both rounds 1 and 2 as labeled data, whereas BTS uses round 2 as the unlabeled one. We can observe that parallel CNN has the lowest accuracy in room B. This is because parallel CNN utilizes both the amplitude and phase of CSI, whereas the spatial information in room B is very similar to that in the case of human presence in both rooms, which makes this CNN-based model fail to predict the room B case accurately. On the other hand, AE-LRCN has lower accuracy than BTS in cases of human presence since spatial and spectral information of CSI is not appropriately extracted. In a nutshell, the proposed BTS system achieves the highest accuracy of $100\%$, $99.48\%$, $95.53\%$, and $99.02\%$ in the respective presence detection cases, which has the benefit of only requiring half number of labels for training.

\begin{figure}
     \centering
     \begin{subfigure}{0.4\textwidth}
         \centering
         \includegraphics[width=1\textwidth]{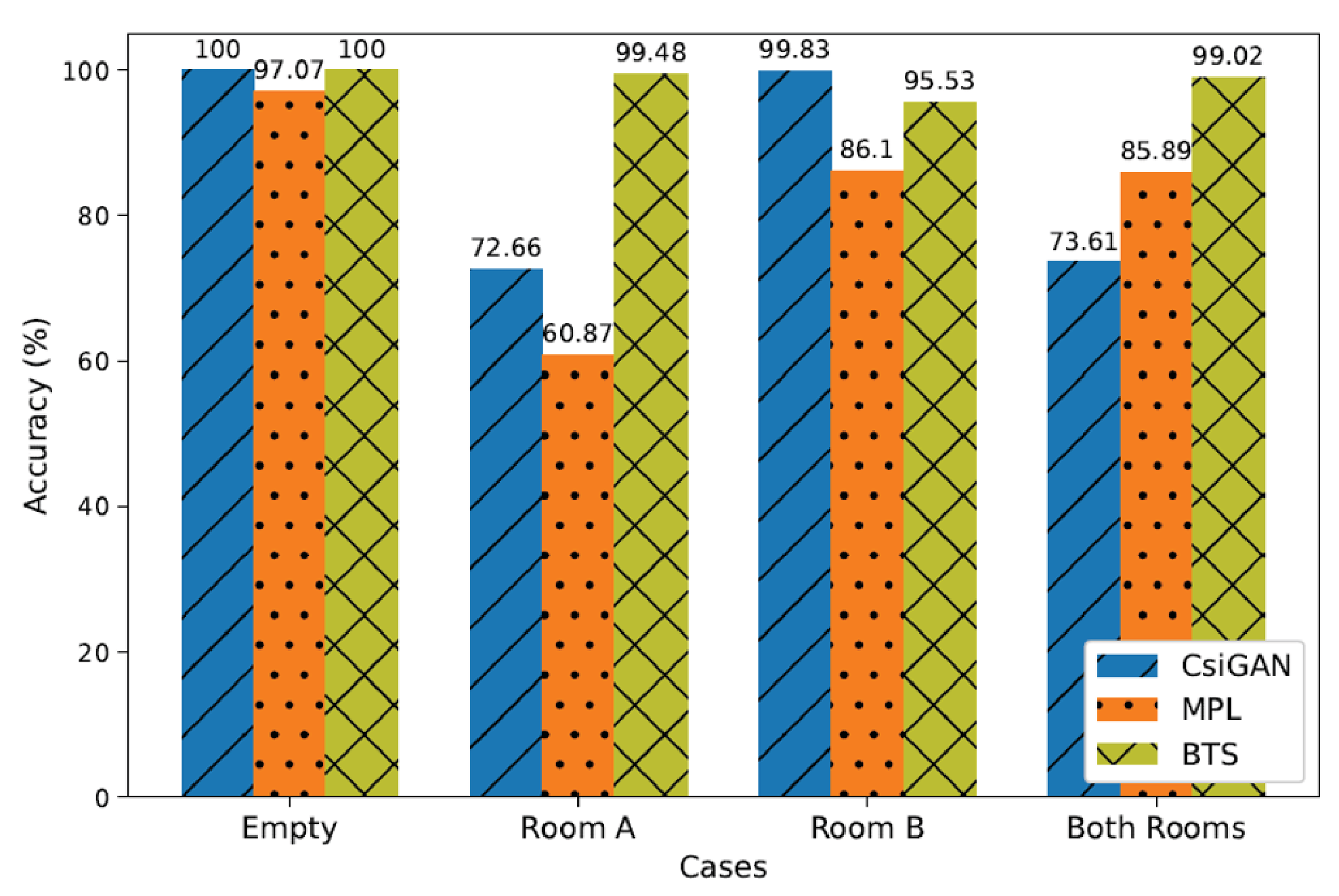}
         \caption{}
         \label{fig:f9_1}
     \end{subfigure}
     \quad
     \begin{subfigure}{0.4\textwidth}
         \centering
         \includegraphics[width=1\textwidth]{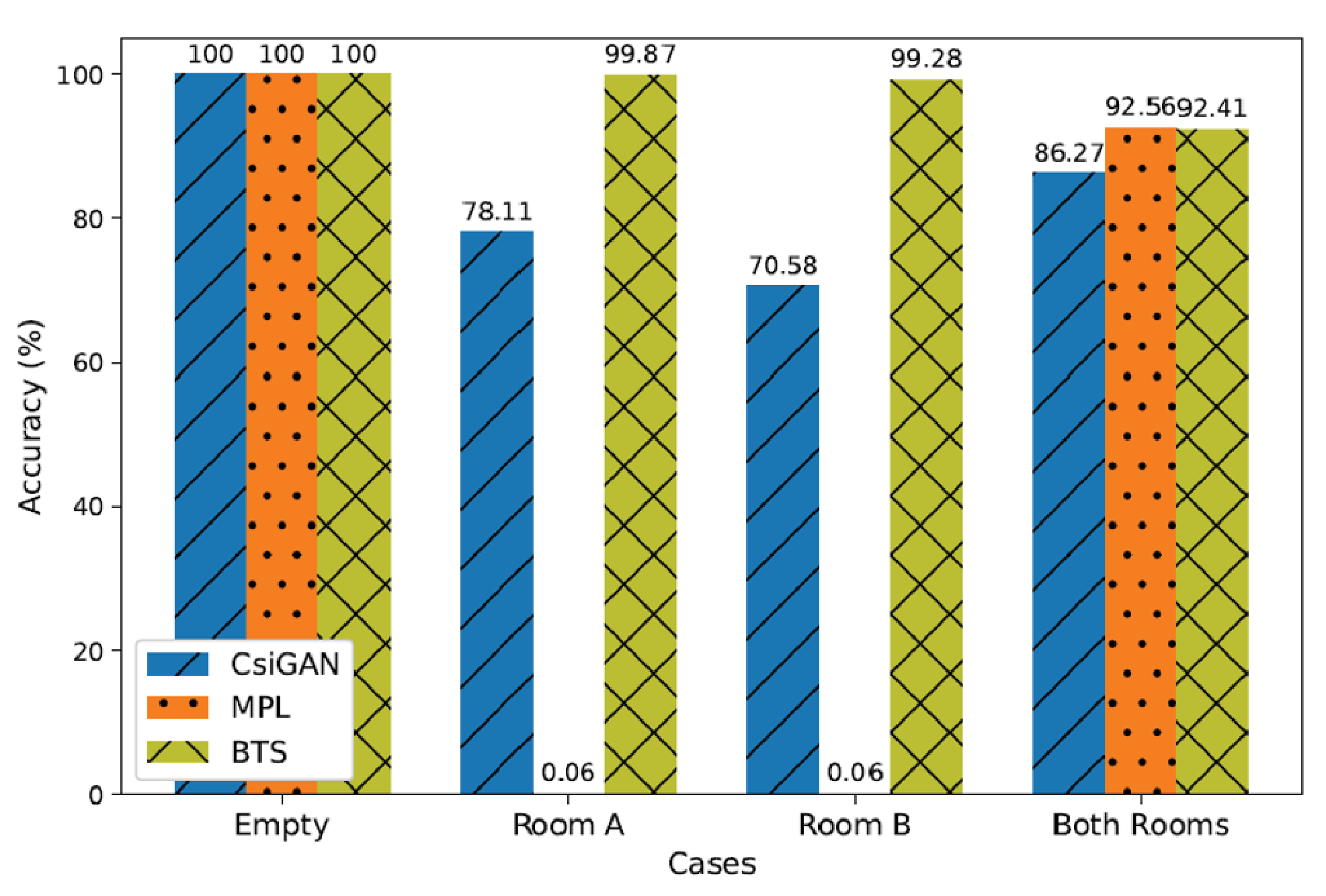}
         \caption{}
         \label{fig:f9_2}
     \end{subfigure}
        \caption{\small Performance of accuracy of BTS comparing to different SSL methods, i.e., CsiGAN and MPL. (a) Dataset in round 1 is labeled and that in round 2 is unlabeled. (b) Dataset in round 1 is labeled and that in round 6 is unlabeled.}
        \label{fig:f9}
\end{figure}

As presented in \fig \ref{fig:f9_1}, we compare BTS with two SSL, i.e., CsiGAN \cite{CsiGAN} and MPL \cite{MPL}. We consider round 1 as the labeled dataset and either round 2 in Fig. \ref{fig:f9_1} or round 6 in Fig. \ref{fig:f9_2} as the unlabeled one, while evaluate data in round 3. We can observe that the highest accuracy can be reached in both plots in the empty case due to simple distinguishment from features. In Fig. \ref{fig:f9_1}, both CsiGAN and BTS both achieve a performance greater than $95\%$ in the case of human presence in room B. It can be seen that CsiGAN and MPL fail to distinguish the cases of presence in room A and both rooms effectively. This is because these benchmarks potentially overfit the features of room B, which results in lower accuracy in the other cases. In contrast, BTS improves upon the original MPL architecture by detecting data drift and considering time-varying effects, which results in a better accuracy. As illustrated in \fig \ref{fig:f9_2}, it reveals a more severe case that considers unlabeled drifted data in round 6. Since MPL does not consider time-varying effects, it can only distinguish the differences between the cases of empty room and human presence in both rooms. Although CsiGAN is capable of generically augmenting adversarial data with similar features, confusing data features in the cases of presence in either room A or B still lead to a lower accuracy than proposed BTS scheme.

We summarize the results of our experiments in Table \ref{summary}. In the SL block, "Teacher (round 1)" indicates that we use a teacher model to conduct supervised learning only with round 1 as the labeled dataset, while "Teacher (rounds 1 and 2)" means that we adopt both rounds 1 and 2 as labeled datasets. Our proposed BTS system outperforms the other SL methods in rounds 3 and 5 and is comparable to other rounds utilizing supervised learning with an accuracy difference lower than $5\%$. Furthermore, in SSL block, "Single ResNet/Transformer TS" indicates that we utilize a single teacher-student network based on either ResNet or Transformer, respectively. Overall, our proposed BTS system, which leverages all designed loss functions and a joint architecture of transformer and ResNet, outperforms other existing SSL benchmarks in all testing rounds by extracting temporal, spatial, and spectral features.

\subsection{Computational Complexity}

\begin{table}[!t]
\centering
\scriptsize
\caption{Computational Complexity}
\begin{tabular}{|l|c|c|c|c|}
\hline
Model & Parameters & \tabincell{c}{Memory \\ Size} & \tabincell{c}{Training\\ Time} & \tabincell{c}{Inference\\ Time} \\ \hline
Parallel CNN \cite{Deep_Learning2} & 789,428 & 3.01 MB & 3.06 s & 216.57 ms \\ \hline
AE-LRCN \cite{Deep_Learning6} & 2,605,552 & 9.94 MB & 1.56 s & 400.85 ms \\ \hline
Single ResNet TS & 11,181,700 & 42.65 MB & 1.49 s & 101.68 ms \\ \hline
Single Transformer TS & 898,852 & 3.43 MB & 1.30 s & 82.64 ms \\ \hline
CsiGAN \cite{CsiGAN} & 40,404,797 & 154.13 MB & 12.95 s & 496.24 ms \\ \hline
MPL \cite{MPL} & 1,466,980 & 5.60 MB & 2.02 s & 281.63 ms \\ \hline
BTS & 24,490,704 & 93.42 MB & 7.45 s & 181.23 ms \\ \hline
\end{tabular} \label{complexity}
\end{table}

Here, we present the analysis of computational complexity of the proposed BTS compared to the benchmarks in terms of the required total number of deep learning model parameters, required memory size per epoch, and training/inference time. Note that training time is calculated per epoch, whilst testing time is averaged over each batch. We evaluate the system with the 12th Gen Intel(R) Core(TM) i9-12900 central processing unit (CPU) and NVIDIA GeForce RTX 4060 graphics processing Unit (GPU). As shown in Table \ref{complexity}, our BTS system requires a total of 24.49 M parameters to be trained owing to the complex TS models reckoning with the respective issues in human presence detection. An acceptable memory size of 93.42 MB is required for BTS compared to benchmark of CsiGAN. The training process of BTS takes the moderate time of 7.45 seconds per epoch, whereas it takes around 181.23 ms of inference time for each batch owing to the data preprocessing and four sub-architectures of BTS. We further notice that the proposed BTS system possesses the same input dimension under the varying numbers of human presence in the room. Since we only detect the human presence and room vacancy, different numbers of people indoor will not affect the computational complexity. To conclude, the proposed BTS system is feasible and implementable in practical applications, which outperforms the other existing solutions in the open literature.

\section{Conclusion} \label{conclusion}

\subsection{Conclusion Remarks}
We have addressed the challenges in indoor adjoining room human presence detection by developing a semi-supervised CSI-based BTS system. The primal/dual TS networks are designed to learn temporal, spatial and spectral features based on transformer and ResNet using SSL learning mechanism. Several loss functions are conceived to deal with time-varying CSI and data drift according to confidence distribution and outlier distance. In experimental trails, we deploy two commercial APs acting as TX/RX respectively in two adjoining room as well as an edge server for human presence detection training and testing. We have evaluated different hyperparameters and network settings as well as all the designed loss functions. In conclusion, our proposed BTS system shows superior performance with an averaged accuracy of around 98\% in the human presence detection task when compared to existing SSL methods. Notably, our system reaches the asymptotic 98\% accuracy of supervised learning, proving its effectiveness in reducing the labeling cost while maintaining high accuracy.

\subsection{Future Potentials}

The proposed BTS system can be extended to more rooms, with each pair conducting the identical algorithm. Owing to multiple overlapped results, additional voting mechanism or fusion technique can be conducted to predict the most possible outcome. However, the increment of number of rooms indicates the higher AP deployment cost and higher overhead of data collection. Also, training of multi-BTS systems will lead to a higher computational complexity order. 
We further notice that current model may face limitations in real-time or large-scale deployments, especially in resource-constrained environments. Such issues can be solved by designing techniques, including (1) model compression and pruning significantly reducing the parameter count and memory footprint while retaining performance, (2) knowledge distillation transferring the learned knowledge of the larger model into a smaller and more efficient one, and (3) optimizations specific to edge hardware such as advanced GPUs and tensor processing units (TPUs) to accelerate both training and inference processes.

Moreover, indoor layout may be complex and nearby corridors or aisles, where undesired people will be detected and regarded as interference. Such case requires further design by removing those undesirable data features. Once the data drift is much severe for the edge to recognize, it will be regarded as a new environment according to the predefined threshold. Therefore, reinforcement learning based technique can be further incorporated into BTS to dynamically measure and adjust the threshold of data drift detection and retraining period. Moreover, with larger room space, AP requires better deployment policy or more equipped antennas for optimally covering the whole space. Higher transmit power is also required for detecting the CSI signals. Under the fixed network architecture, deploying the cost-effective metasufaces having comparably higher channel diversities become a promising auxiliary \cite{future1}. Higher resolution of CSI data can be obtained with the aid of metasufaces. With higher dimension of data, more sensing tasks such as people counting, localization, and tracking can be simultaneously conducted.

\footnotesize
\bibliographystyle{IEEEtran}
\bibliography{IEEEabrv,myReference}

\begin{thebibliography}{10}
\providecommand{\url}[1]{#1}
\csname url@samestyle\endcsname
\providecommand{\newblock}{\relax}
\providecommand{\bibinfo}[2]{#2}
\providecommand{\BIBentrySTDinterwordspacing}{\spaceskip=0pt\relax}
\providecommand{\BIBentryALTinterwordstretchfactor}{4}
\providecommand{\BIBentryALTinterwordspacing}{\spaceskip=\fontdimen2\font plus
\BIBentryALTinterwordstretchfactor\fontdimen3\font minus
  \fontdimen4\font\relax}
\providecommand{\BIBforeignlanguage}[2]{{%
\expandafter\ifx\csname l@#1\endcsname\relax
\typeout{** WARNING: IEEEtran.bst: No hyphenation pattern has been}%
\typeout{** loaded for the language `#1'. Using the pattern for}%
\typeout{** the default language instead.}%
\else
\language=\csname l@#1\endcsname
\fi
#2}}
\providecommand{\BIBdecl}{\relax}
\BIBdecl

\bibitem{add5}
Y.~Ge, A.~Taha, S.~A. Shah, K.~Dashtipour, S.~Zhu, J.~Cooper, Q.~H. Abbasi, and
  M.~A. Imran, ``Contactless wifi sensing and monitoring for future healthcare
  - emerging trends, challenges, and opportunities,'' \emph{IEEE Reviews in
  Biomedical Engineering}, vol.~16, pp. 171--191, 2023.

\bibitem{Smartphone1}
J.~Jiao, F.~Li, Z.~Deng, and W.~Ma, ``A smartphone camera-based indoor
  positioning algorithm of crowded scenarios with the assistance of deep
  {CNN},'' \emph{Sensors}, vol.~17, no.~4, 2017.

\bibitem{Smartphone2}
M.~Oplenskedal, P.~Herrmann, and A.~Taherkordi, ``{DeepMatch2}: A comprehensive
  deep learning-based approach for in-vehicle presence detection,''
  \emph{Information Systems}, vol. 108, 2022.

\bibitem{Smartwatch1}
A.~Filippoupolitis, W.~Oliff, B.~Takand, and G.~Loukas, ``Location-enhanced
  activity recognition in indoor environments using off the shelf smart watch
  technology and {BLE} beacons,'' \emph{Sensors}, vol.~17, no.~6, 2017.

\bibitem{Smartwatch2}
R.~Dai, C.~Lu, M.~Avidan, and T.~Kannampallil, ``{RespWatch}: Robust
  measurement of respiratory rate on smartwatches with photoplethysmography,''
  in \emph{Proceedings of ACM/IEEE International Conference on
  Internet-of-Things Design and Implementation (IoTDI)}, 2021, pp. 208--220.

\bibitem{IPcamera1}
L.~Zhang, T.~Zhou, and B.~Lian, ``Integrated {IMU} with faster {R-CNN} aided
  visual measurements from {IP} cameras for indoor positioning,''
  \emph{Sensors}, vol.~18, no.~9, 2018.

\bibitem{IPcamera2}
R.~Marroquin, J.~Dubois, and C.~Nicolle, ``{WiseNET}: An indoor multi-camera
  multi-space dataset with contextual information and annotations for people
  detection and tracking,'' \emph{Data in Brief}, vol.~27, 2019.

\bibitem{IPcamera3}
L.~Chen and S.~Li, ``Human motion target posture detection algorithm using
  semi-supervised learning in internet of things,'' \emph{IEEE Access}, vol.~9,
  pp. 90\,529--90\,538, 2021.

\bibitem{add1}
J.~Cha, K.~Yoo, D.~Choi, and Y.~Kim, ``Human presence detection using
  ultrashort-range {FMCW} radar based on {DCNN},'' \emph{IEEE Sensors Journal},
  vol.~24, no.~16, pp. 26\,258--26\,265, 2024.

\bibitem{add2}
R.~Shahbazian and I.~Trubitsyna, ``Human sensing by using radio frequency
  signals: A survey on occupancy and activity detection,'' \emph{IEEE Access},
  vol.~11, pp. 40\,878--40\,904, 2023.

\bibitem{CSI1}
J.~Xiao, K.~Wu, Y.~Yi, and L.~M. Ni, ``{FIFS}: Fine-grained indoor
  fingerprinting system,'' in \emph{Proceedings of IEEE International
  Conference on Computer Communications and Networks (ICCCN)}, 2012, pp. 1--7.

\bibitem{CSI2}
R.~Zhou, X.~Lu, P.~Zhao, and J.~Chen, ``Device-free presence detection and
  localization with {SVM} and {CSI} fingerprinting,'' \emph{IEEE Sensors
  Journal}, vol.~17, no.~23, pp. 7990--7999, 2017.

\bibitem{CSI3}
J.~Lv, D.~Man, W.~Yang, X.~Du, and M.~Yu, ``Robust {WLAN}-based indoor
  intrusion detection using {PHY} layer information,'' \emph{IEEE Access},
  vol.~6, pp. 30\,117--30\,127, 2018.

\bibitem{acm}
L.-H. Shen, K.-T. Feng, and L.~Hanzo, ``Five facets of {6G}: Research
  challenges and opportunities,'' \emph{ACM Computing Surveys}, vol.~55,
  no.~11, pp. 1--39, 2023.

\bibitem{add6}
V.~V. Ratnam, H.~Chen, H.~H. Chang, A.~Sehgal, and J.~Zhang, ``Optimal
  preprocessing of wifi {CSI} for sensing applications,'' \emph{IEEE
  Transactions on Wireless Communications}, pp. 1--1, 2024.

\bibitem{add3}
Y.~Gu, J.~Chen, K.~He, C.~Wu, Z.~Zhao, and R.~Du, ``Wifileaks: Exposing
  stationary human presence through a wall with commodity mobile devices,''
  \emph{IEEE Transactions on Mobile Computing}, vol.~23, no.~6, pp. 6997--7011,
  2024.

\bibitem{Deep_Learning1}
Y.-M. Huang, A.-H. Hsiao, C.-J. Chiu, K.-T. Feng, and P.-H. Tseng,
  ``Device-free multiple presence detection using {CSI} with machine learning
  methods,'' in \emph{Proceedings of IEEE Vehicular Technology Conference
  (VTC-Fall)}, 2019, pp. 1--5.

\bibitem{Deep_Learning2}
Y.~Liu, T.~Wang, Y.~Jiang, and B.~Chen, ``Harvesting ambient {RF} for presence
  detection through deep learning,'' \emph{IEEE Transactions on Neural Networks
  and Learning Systems}, vol.~33, no.~4, pp. 1571--1583, 2022.

\bibitem{Deep_Learning3}
Y.~Zhang, C.~Qu, and Y.~Wang, ``An indoor positioning method based on {CSI} by
  using features optimization mechanism with {LSTM},'' \emph{IEEE Sensors
  Journal}, vol.~20, no.~9, pp. 4868--4878, 2020.

\bibitem{Deep_Learning4}
S.~M. Bokhari, S.~Sohaib, A.~R. Khan, M.~Shafi, and A.~U.~R. Khan, ``{DGRU}
  based human activity recognition using channel state information,''
  \emph{Measurement}, vol. 167, 2021.

\bibitem{Deep_Learning5}
F.-Y. Chu, C.-J. Chiu, A.-H. Hsiao, K.-T. Feng, and P.-H. Tseng, ``Wifi
  {CSI}-based device-free multi-room presence detection using conditional
  recurrent network,'' in \emph{Proceedings of IEEE Vehicular Technology
  Conference (VTC-Spring)}, 2021, pp. 1--5.

\bibitem{Deep_Learning6}
H.~Zou, Y.~Zhou, J.~Yang, H.~Jiang, L.~Xie, and C.~J. Spanos, ``{DeepSense}:
  Device-free human activity recognition via autoencoder long-term recurrent
  convolutional network,'' in \emph{Proceedings of IEEE International
  Conference on Communications (ICC)}, 2018, pp. 1--6.

\bibitem{Deep_Learning7}
Q.~Li, H.~Qu, Z.~Liu, N.~Zhou, W.~Sun, S.~Sigg, and J.~Li, ``{AF-DCGAN}:
  Amplitude feature deep convolutional {GAN} for fingerprint construction in
  indoor localization systems,'' \emph{IEEE Transactions on Emerging Topics in
  Computational Intelligence}, vol.~5, no.~3, pp. 468--480, 2021.

\bibitem{Deep_Learning8}
X.~Wang, L.~Gao, S.~Mao, and S.~Pandey, ``{CSI}-based fingerprinting for indoor
  localization: A deep learning approach,'' \emph{IEEE Transactions on
  Vehicular Technology}, vol.~66, no.~1, pp. 763--776, 2017.

\bibitem{Deep_Learning9}
X.~Wang, L.~Gao, and S.~Mao, ``{CSI} phase fingerprinting for indoor
  localization with a deep learning approach,'' \emph{IEEE Internet of Things
  Journal}, vol.~3, no.~6, pp. 1113--1123, 2016.

\bibitem{add7}
Y.~Zhang, G.~Wang, H.~Liu, W.~Gong, and F.~Gao, ``Wifi-based indoor human
  activity sensing: A selective sensing strategy and a multilevel feature
  fusion approach,'' \emph{IEEE Internet of Things Journal}, vol.~11, no.~18,
  pp. 29\,335--29\,347, 2024.

\bibitem{add4}
I.~Ahmad, A.~Ullah, and W.~Choi, ``Wifi-based human sensing with deep learning:
  Recent advances, challenges, and opportunities,'' \emph{IEEE Open Journal of
  the Communications Society}, vol.~5, pp. 3595--3623, 2024.

\bibitem{new_attention}
L.-H. Shen, A.-H. Hsiao, K.-I. Lu, and K.-T. Feng, ``Attention-enhanced deep
  learning for device-free through-the-wall presence detection using indoor
  wifi systems,'' \emph{IEEE Sensors Journal}, vol.~24, no.~4, pp. 5288--5302,
  2024.

\bibitem{tk_rnn}
A.~Bryant and K.~Cios, ``{RNN-DBSCAN}: A density-based clustering algorithm
  using reverse nearest neighbor density estimates,'' \emph{IEEE Transactions
  on Knowledge and Data Engineering}, vol.~30, no.~6, pp. 1109--1121, 2018.

\bibitem{new_chu}
L.-H. Shen, A.-H. Hsiao, F.-Y. Chu, and K.-T. Feng, ``Time-selective {RNN} for
  device-free multiroom human presence detection using wifi {CSI},'' \emph{IEEE
  Transactions on Instrumentation and Measurement}, vol.~73, pp. 1--17, 2024.

\bibitem{tk_lstm}
Q.~Li, J.~Tan, J.~Wang, and H.~Chen, ``A multimodal event-driven {LSTM} model
  for stock prediction using online news,'' \emph{IEEE Transactions on
  Knowledge and Data Engineering}, vol.~33, no.~10, pp. 3323--3337, 2021.

\bibitem{time_varying}
G.~Lebrun, J.~Gao, and M.~Faulkner, ``{MIMO} transmission over a time-varying
  channel using {SVD},'' \emph{IEEE Transactions on Wireless Communications},
  vol.~4, no.~2, pp. 757--764, 2005.

\bibitem{tk_ssl}
X.~Yang, Z.~Song, I.~King, and Z.~Xu, ``A survey on deep semi-supervised
  learning,'' \emph{IEEE Transactions on Knowledge and Data Engineering}, pp.
  1--20, 2022.

\bibitem{MCBAR}
D.~Wang, J.~Yang, W.~Cui, L.~Xie, and S.~Sun, ``Multimodal {CSI}-based human
  activity recognition using {GANs},'' \emph{IEEE Internet of Things Journal},
  vol.~8, no.~24, pp. 17\,345--17\,355, 2021.

\bibitem{CsiGAN}
C.~Xiao, D.~Han, Y.~Ma, and Z.~Qin, ``{CsiGAN}: Robust channel state
  information-based activity recognition with {GANs},'' \emph{IEEE Internet of
  Things Journal}, vol.~6, no.~6, pp. 10\,191--10\,204, 2019.

\bibitem{DADA}
Y.-S. Chen, Y.-C. Chang, and C.-Y. Li, ``A semi-supervised transfer learning
  with dynamic associate domain adaptation for human activity recognition using
  wifi signals,'' \emph{Sensors}, vol.~21, no.~24, 2021.

\bibitem{SemiC_HAR}
D.~Liu and T.~Abdelzaher, ``Semi-supervised contrastive learning for human
  activity recognition,'' in \emph{Proceedings of IEEE International Conference
  on Distributed Computing in Sensor Systems (DCOSS)}, 2021, pp. 45--53.

\bibitem{CycleGAN}
J.-Y. Zhu, T.~Park, P.~Isola, and A.~A. Efros, ``Unpaired image-to-image
  translation using cycle-consistent adversarial networks,'' in
  \emph{Proceedings of IEEE/CVF International Conference on Computer Vision
  (ICCV)}, 2017.

\bibitem{kn}
G.~Hinton, O.~Vinyals, and J.~Dean, ``Distilling the knowledge in a neural
  network,'' in \emph{Proceedings of Advances in Neural Information Processing
  Systems (NeurIPS)}, 2015.

\bibitem{MPL}
H.~Pham, Z.~Dai, Q.~Xie, and Q.~V. Le, ``Meta pseudo labels,'' in
  \emph{Proceedings of IEEE/CVF Conference on Computer Vision and Pattern
  Recognition (CVPR)}, 2021, pp. 11\,557--11\,568.

\bibitem{tk_ts}
X.~Zhao, M.~Yang, Q.~Qu, R.~Xu, and J.~Li, ``Exploring privileged features for
  relation extraction with contrastive student-teacher learning,'' \emph{IEEE
  Transactions on Knowledge and Data Engineering}, pp. 1--1, 2022.

\bibitem{Attention}
A.~Vaswani, N.~Shazeer, N.~Parmar, J.~Uszkoreit, L.~Jones, A.~N. Gomez,
  L.~Kaiser, and I.~Polosukhin, ``Attention is all you need,'' in
  \emph{Proceedings of Advances in Neural Information Processing Systems
  (NeurIPS)}, vol.~30, 2017.

\bibitem{ResNet}
K.~He, X.~Zhang, S.~Ren, and J.~Sun, ``Deep residual learning for image
  recognition,'' in \emph{Proceedings of IEEE/CVF Conference on Computer Vision
  and Pattern Recognition (CVPR)}, 2016, pp. 770--778.

\bibitem{SAS_PD}
K.-J. Chen, A.-H. Hsiao, C.-J. Chiu, and K.-T. Feng, ``Self-attention based
  semi-supervised learning for time-varying wi-fi {CSI}-based adjoining room
  presence detection,'' in \emph{Proceedings of IEEE Vehicular Technology
  Conference (VTC-Spring)}, 2022, pp. 1--5.

\bibitem{new_cronos}
L.-H. Shen, C.-C. Hsieh, A.-H. Hsiao, and K.-T. Feng, ``{CRONOS}: Colorization
  and contrastive learning for device-free {NLoS} human presence detection
  using wi-fi {CSI},'' \emph{IEEE Internet of Things Journal}, vol.~11, no.~3,
  pp. 5491--5510, 2024.

\bibitem{DeepFi}
X.~Wang, L.~Gao, S.~Mao, and S.~Pandey, ``{DeepFi}: Deep learning for indoor
  fingerprinting using channel state information,'' in \emph{Proceedings of
  IEEE Wireless Communications and Networking Conference (WCNC)}, 2015, pp.
  1666--1671.

\bibitem{contrastive}
T.~Chen, S.~Kornblith, M.~Norouzi, and G.~Hinton, ``A simple framework for
  contrastive learning of visual representations,'' in \emph{Proceedings of ACM
  International Conference on Machine Learning (ICML)}, 2020, pp. 1597--1607.

\bibitem{BYOL}
J.-B. Grill, F.~Strub, F.~Altch\'{e}, C.~Tallec, P.~Richemond, E.~Buchatskaya,
  C.~Doersch, B.~Avila~Pires, Z.~Guo, M.~Gheshlaghi~Azar, B.~Piot,
  k.~kavukcuoglu, R.~Munos, and M.~Valko, ``Bootstrap your own latent - a new
  approach to self-supervised learning,'' in \emph{Proceedings of Advances in
  Neural Information Processing Systems (NeurIPS)}, vol.~33, 2020, pp.
  21\,271--21\,284.

\bibitem{csitool}
Y.~Xie, Z.~Li, and M.~Li, ``Precise power delay profiling with commodity
  wi-fi,'' \emph{IEEE Transactions on Mobile Computing}, vol.~18, no.~6, pp.
  1342--1355, 2019.

\bibitem{future1}
L.-H. Shen, K.-T. Feng, T.-S. Lee, Y.-C. Lin, S.-C. Lin, C.-C. Chang, and S.-F.
  Chang, ``{AI}-enabled unmanned vehicle-assisted reconfigurable intelligent
  surfaces: Deployment, prototyping, experiments, and opportunities,''
  \emph{IEEE Network}, pp. 1--1, 2024.

\end{thebibliography}

\end{document}